\documentclass{article}
\usepackage{iclr2027_conference,times}

\usepackage{amsmath,amsfonts,bm}

\def\eqref#1{equation~\ref{#1}}
\def\1{\bm{1}}

\DeclareMathAlphabet{\mathsfit}{\encodingdefault}{\sfdefault}{m}{sl}
\SetMathAlphabet{\mathsfit}{bold}{\encodingdefault}{\sfdefault}{bx}{n}

\usepackage{hyperref}
\usepackage{url}
\usepackage{graphicx}
\usepackage{wrapfig}
\usepackage{booktabs}
\usepackage{nicematrix}
\usepackage{tikz}
\usepackage{diagbox}
\usepackage{subcaption}
\usepackage{tabularx}
\usepackage{array}
\usepackage{multirow}
\usepackage[most]{tcolorbox}
\usepackage{listings}
\usepackage[normalem]{ulem}
\usepackage[table]{xcolor}
\usepackage{bbm}
\usepackage{adjustbox}
\usepackage{arydshln}

\usetikzlibrary{positioning,calc,arrows.meta,shapes.geometric,fit,backgrounds,spy}

\hypersetup{
    colorlinks=true,
    allcolors=blue!50!black
}

\renewcommand{\arraystretch}{1.2}

\newcommand{\redstrike}[1]{%
  \textcolor{red}{\sout{\textcolor{black}{#1}}}%
}
\newcommand{\blueadd}[1]{%
  \textcolor{blue!75!black}{#1}%
}
\newcommand{\greencorr}[1]{%
  \textcolor{green!45!black}{#1}%
}
\newcommand{\smath}[1]{\scalebox{0.65}{$#1$}}
\newcommand{\smathsmall}[1]{\scalebox{0.55}{$#1$}}
\newcommand{\headercell}[1]{\normalsize\textbf{#1}}
\definecolor{promptAccent}{RGB}{15,56,112}
\definecolor{promptBack}{RGB}{245,248,255}

\lstdefinestyle{mystyle}{
  basicstyle=\fontsize{6}{4.5}\ttfamily,
  breaklines=true,
  frame=single,
  backgroundcolor=\color{gray!5},
  columns=fullflexible
}

\newtcolorbox{PromptBox}[2][]{%
  enhanced, breakable,
  colback=promptBack, colframe=promptAccent,
  coltitle=white, colbacktitle=promptAccent,
  fonttitle=\bfseries\small\ttfamily,
  fontupper=\ttfamily\scriptsize,
  title={#2},
  boxed title style={boxrule=0pt, colframe=promptAccent, colback=promptAccent,
    top=2pt, bottom=2pt, left=6pt, right=6pt},
  attach boxed title to top left={yshift=-1.5mm, xshift=4mm},
  boxrule=0.75pt, arc=2pt, boxsep=5pt,
  before skip=6pt, after skip=6pt,
  borderline west={2pt}{0pt}{promptAccent},
  drop shadow southeast,
  sharp corners=south,
  before upper={\vspace{2pt}},
  #1
}

\title{%
  \raisebox{-1.5ex}{\includegraphics[height=1.2cm]{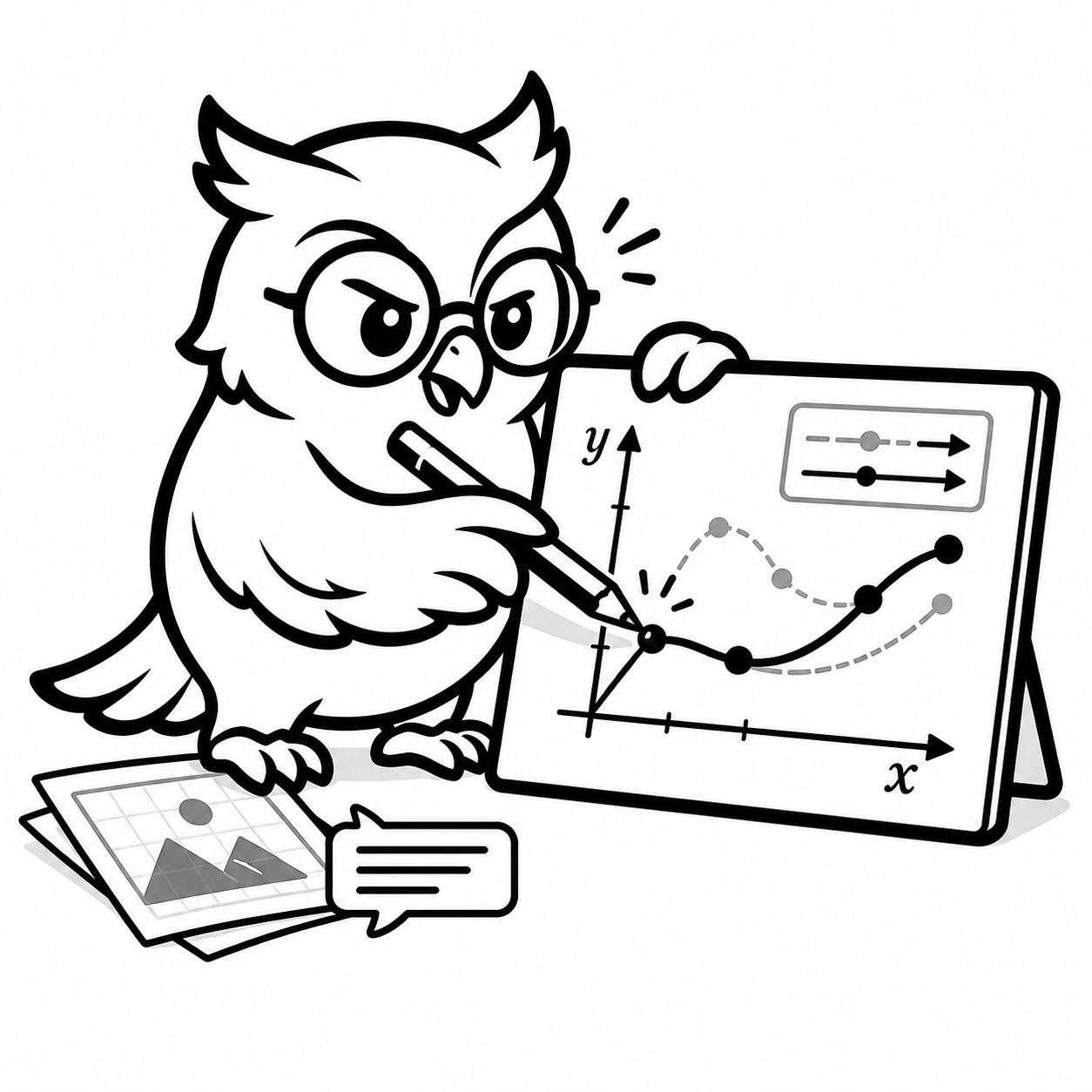}}%
  \hspace{0.05em}
  EdiTikZ: Scientific Figure Editing from Revision Trajectories
}

\author{%
Christian Greisinger\textsuperscript{1}, Zhixue Zhao\textsuperscript{2}, Steffen Eger\textsuperscript{1}\\
{\small
\textsuperscript{1}University of Technology Nuremberg \texttt{\{christian.greisinger,steffen.eger\}@utn.de}}\\
{\small
\textsuperscript{2}University of Sheffield \texttt{zhixue.zhao@sheffield.ac.uk}}%
}

\iclrfinalcopy
\begin{document}

\maketitle
\lhead{Preprint}

\begin{abstract}
Vision-language models (VLMs) have shown strong performance in generating scientific figures from text or images. However, publication-ready figures often require iterative refinement, making scientific figure editing an important yet largely unexplored step toward interactive figure creation. 
Existing approaches rely on costly proprietary agentic systems, focus primarily on evaluation, or construct training supervision from synthetically generated edits. Instead, we leverage naturally occurring scientific revision and development trajectories as a scalable source of supervision. To this end, we introduce DaEdiTikZ, the first large-scale dataset of revision-derived scientific figure edits, constructed by mining 391K plausible TikZ edit pairs from arXiv, GitHub, and TeX SE and inferring 781K directed edit instructions with a VLM conditioned on rendered figures and TikZ code. We further introduce DaEdiTikZ-Bench, a human-refined benchmark with 690 instances, and train two compact Qwen3.5-based EdiTikZ models (4B and 9B) by jointly learning image-to-TikZ reconstruction and instruction-conditioned editing, followed by reinforcement learning (RL) with complementary rewards for rendered fidelity and edit application. Automatic evaluation places our 9B model above all tested baselines, while human evaluation with 9 annotators and 4,320 ratings places it above GPT-5.6-Sol and on par with Gemini-3.1-Pro. Under severe out-of-distribution shifts, it remains competitive with GPT-5.6-Sol near its 2K training sequence-length regime. Models and datasets are available on
\href{https://huggingface.co/collections/nllg/editikz}{%
  \raisebox{-0.18\height}{\includegraphics[height=1.1em]{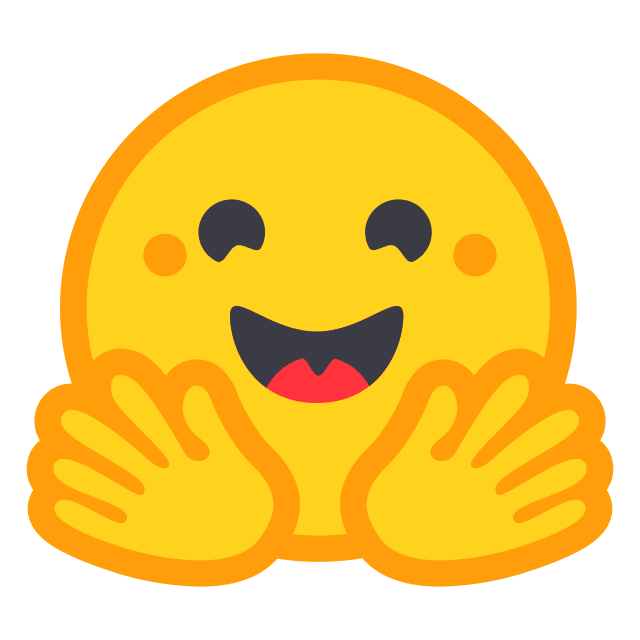}}%
  \,Hugging Face}.
Code will be released on
\href{https://github.com/NL2G/EdiTikZ}{%
  \raisebox{-0.18\height}{\includegraphics[height=1.1em]{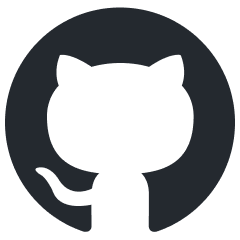}}%
  \,GitHub}.
\end{abstract}

\section{Introduction}
\label{sec:Introduction}

VLMs are increasingly capable of assisting researchers in multimodal tasks~\citep{eger2026transformingsciencelargelanguage}, including understanding and generating figures~\citep{li-etal-2024-multimodal-arxiv, NEURIPS2024_CharXiv}, tables~\citep{NEURIPS_SciGen}, slides~\citep{Ge_2025_CVPR_AutoPresent}, and posters~\citep{NEURIPS2025_Paper2Poster}. These advances are driven by improvements in multimodal alignment~\citep{NEURIPS2023_VisualInstructionTuning}, reasoning~\citep{zhang2024multimodalchainofthoughtreasoninglanguage, huang2026visionr1incentivizingreasoningcapability}, and agentic systems~\citep{koh-etal-2024-visualwebarena} that combine planning and tool use to tackle complex scientific workflows~\citep{sun2026scienceboardevaluatingmultimodalautonomous}. Despite this progress, publication-ready figures rarely emerge in a single generation and typically require revisions to their content, layout, and visual details, making scientific figure editing an important yet underexplored capability.

Graphics programming languages such as TikZ are the de facto standard in academia due to their precision, interpretability, and seamless integration into the \LaTeX{} ecosystem. However, their diverse syntax and steep learning curve make them difficult for humans to master~\citep{belouadi2024automatikz}. Prior work has focused on generating TikZ from text~\citep{greisinger2026tikzilla} or images~\citep{belouadi2024detikzify}, whereas existing editing efforts rely on proprietary agentic systems~\citep{lin2026autofigureeditgeneratingeditablescientific}, target specialized domains such as charts~\citep{zhao-etal-2025-chartedit}, or focus primarily on evaluation~\citep{rahman2026viseditbenchvisionlanguagemodelsedit,bo2026diagrammmumultimodalbenchmarkscientific}. Large-scale training supervision remains limited and predominantly synthetic~\citep{wang2026disciplinegen1mlargescaledatasetmultidisciplinary,bo2026diagrammmumultimodalbenchmarkscientific}.

In this work, we take a different perspective. Scientific figures naturally evolve through iterative human revisions during research, paper writing, and community discussions. These revisions capture rich but previously overlooked expert decisions about how figures should change, yet remain unused as supervision for multimodal models. Inspired by how early instruction-tuning methods leverage naturally occurring software revisions~\citep{muennighoff2024octopack, wei2024coeditor, li-etal-2024-instructcoder}, we introduce a scalable framework that recovers plausible scientific figure revision pairs from real-world repositories. Applied to TikZ figures from arXiv, GitHub, and TeX SE, this yields DaEdiTikZ, the first large-scale dataset of revision-derived scientific figure edits, containing 391K edit pairs. Since figures and their programs already exist, we synthesize only the missing edit instruction using a VLM conditioned on rendered figures and TikZ code, yielding 781K directed editing instances. We also introduce DaEdiTikZ-Bench, a human-refined benchmark with 690 editing instances.

Building on DaEdiTikZ, we train two small Qwen3.5-based EdiTikZ models that jointly learn to reconstruct figures as TikZ and edit them from instructions, followed by RL with complementary rewards for rendered fidelity and edit application. Across three human-evaluation criteria on DaEdiTikZ-Bench, our 9B model performs above GPT-5.6-Sol and on par with Gemini-3.1-Pro. Post-training gains transfer even beyond the 2K-token training horizon to substantially more complex out-of-distribution figures from SPIQA and CharXiv. Table~\ref{tab:editikz_examples} shows representative editing results. Our key contributions are as follows:

\setlength{\aboverulesep}{0pt}
\setlength{\belowrulesep}{0pt}
\setlength{\tabcolsep}{2pt}
\renewcommand{\arraystretch}{1.03}
\setlength{\abovetopsep}{0pt}

\newlength{\descw}
\setlength{\descw}{0.25\textwidth}

\newlength{\imgw}
\setlength{\imgw}{0.136\textwidth}

\newlength{\imgh}
\setlength{\imgh}{2.6cm}

\newcolumntype{Y}{>{\raggedright\arraybackslash}p{\descw}}
\newcolumntype{C}{>{\centering\arraybackslash}p{\imgw}}

\newcommand{\desc}[1]{%
  \begin{minipage}[t]{\linewidth}
    \vspace{-6pt}%
    \raggedright
    \fontsize{4}{4.6}\selectfont
    \setlength{\parskip}{0pt}%
    \setlength{\parindent}{0pt}%
    #1%
  \end{minipage}%
}

\newcommand{\imgcell}[1]{%
  \begin{minipage}[t]{\imgw}\vspace{-6pt}\centering
    \includegraphics[width=0.95\linewidth,height=\imgh,keepaspectratio]{#1}%
 \end{minipage}%
}
\newcommand{\imgcellbad}[2][red]{%
  \begin{minipage}[t]{\imgw}\vspace{-6pt}\centering
    \tcbox[
      enhanced,
      arc=2mm,
      colframe=#1,
      colback=#1!4,
      boxrule=0.5pt,
      left=-0.4mm,right=-0.4mm,top=-0.4mm,bottom=-0.4mm,
      nobeforeafter,
      width=\linewidth,
    ]{%
      \includegraphics[width=0.95\linewidth,height=\imgh,keepaspectratio]{#2}%
    }%
  \end{minipage}%
}
\newcommand{\imgcellokay}[2][orange]{%
  \begin{minipage}[t]{\imgw}\vspace{-6pt}\centering
    \tcbox[
      enhanced,
      arc=2mm,
      colframe=#1,
      colback=#1!4,
      boxrule=0.5pt,
      left=-0.4mm,right=-0.4mm,top=-0.4mm,bottom=-0.4mm,
      nobeforeafter,
      width=\linewidth,
    ]{%
      \includegraphics[width=0.95\linewidth,height=\imgh,keepaspectratio]{#2}%
    }%
  \end{minipage}%
}
\newcommand{\imgcellgood}[2][yellow]{%
  \begin{minipage}[t]{\imgw}\vspace{-6pt}\centering
    \tcbox[
      enhanced,
      arc=2mm,
      colframe=#1,
      colback=#1!4,
      boxrule=0.5pt,
      left=-0.4mm,right=-0.4mm,top=-0.4mm,bottom=-0.4mm,
      nobeforeafter,
      width=\linewidth,
    ]{%
      \includegraphics[width=0.95\linewidth,height=\imgh,keepaspectratio]{#2}%
    }%
  \end{minipage}%
}
\newcommand{\imgcellverygood}[2][green]{%
  \begin{minipage}[t]{\imgw}\vspace{-6pt}\centering
    \tcbox[
      enhanced,
      arc=2mm,
      colframe=#1,
      colback=#1!4,
      boxrule=0.5pt,
      left=-0.4mm,right=-0.4mm,top=-0.4mm,bottom=-0.4mm,
      nobeforeafter,
      width=\linewidth,
    ]{%
      \includegraphics[width=0.95\linewidth,height=\imgh,keepaspectratio]{#2}%
    }%
  \end{minipage}%
}

\newcommand{\legendbox}[1]{%
  \raisebox{-0.75mm}{%
    \tcbox[
      enhanced,
      arc=1mm,
      colframe=#1,
      colback=#1!15,
      boxrule=0.5pt,
      left=0mm,right=0mm,top=0mm,bottom=0mm,
      nobeforeafter,
    ]{\rule{1mm}{1mm}}%
  }%
}

\newcommand{\exrowone}[6]{%
  \imgcell{#1} &
  \desc{#2} &
  \imgcell{#3} &
  \imgcellokay{#4}
  {\fontsize{8}{8}\selectfont
  E:\textbf{4}\quad
  P:\textbf{5}\quad
  Q:\textbf{5}} 
  &
  \imgcellgood{#5}
  {\fontsize{8}{8}\selectfont
  E:\textbf{5}\quad
  P:\textbf{6}\quad
  Q:\textbf{6}} 
  &
  \imgcellgood{#6}
  {\fontsize{8}{8}\selectfont
  E:\textbf{6}\quad
  P:\textbf{6}\quad
  Q:\textbf{6}} 
  \\
}

\newcommand{\exrowtwo}[6]{%
  \imgcell{#1} &
  \desc{#2} &
  \imgcell{#3} &
  \imgcellgood{#4}
  {\fontsize{8}{8}\selectfont
  E:\textbf{5}\quad
  P:\textbf{4}\quad
  Q:\textbf{6}} 
  &
  \imgcellokay{#5}
  {\fontsize{8}{8}\selectfont
  E:\textbf{3}\quad
  P:\textbf{3}\quad
  Q:\textbf{5}} 
  &
  \imgcellverygood{#6}
  {\fontsize{8}{8}\selectfont
  E:\textbf{7}\quad
  P:\textbf{6}\quad
  Q:\textbf{7}} 
  \\
}

\newcommand{\exrowthree}[6]{%
  \imgcell{#1} &
  \desc{#2} &
  \imgcell{#3} &
  \imgcellokay{#4} 
  {\fontsize{8}{8}\selectfont
  E:\textbf{3}\quad
  P:\textbf{5}\quad
  Q:\textbf{4}} 
  &
  \imgcellbad{#5} 
  {\fontsize{8}{8}\selectfont
  E:\textbf{1}\quad
  P:\textbf{2}\quad
  Q:\textbf{2}} 
  &
  \imgcellverygood{#6} 
  {\fontsize{8}{8}\selectfont
  E:\textbf{7}\quad
  P:\textbf{7}\quad
  Q:\textbf{7}} 
  \\
}

\newcommand{\exrowfour}[6]{%
  \imgcell{#1} &
  \desc{#2} &
  \imgcell{#3} &
  \imgcellgood{#4} 
  {\fontsize{8}{8}\selectfont
  E:\textbf{6}\quad
  P:\textbf{5}\quad
  Q:\textbf{5}} 
  &
  \imgcellokay{#5} 
  {\fontsize{8}{8}\selectfont
  E:\textbf{4}\quad
  P:\textbf{3}\quad
  Q:\textbf{4}} 
  &
  \imgcellgood{#6} 
  {\fontsize{8}{8}\selectfont
  E:\textbf{6}\quad
  P:\textbf{5}\quad
  Q:\textbf{6}} 
  \\
}

\begin{table}[t]
\centering
\caption{Scientific figure edits by GPT-5.6-Sol and our EdiTikZ-9B models before and after RL. Models receive the source image and VLM-generated edit instruction. Human annotations score edit application (E), source preservation (P), and visual quality (Q). Overall quality: \legendbox{green} very good, \legendbox{yellow} good, \legendbox{orange} bad, \legendbox{red} very bad. More examples are in Appendix~\ref{subsubsec:Examples}.}
\label{tab:editikz_examples}
\begin{tabularx}{\textwidth}{@{}CYCCCC@{}}
    \toprule
    \headercell{Source} &
    \headercell{Edit Instruction} &
    \headercell{Ground Truth} &
    \headercell{GPT-5.6-Sol} &
    \headercell{EdiTikZ-9B} &
    \headercell{EdiTikZ-9B-RL} \\
    \midrule

    \exrowone
      {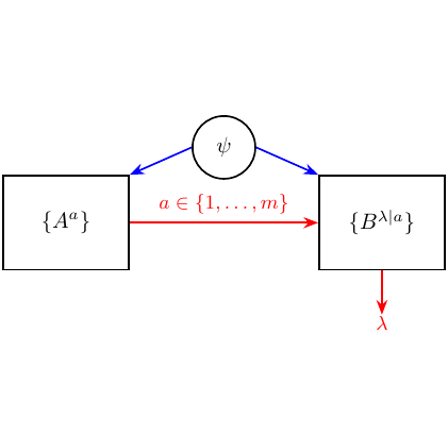}
      {The circular node labeled \smath{\psi} at the top is changed from having a black outline to having a light green fill. The rectangular boxes labeled \smath{A^a} and \smath{B^{\dots}} are changed from having a white fill to having a light blue fill. The label inside the right-hand box is changed from \smath{\{B^{\lambda|a}\}} to \smath{\{B^b\}}. A new rectangular box with a light red fill, labeled \smath{p(\lambda|a,b)}, is added below the right-hand box. The horizontal red arrow labeled \smath{a \in \{1,\dots,m\}} is rerouted. It now starts from the bottom of the left box, goes down, turns right, and points to the left side of the new red box. A new vertical red arrow labeled \smath{b} is added, connecting the bottom of the right box to the top of the new red box. The vertical red arrow labeled \smath{\lambda} is moved to originate from the bottom of the new red box instead of the right-hand box.}
      {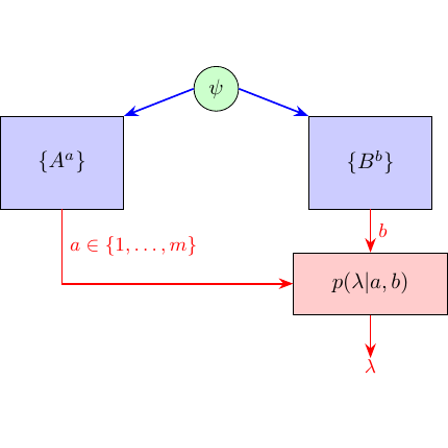}
      {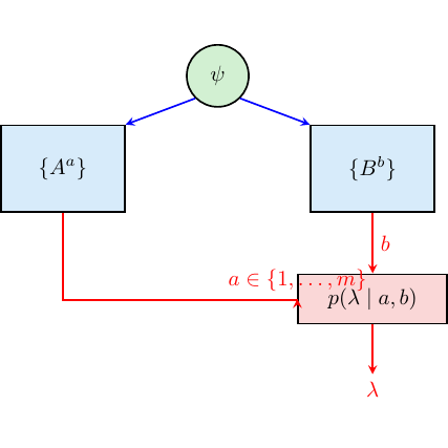}
      {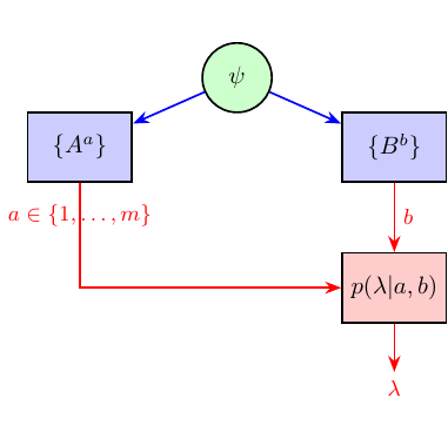}
      {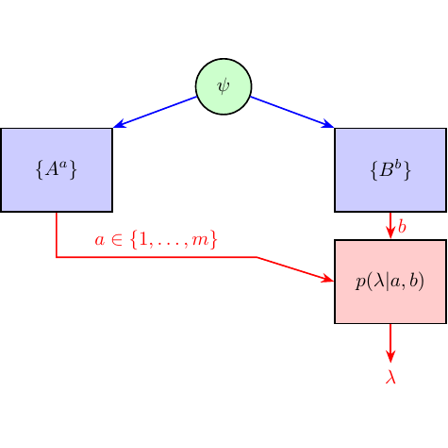}
    \midrule
    \exrowtwo
      {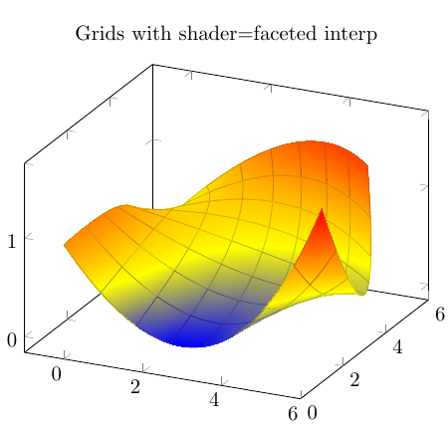}
      {The title 'Grids with shader=faceted interp' at the top of the plot is removed. The surface geometry is changed from a custom set of coordinates to a mathematical function (hyperbolic paraboloid), resulting in a saddle shape with a peak in the top-left middle and a valley in the bottom middle. The axis ranges are modified: the x-axis now spans from -2 to 2, the y-axis spans from -2 to 2, and the z-axis spans from -4 to 4.}
      {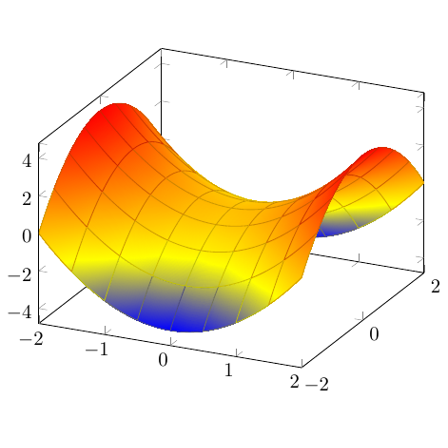}
      {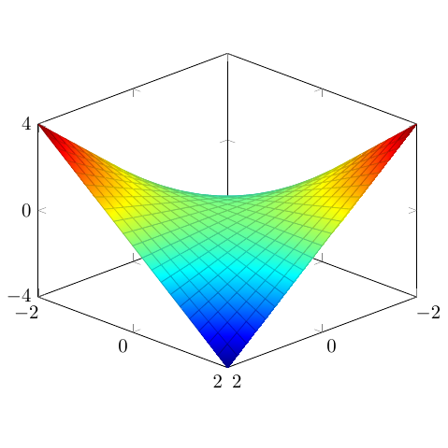}
      {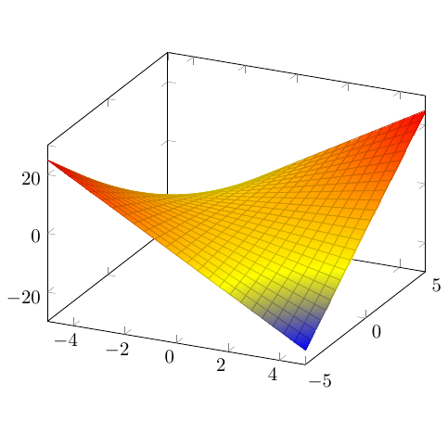}
      {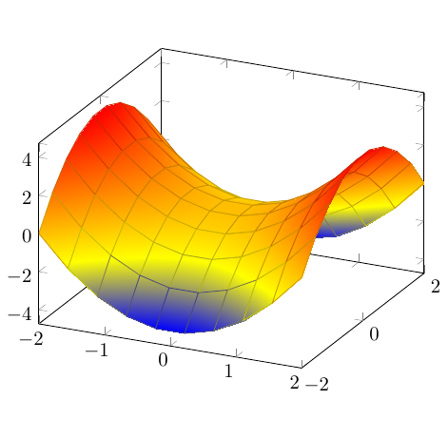}
    \midrule
    \exrowthree
      {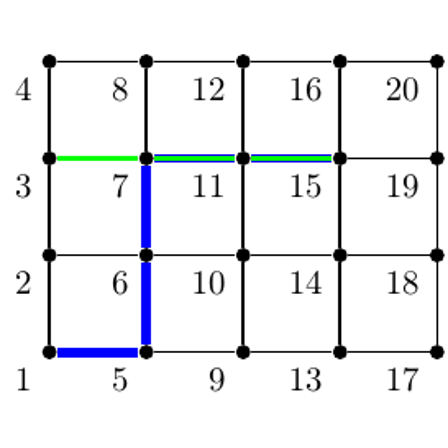}
      {The thick blue path (path1) is rerouted. In the first image, it goes from node 6 up to node 7, then right to node 11, and finally to node 15. In the second image, the path from node 6 goes right to node 10, then right to node 14, and finally up to node 15. The thin green path (path2) is rerouted. In the first image, it goes from node 7 right to node 11, then right to node 15. In the second image, the path from node 7 goes down to node 6, then right to node 10, then right to node 14, and finally up to node 15.}
      {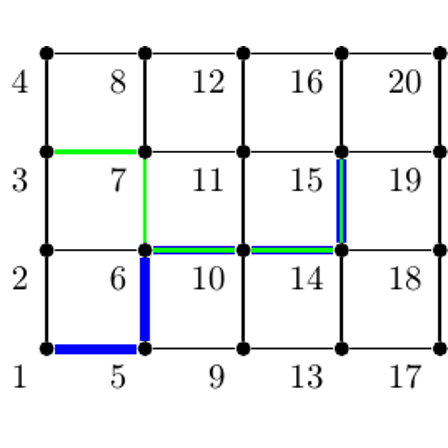}
      {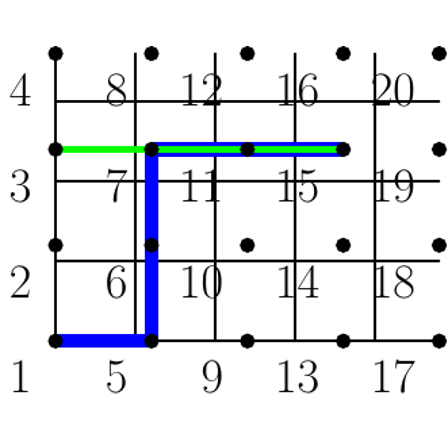}
      {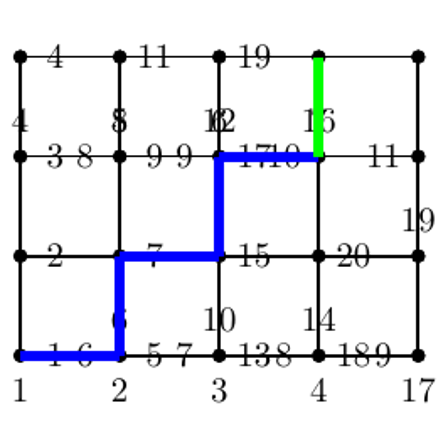}
      {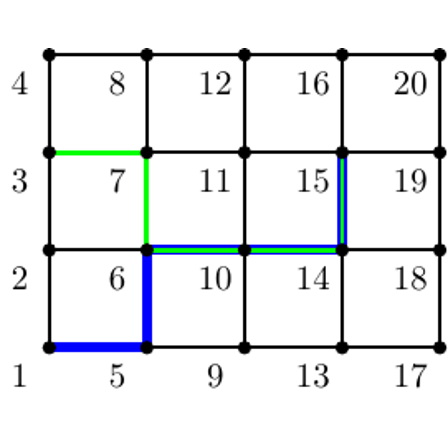}
    \midrule
    \exrowfour
      {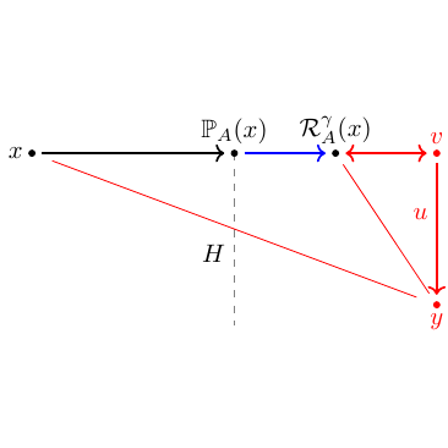}
      {The point labeled \smath{\mathcal{R}_A^\gamma(x)} is moved from the right of \smath{\mathbb{P}_A(x)} to the left of it (between \smath{x} and \smath{\mathbb{P}_A(x)}). The black arrow moves through the point \smath{\mathcal{R}_A^\gamma(x)} to point to \smath{\mathbb{P}_A(x)} while the blue arrow from \smath{\mathbb{P}_A(x)} to \smath{\mathcal{R}_A^\gamma(x)} is reversed and positioned slightly below the black arrow. The red point \smath{v} and the red point \smath{y} are moved horizontally to the left, closer to the vertical dashed line \smath{H}. The red double-headed arrow now points from \smath{\mathbb{P}_A(x)} to \smath{v}. The two red lines connecting the top points (\smath{x} and \smath{\mathcal{R}_A^\gamma(x)}) to the bottom point \smath{y} are changed to double-headed arrows.}
      {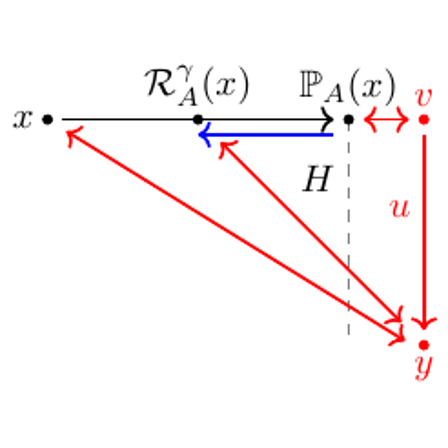}
      {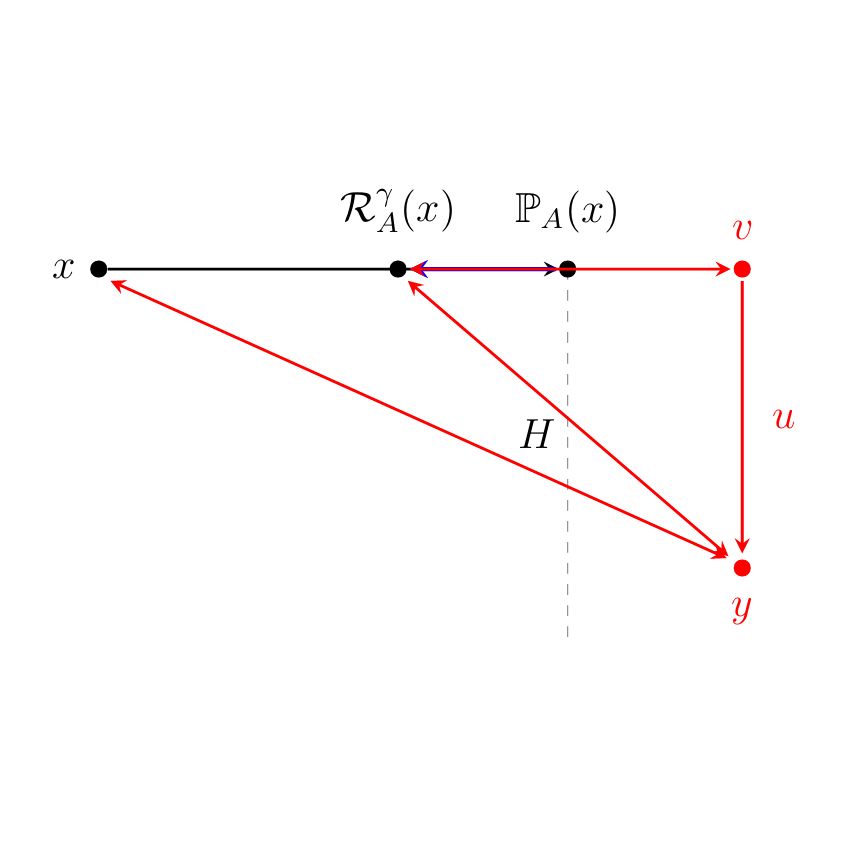}
      {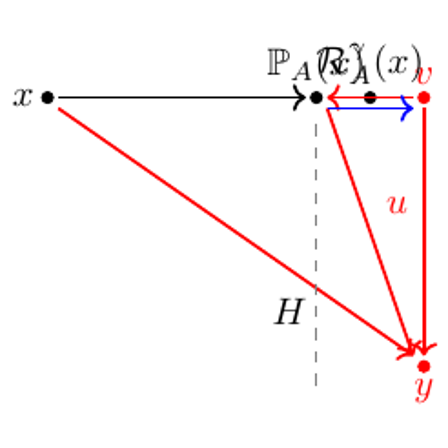}
      {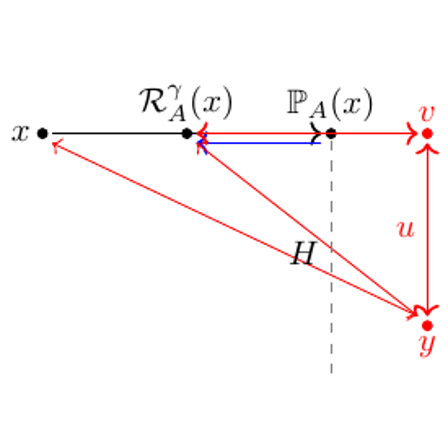}
    \bottomrule
\end{tabularx}
\end{table}

\begin{itemize}
\item \textbf{Revision-Derived Supervision:}
We introduce a scalable framework for recovering plausible edit pairs from naturally occurring collections of related scientific figures.
\item \textbf{Dataset and Benchmark:}
We release DaEdiTikZ with 391K plausible edit pairs, yielding 781K directed editing instances, and DaEdiTikZ-Bench with 690 human-refined instances.
\item \textbf{Editing-Specific Post-Training:}
We jointly train reconstruction and editing during SFT followed by GDPO with complementary rewards for rendered fidelity and edit application.
\item \textbf{EdiTikZ Models:}
We release 4B and 9B EdiTikZ models. EdiTikZ-9B-RL outperforms all tested baselines in automatic evaluation and surpasses GPT-5.6-Sol in human evaluation.
\end{itemize}

\section{Related Work}
\label{sec:Related Work}

\paragraph{Generating Scientific Figures with Graphics Programs}
For TikZ, prior work generates code from text~\citep{belouadi2024automatikz, belouadi2025tikzero, greisinger2026tikzilla}, or reconstructs it from images~\citep{belouadi2024detikzify, zeng2026davinci, lin2026scientificgraphicsprogramsynthesis}. Other work targets SVG~\citep{rodriguez2025cvpr-starvector,Wu2024Chat2SVGVG,zou-etal-2024-vgbench}, Python~\citep{ni-etal-2025-viscoder,yang-etal-2024-matplotagent}, multiple visualization languages~\citep{zhang2025scimage,ni2026viscoder}, or generates diagrams from documents~\citep{zhu2026paperbanana, Guan2026GENFIG1VS, mondal-etal-2024-scidoc2diagrammer}. However, these methods generate figures from scratch instead of modifying them.

\paragraph{Scientific Figure Editing}
Prior work studies editing of charts~\citep{zhao-etal-2025-chartedit,li2026charts}, SVGs~\citep{kuchar2025vectoreditsdatasetbenchmarkinstructionbased, lin2026autofigureeditgeneratingeditablescientific}, TikZ~\citep{Wei_2025_CVPR}, and rasters~\citep{zhao2026craftermultiagentharnesseditable} using agentic systems. Concurrent work includes S1-Omni-Image~\citep{li2026s1omniimageunifiedmodelscientific}, which unifies scientific-image understanding, generation, and editing, and DisciplineGen-1M~\citep{wang2026disciplinegen1mlargescaledatasetmultidisciplinary}, which constructs OCR-based synthetic editing supervision. Released during the final preparation of this manuscript, VisEditBench~\citep{rahman2026viseditbenchvisionlanguagemodelsedit} benchmarks Matplotlib/Vega-Lite code editing from multimodal feedback, while Diagram-MMU~\citep{bo2026diagrammmumultimodalbenchmarkscientific} benchmarks image-conditioned TikZ editing using template-constructed modifications across six diagram types. In contrast, we construct large-scale training supervision from plausible pairs of human-authored scientific figures and synthesize only the missing edit instruction. See Appendix~\ref{subsec:Related Work} for broader image-editing work.

\paragraph{RL from Rendering Feedback}
Rendered-feedback RL has been applied to SVG~\citep{NEURIPS2025_rlrf,zeng2026davinci,rodriguez2026vectorgym} and TikZ generation~\citep{greisinger2026tikzilla,lin2026scientificgraphicsprogramsynthesis}, using perceptual, domain-specific, code-based, and self-consistency rewards. Recent methods use VLM feedback to compare charts~\citep{Tang_2026_CVPR_MM-ReCoder} or answer instance-specific visual questions~\citep{yang-etal-2026-omnidiagram}. Scientific figure editing instead requires preserving source content while applying localized changes. We therefore combine global rendered similarity with a source-conditioned, target-reference-free VLM verifier for individual requested edits.

\section{Dataset and Benchmark}
\label{sec:Dataset and Benchmark}

\begin{figure*}[t]
\centering

\begin{tikzpicture}[
x=1cm,
y=1cm,
stage/.style={
    draw=gray!65,
    dashed,
    rounded corners=4pt,
    fill=gray!5,
    line width=0.3pt,
    inner sep=0pt
},
branch/.style={
    draw=gray!55,
    rounded corners=3pt,
    fill=white,
    line width=0.3pt,
    inner sep=0pt
},
paneltitle/.style={
    font=\bfseries\tiny,
    anchor=south west,
    inner sep=0pt
},
versionlabel/.style={
    font=\bfseries\tiny,
    anchor=south,
    inner sep=1pt
},
dots/.style={
    font=\bfseries\small,
    anchor=south,
    inner sep=1pt
},
img/.style={
    inner sep=0pt,
    outer sep=0pt,
    anchor=center
},
sourceimg/.style={
    draw=gray!45,
    line width=0.2pt,
    inner sep=0.1pt,
    outer sep=0pt,
    anchor=center
},
revision/.style={
    <->,
    >=stealth,
    line width=0.4pt
}
]

\node[
    stage,
    anchor=south west,
    minimum width=5.10cm,
    minimum height=2.65cm
] (stageone) at (0,0) {};

\node[paneltitle]
at ($(stageone.north west)+(0,0.04cm)$)
{1) Cross-Version Revisions};

\node[sourceimg] (paperone)
at ($(stageone.west)+(0.78cm,0)$) {%
    \includegraphics[
        width=2cm,
        height=2cm
    ]{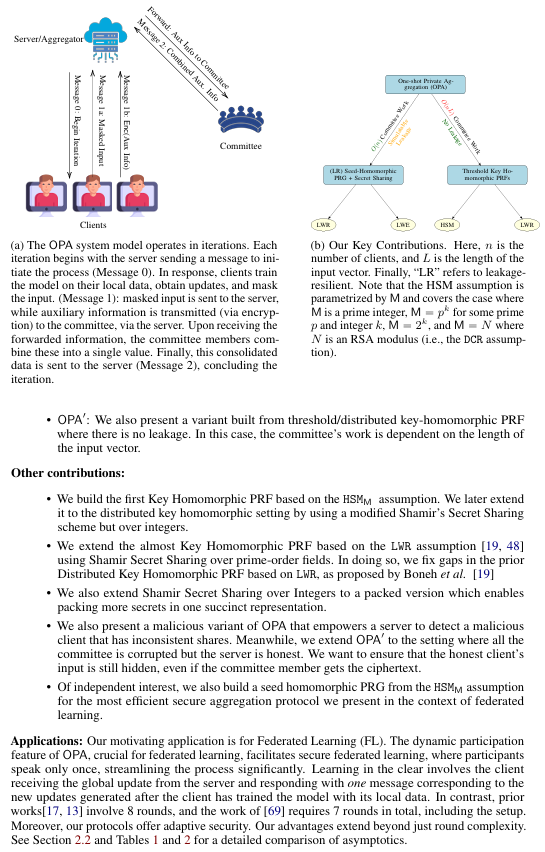}%
};

\node[sourceimg] (papertwo)
at ($(paperone.east)+(0.95cm,0)$) {%
    \includegraphics[
        width=2cm,
        height=2cm
    ]{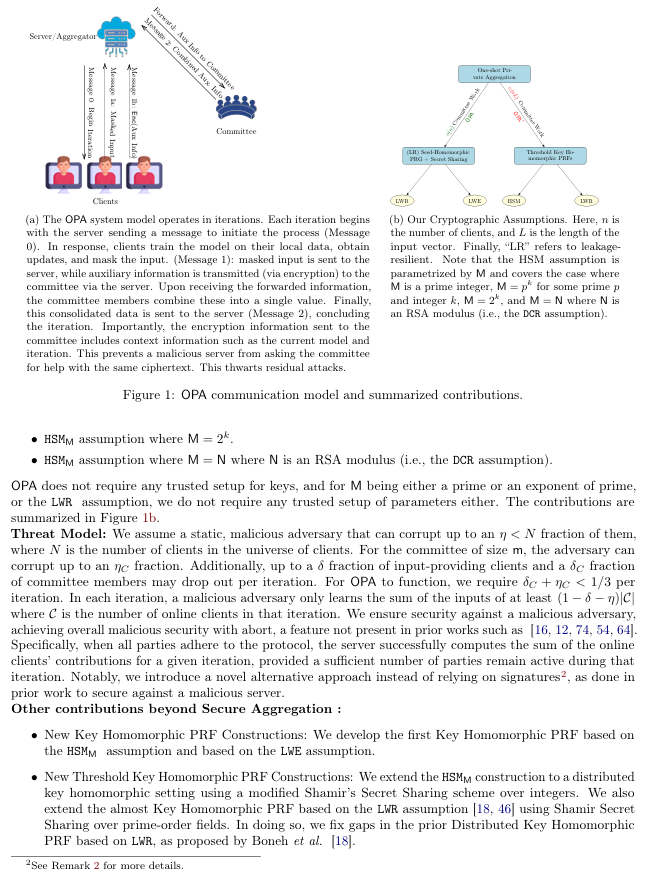}%
};

\node[versionlabel]
at ($(paperone.north)+(0,0.0cm)$)
{$V_1$};

\node[versionlabel]
at ($(papertwo.north)+(0,0.0cm)$)
{$V_N$};

\node[dots]
at ($(paperone.east)+(0.105cm,-0.18cm)$)
{$\vdots$};

\node[
    branch,
    anchor=east,
    minimum width=1.50cm,
    minimum height=2.45cm
] (pairone)
at ($(stageone.east)+(-0.10cm,0)$) {};

\node[
    img,
    draw=red!85!black,
    line width=1.1pt
] (diffone)
at ($(pairone.center)+(0,0.70cm)$) {%
    \includegraphics[
        width=1.4cm,
        height=1.4cm
    ]{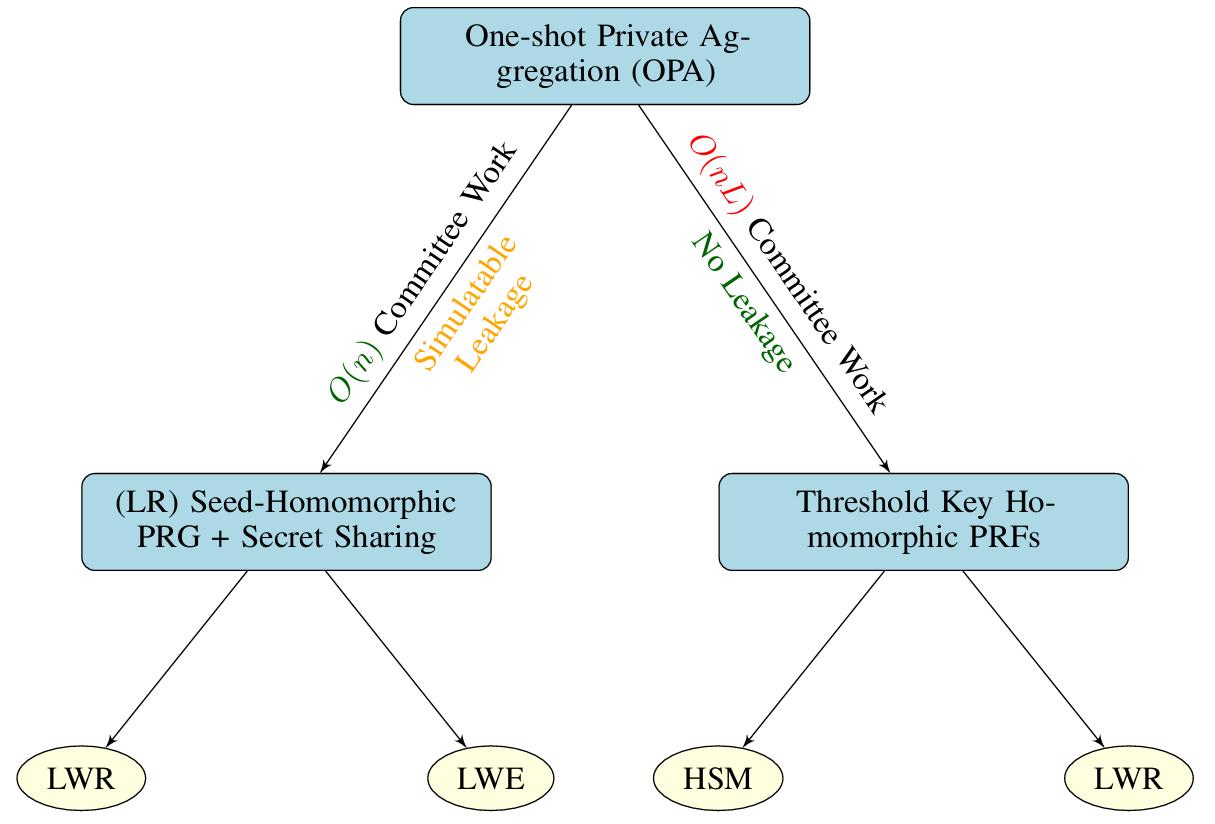}%
};

\node[
    img,
    draw=green!60!black,
    line width=1.1pt
] (difftwo)
at ($(pairone.center)+(0,-0.70cm)$) {%
    \includegraphics[
        width=1.4cm,
        height=1.4cm
    ]{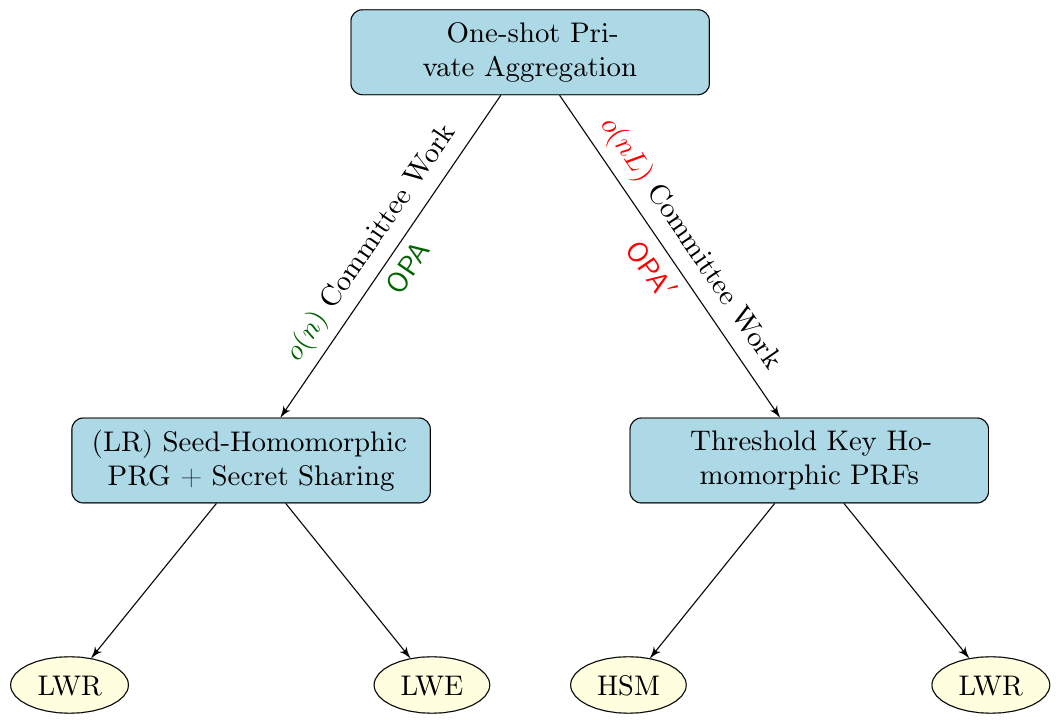}%
};

\draw[revision]
($(diffone.south)+(0,-0.03cm)$)
--
($(difftwo.north)+(0,0.03cm)$);

\coordinate (paperonefocus)
at ($(paperone.center)+(0.36cm,0.64cm)$);

\coordinate (papertwofocus)
at ($(papertwo.center)+(0.39cm,0.69cm)$);

\node[
    draw=red!85!black,
    line width=0.55pt,
    minimum width=0.55cm,
    minimum height=0.39cm,
    inner sep=0pt
] (paperonebox) at (paperonefocus) {};

\node[
    draw=green!60!black,
    line width=0.55pt,
    minimum width=0.50cm,
    minimum height=0.35cm,
    inner sep=0pt
] (papertwobox) at (papertwofocus) {};

\draw[
    red!85!black,
    ->,
    >=stealth,
    line width=0.55pt
]
(paperonebox.east)
to[out=20,in=150]
($(diffone.west)+(-0.01,0.0cm)$);

\draw[
    green!60!black,
    ->,
    >=stealth,
    line width=0.55pt
]
(papertwobox.east)
to[out=-20,in=150]
($(difftwo.west)+(-0.01,0.0cm)$);

\coordinate (redfocus)
at ($(diffone.center)+(-0.2cm,0.16cm)$);

\node[
    draw=red!85!black,
    line width=0.55pt,
    minimum width=0.27cm,
    minimum height=0.33cm,
    inner sep=0pt
] (redsource) at (redfocus) {};

\coordinate (redzoomcenter)
at ($(diffone.center)+(0.4125cm,0.121cm)$);

\begin{scope}
    \clip
        ($(redzoomcenter)+(-0.297cm,-0.363cm)$)
        rectangle
        ($(redzoomcenter)+(0.297cm,0.363cm)$);

    \node[inner sep=0pt]
    at ($(redzoomcenter)+(0.44cm,-0.352cm)$) {%
        \includegraphics[
            width=3.08cm
        ]{structure/revision_figures/tikz_diff_v1.png}%
    };
\end{scope}

\draw[
    red!85!black,
    line width=0.55pt
]
($(redzoomcenter)+(-0.297cm,-0.363cm)$)
rectangle
($(redzoomcenter)+(0.297cm,0.363cm)$);

\draw[
    red!85!black,
    line width=0.45pt
]
(redsource.north east)
--
($(redzoomcenter)+(-0.297cm,0.363cm)$);

\draw[
    red!85!black,
    line width=0.45pt
]
(redsource.south east)
--
($(redzoomcenter)+(-0.297cm,-0.363cm)$);

\coordinate (greenfocus)
at ($(difftwo.center)+(-0.2cm,0.16cm)$);

\node[
    draw=green!60!black,
    line width=0.55pt,
    minimum width=0.27cm,
    minimum height=0.34cm,
    inner sep=0pt
] (greensource) at (greenfocus) {};

\coordinate (greenzoomcenter)
at ($(difftwo.center)+(0.4125cm,0.113cm)$);

\begin{scope}
    \clip
        ($(greenzoomcenter)+(-0.297cm,-0.374cm)$)
        rectangle
        ($(greenzoomcenter)+(0.297cm,0.374cm)$);

    \node[inner sep=0pt]
    at ($(greenzoomcenter)+(0.44cm,-0.352cm)$) {%
        \includegraphics[
            width=3.08cm
        ]{structure/revision_figures/tikz_diff_v3.png}%
    };
\end{scope}

\draw[
    green!60!black,
    line width=0.55pt
]
($(greenzoomcenter)+(-0.297cm,-0.374cm)$)
rectangle
($(greenzoomcenter)+(0.297cm,0.374cm)$);

\draw[
    green!60!black,
    line width=0.45pt
]
(greensource.north east)
--
($(greenzoomcenter)+(-0.297cm,0.374cm)$);

\draw[
    green!60!black,
    line width=0.45pt
]
(greensource.south east)
--
($(greenzoomcenter)+(-0.297cm,-0.374cm)$);

\node[
    stage,
    anchor=south west,
    minimum width=4.75cm,
    minimum height=2.65cm
] (stagetwo)
at ($(stageone.south east)+(0.12cm,0)$) {};

\node[paneltitle]
at ($(stagetwo.north west)+(0,0.04cm)$)
{2) Related Figures/Subfigures};

\node[sourceimg] (subpaperone)
at ($(stagetwo.west)+(0.95cm,0.1cm)$) {%
    \includegraphics[
        width=2.2cm,
        height=2.2cm
    ]{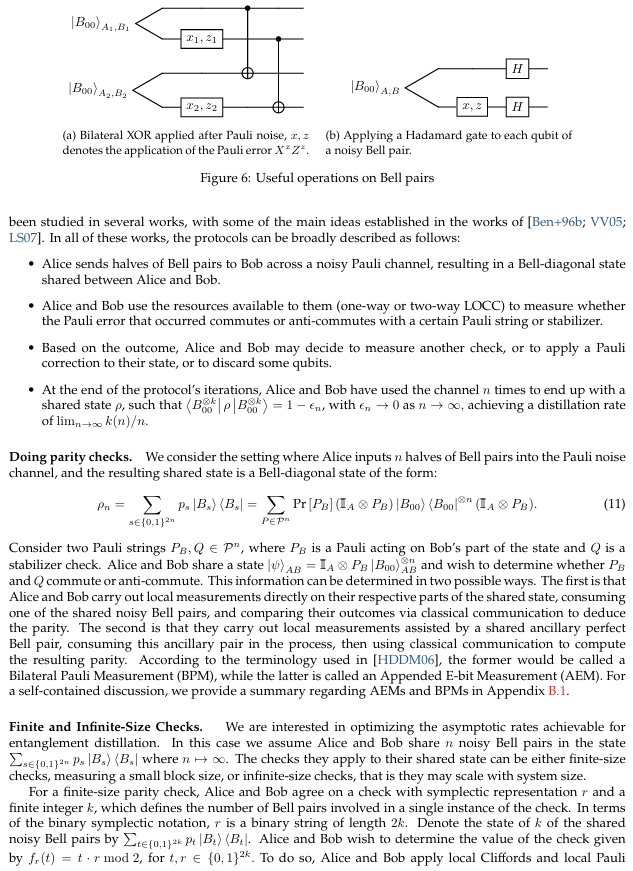}%
};

\node[sourceimg] (subpapertwo)
at ($(subpaperone.center)+(0.6cm,-0.2cm)$) {%
    \includegraphics[
        width=2.2cm,
        height=2.2cm
    ]{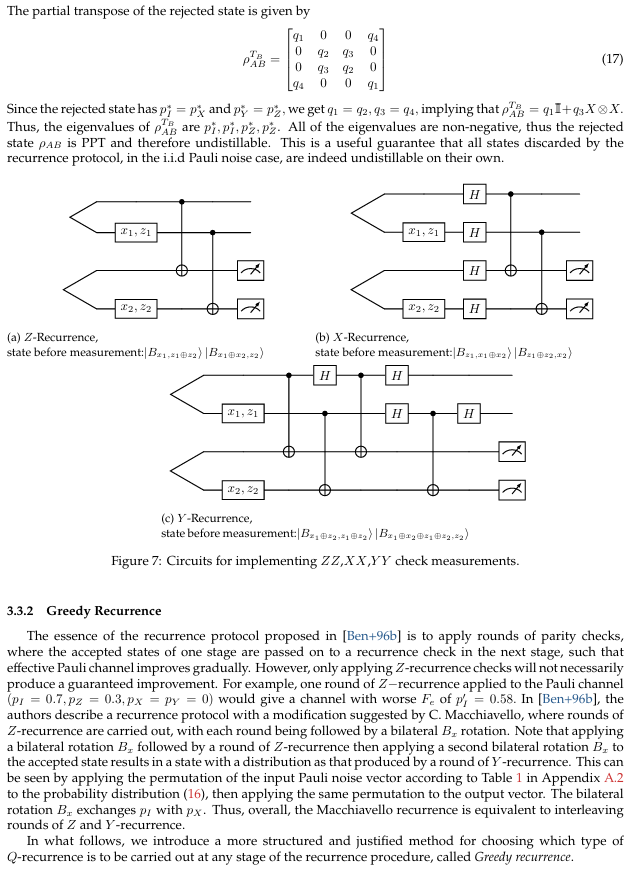}%
};

\node[
    branch,
    anchor=east,
    minimum width=1.9cm,
    minimum height=2.45cm
] (pairtwo)
at ($(stagetwo.east)+(-0.10cm,0)$) {};

\node[
    img,
    draw=red!85!black,
    line width=1.1pt
] (subone)
at ($(pairtwo.center)+(0,0.70cm)$) {%
    \includegraphics[
        width=1.7cm,
        height=1.7cm
    ]{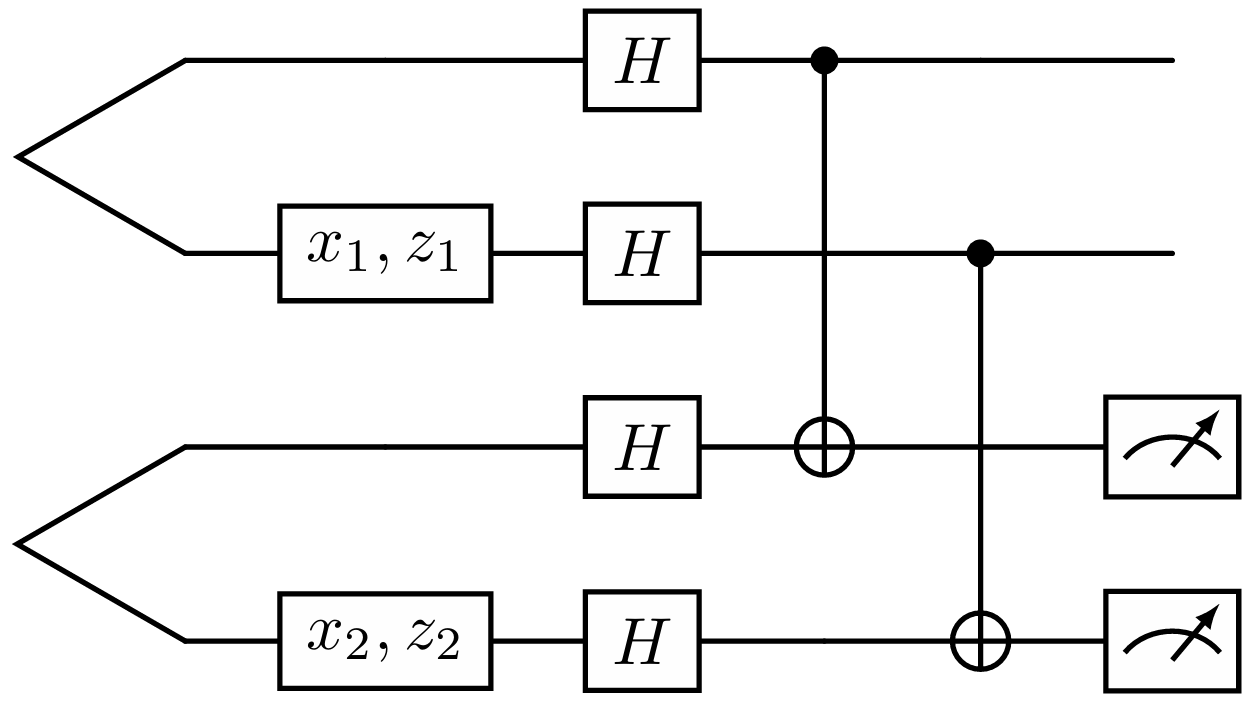}%
};

\node[
    img,
    draw=green!60!black,
    line width=1.1pt
] (subtwo)
at ($(pairtwo.center)+(0,-0.70cm)$) {%
    \includegraphics[
        width=1.7cm,
        height=1.7cm
    ]{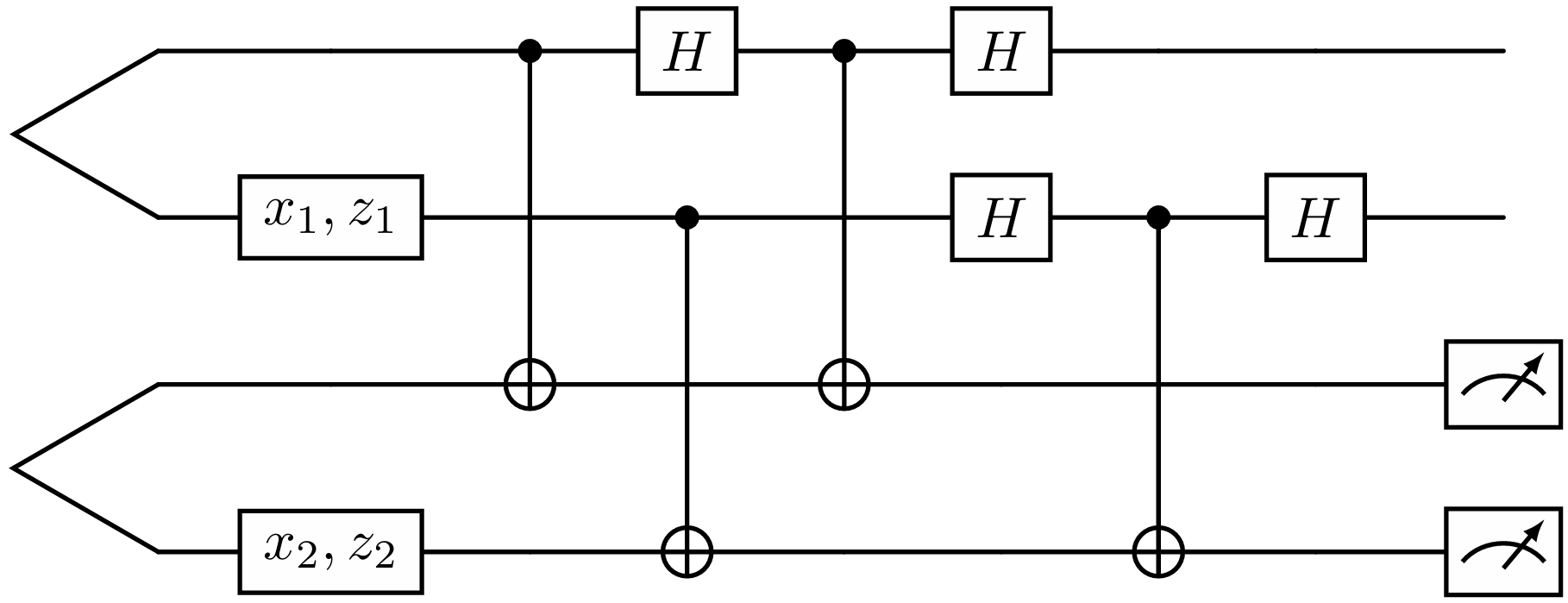}%
};

\draw[revision]
($(subone.south)+(0,-0.03cm)$)
--
($(subtwo.north)+(0,0.03cm)$);

\coordinate (subpapertwopaironefocus)
at ($(subpapertwo.center)+(0.39cm,0.465cm)$);

\coordinate (subpapertwopairtwofocus)
at ($(subpapertwo.center)+(0.08cm,0.005cm)$);

\node[
    draw=red!85!black,
    line width=0.55pt,
    minimum width=0.645cm,
    minimum height=0.37cm,
    inner sep=0pt
] (subpapertwopaironebox) at (subpapertwopaironefocus) {};

\node[
    draw=green!60!black,
    line width=0.55pt,
    minimum width=0.93cm,
    minimum height=0.37cm,
    inner sep=0pt
] (subpapertwopairtwobox) at (subpapertwopairtwofocus) {};

\draw[
    red!85!black,
    ->,
    >=stealth,
    line width=0.55pt
]
(subpapertwopaironebox.east)
to[out=0,in=180]
($(subone.west)+(-0.01,0.0cm)$);

\draw[
    green!60!black,
    ->,
    >=stealth,
    line width=0.55pt
]
(subpapertwopairtwobox.east)
to[out=0,in=180]
($(subtwo.west)+(-0.01,0.0cm)$);

\node[
    stage,
    anchor=south west,
    minimum width=3.80cm,
    minimum height=2.65cm
] (stagethree)
at ($(stagetwo.south east)+(0.12cm,0)$) {};

\node[paneltitle]
at ($(stagethree.north west)+(0,0.04cm)$)
{3) TeX SE Discussions};

\node[sourceimg] (discussionone)
at ($(stagethree.west)+(0.8cm,0.12cm)$) {%
    \includegraphics[
        width=2.2cm,
        height=2.2cm
    ]{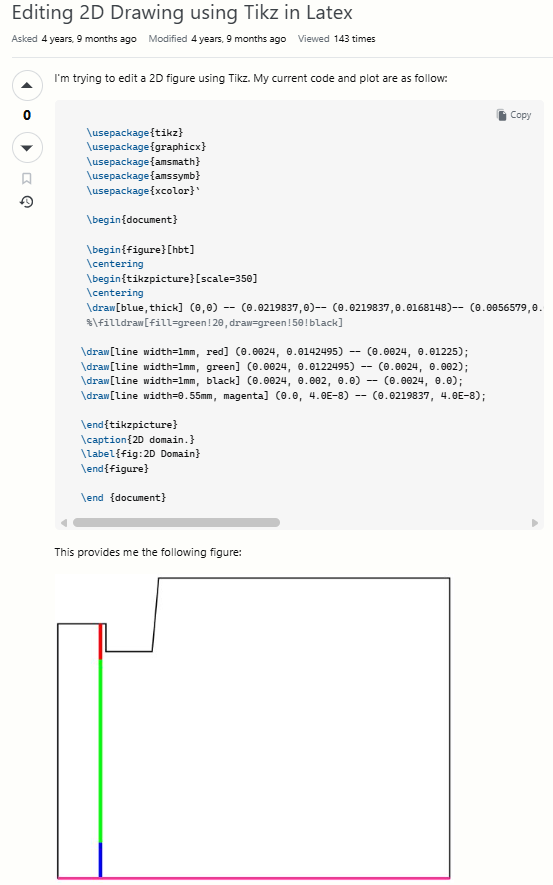}%
};

\coordinate (discussiononefocus)
at ($(discussionone.center)+(0.0cm,-0.707cm)$);

\node[
    draw=red!85!black,
    line width=0.55pt,
    minimum width=1.15cm,
    minimum height=0.8cm,
    inner sep=0pt
] (discussiononebox) at (discussiononefocus) {};

\node[sourceimg] (discussiontwo)
at ($(discussionone.center)+(0.45cm,-0.24cm)$) {%
    \includegraphics[
        width=2.2cm,
        height=2.2cm
    ]{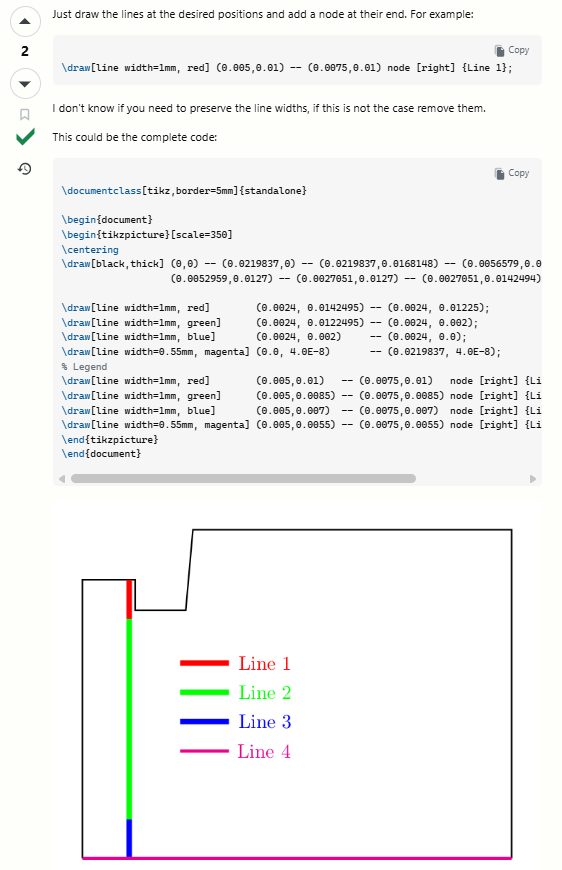}%
};

\coordinate (discussiontwofocus)
at ($(discussiontwo.center)+(0.04cm,-0.655cm)$);

\node[
    draw=green!60!black,
    line width=0.55pt,
    minimum width=1.15cm,
    minimum height=0.88cm,
    inner sep=0pt
] (discussiontwobox) at (discussiontwofocus) {};

\node[
    branch,
    anchor=east,
    minimum width=1.34cm,
    minimum height=2.45cm
] (pairthree)
at ($(stagethree.east)+(-0.10cm,0)$) {};

\node[
    img,
    draw=red!85!black,
    line width=1.10pt
] (texone)
at ($(pairthree.center)+(0,0.70cm)$) {%
    \includegraphics[
        width=1.2cm,
        height=1.2cm
    ]{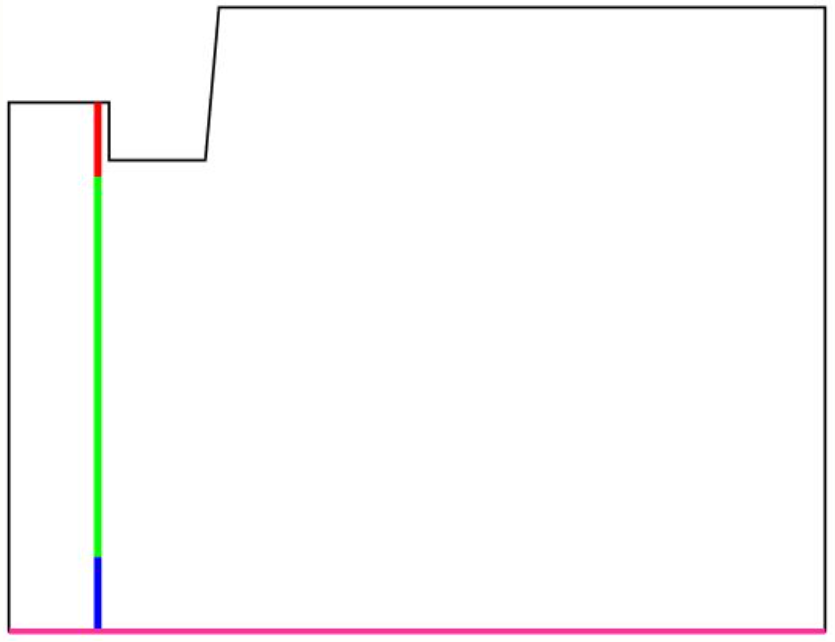}%
};

\node[
    img,
    draw=green!60!black,
    line width=1.10pt
] (textwo)
at ($(pairthree.center)+(0,-0.70cm)$) {%
    \includegraphics[
        width=1.2cm,
        height=1.2cm
    ]{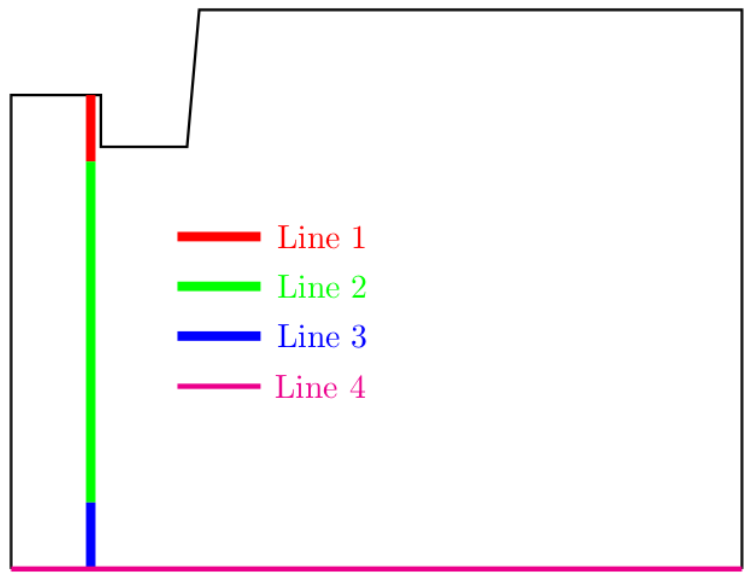}%
};

\draw[revision]
($(texone.south)+(0,-0.03cm)$)
--
($(textwo.north)+(0,0.03cm)$);

\draw[
    red!85!black,
    ->,
    >=stealth,
    line width=0.55pt
]
($(discussiononebox.north)+(-0.40cm,0)$)
to[out=90,in=180]
($(texone.west)+(-0.01cm,0.0cm)$);

\draw[
    green!60!black,
    ->,
    >=stealth,
    line width=0.55pt
]
(discussiontwobox.east)
to[out=0,in=180]
($(textwo.west)+(-0.01,0.0cm)$);

\end{tikzpicture}

\caption{Sources of figure editing supervision. We recover plausible edit pairs from cross-version revisions (left), related figures and subfigures within shared scientific contexts (middle), and iterative refinements in TeX SE discussions (right). Red and green highlight paired variants in their sources.}
\label{fig:revision_sources}

\end{figure*}

\paragraph{Revision-Derived Editing Supervision}
Our key observation is that plausible scientific figure edits naturally arise throughout scientific revision and development processes, including (i) figures modified across arXiv or GitHub versions, (ii) related (sub-)figures in the same paper or repository, (iii) alternative TikZ programs retained in source files but not rendered in the document, and (iv) iterative refinements in TeX SE discussions (Figure~\ref{fig:revision_sources}). Exact figure lineage is difficult to recover reliably as figures may be added, removed, renamed, reordered, or moved across files, while surrounding anchors such as captions, references, and related text can also change. We therefore identify semantically similar pairs within shared scientific contexts and retain plausible editing transformations.

\setkeys{Gin}{keepaspectratio}

\paragraph{Collecting Scientific Revision Traces}
We extend DaTikZ-V4~\citep{greisinger2026tikzilla} by recovering TikZ from all historical versions of arXiv submissions containing \texttt{tikzpicture}, \texttt{circuitikz}, or \texttt{tikzcd}. We apply the TikZilla preprocessing pipeline, including document expansion, subfigure extraction, code standardization, dynamic package inclusion, filtering, rendering, and deduplication on the standardized TikZ body. Across 91K arXiv submissions, 38K contain at least two versions with modified TikZ code. Historical versions contribute 0.77M additional figures, increasing the unique arXiv corpus from 1.47M to 2.38M. Combined with GitHub and TeX SE, this yields a candidate corpus of 2.91M unique TikZ figures.

\paragraph{Recovering Plausible Edit Pairs}
We group figures by arXiv submission across versions, GitHub repository, and TeX SE discussion thread, yielding 222K groups, of which 123K contain at least two unique figures. We prune groups above the 90th size percentile and compute within-group cosine similarities using \texttt{DeTikZify-V2}'s image encoder. To determine the filtering threshold, we manually evaluate 50 pairs in each of eight similarity intervals (0.92--0.9999, width 0.01) and retain intervals containing fewer than 15\% implausible transformations (Table~\ref{tab:semantic_similarity_examples}). This produces 430,442 candidate pairs from 87,051 contributing groups, connecting 589,986 unique figures.

\newlength{\simblockwidth}
\setlength{\simblockwidth}{0.238\textwidth}

\newlength{\simimagewidth}
\setlength{\simimagewidth}{0.108\textwidth}

\newlength{\simimageheight}
\setlength{\simimageheight}{1.65cm}

\newcommand{\simimage}[1]{%
  \begin{minipage}[c][\simimageheight][c]{\simimagewidth}
    \centering
    \includegraphics[
      width=\linewidth,
      height=0.94\simimageheight,
      keepaspectratio
    ]{#1}%
  \end{minipage}%
}

\newcommand{\simrange}[1]{%
  \fontsize{8}{9}\selectfont
  \textbf{#1}%
}

\newcommand{\simsubheader}[1]{%
  \fontsize{7}{8}\selectfont
  \textbf{#1}%
}

\newcommand{\similarityblock}[5]{%
  \begin{tikzpicture}[
    remember picture,
    baseline=(#1.base)
  ]
    \node[
      name=#1,
      anchor=center,
      inner xsep=2pt,
      inner ysep=3pt,
      outer sep=0pt,
      fill=#2,
      minimum width=\simblockwidth
    ]{%
      \begin{minipage}{\dimexpr\simblockwidth-4pt\relax}
        \centering

        \simrange{#3}

        \vspace{-5.5pt}
        \rule{\linewidth}{0.4pt}
        \vspace{-9pt}

        \begin{tabular}{@{}cc@{}}
          \simimage{#4} &
          \simimage{#5}
        \end{tabular}

        \vspace{-2pt}
      \end{minipage}%
    };
  \end{tikzpicture}%
}

\begin{table*}[t]
  \centering
  \caption{Examples of scientific-figure edit pairs across semantic-similarity intervals and the percentage of implausible editing transformations in each interval. Gray cells denote excluded intervals.}
  \label{tab:semantic_similarity_examples}

  \setlength{\tabcolsep}{2pt}
  \renewcommand{\arraystretch}{1}

  \begin{tabular}{@{}cccc@{}}
    \toprule
    \similarityblock
      {simtopone}
      {black!10}
      {$\boldsymbol{0.92\text{--}0.93}$ $\boldsymbol{(28\%)}$}
      {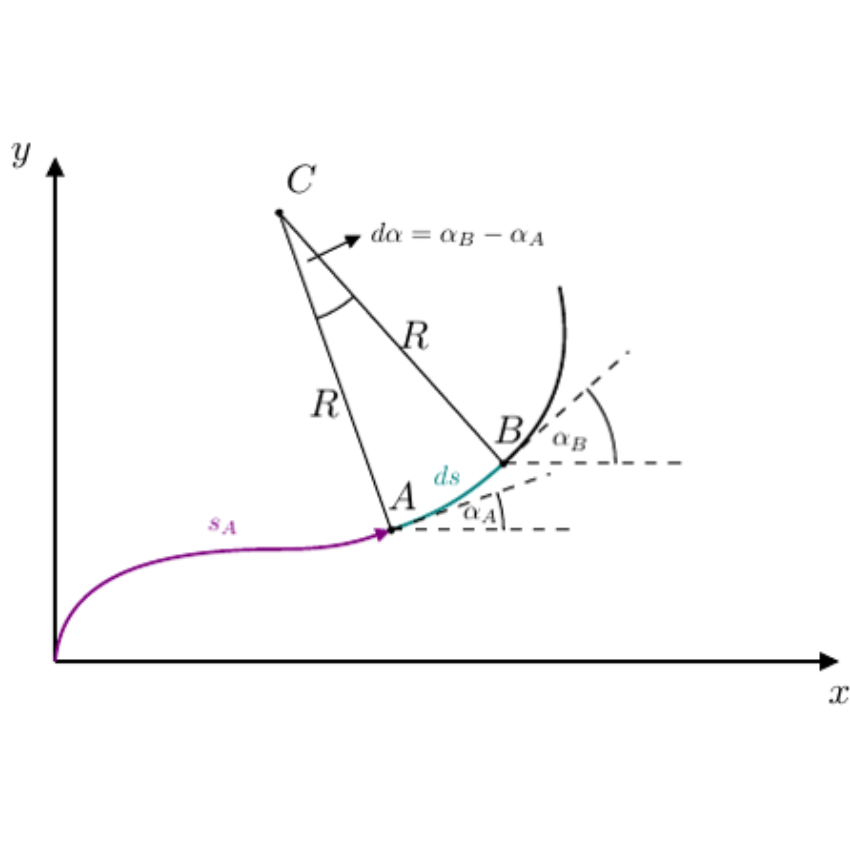}
      {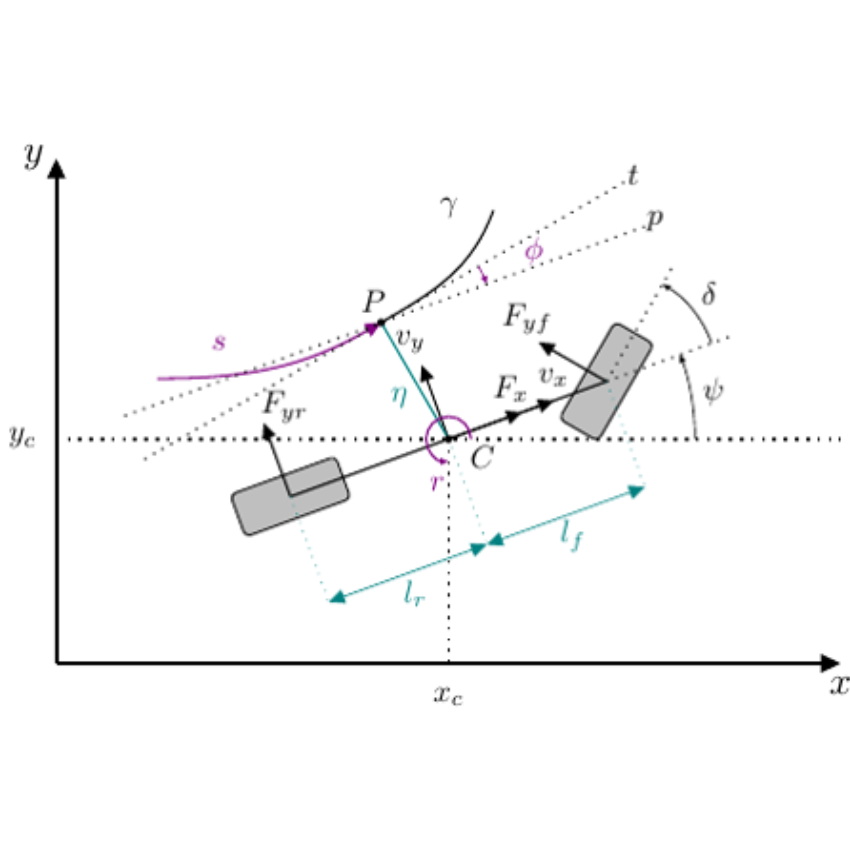}
    &
    \similarityblock
      {simtoptwo}
      {black!10}
      {$\boldsymbol{0.94\text{--}0.95}$ $\boldsymbol{(18\%)}$}
      {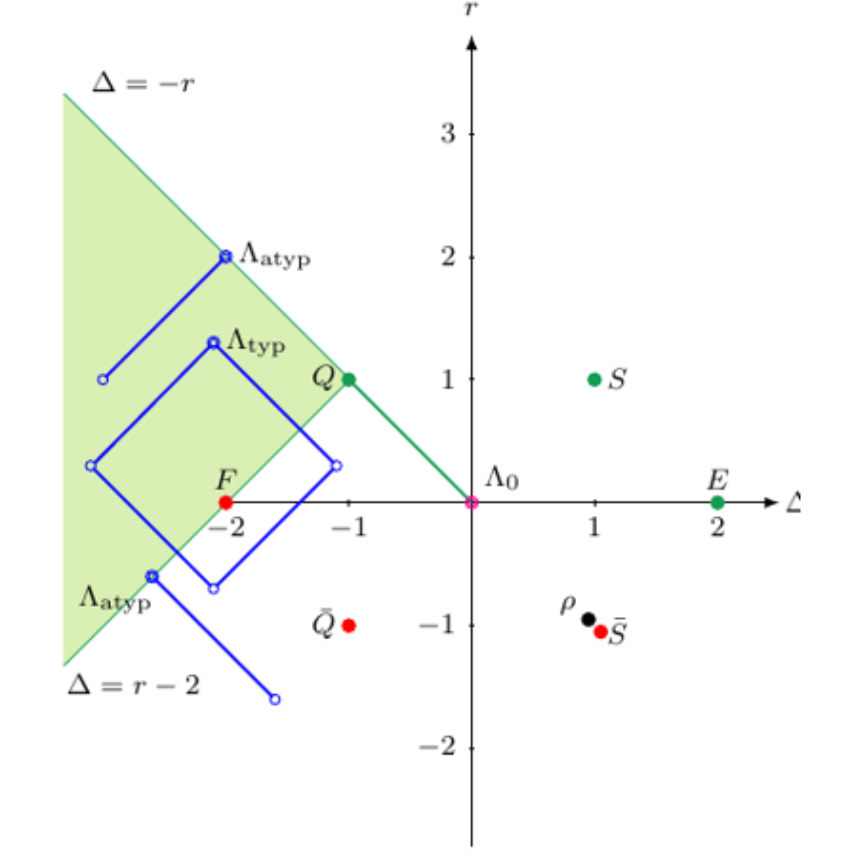}
      {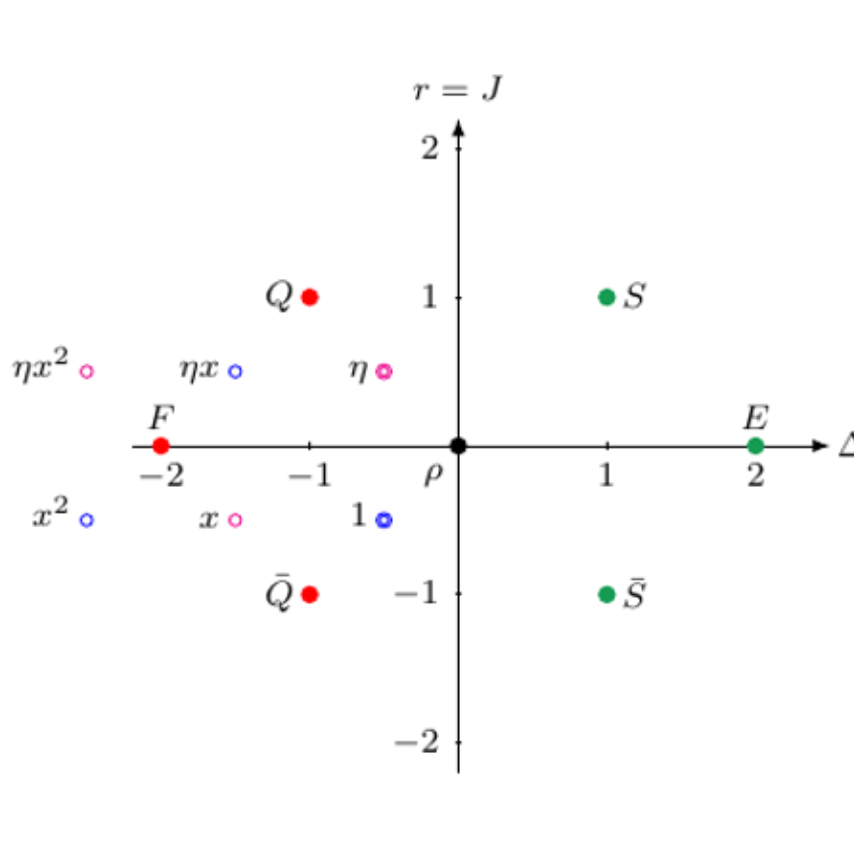}
    &
    \similarityblock
      {simtopthree}
      {white}
      {$\boldsymbol{0.96\text{--}0.97}$ $\boldsymbol{(8\%)}$}
      {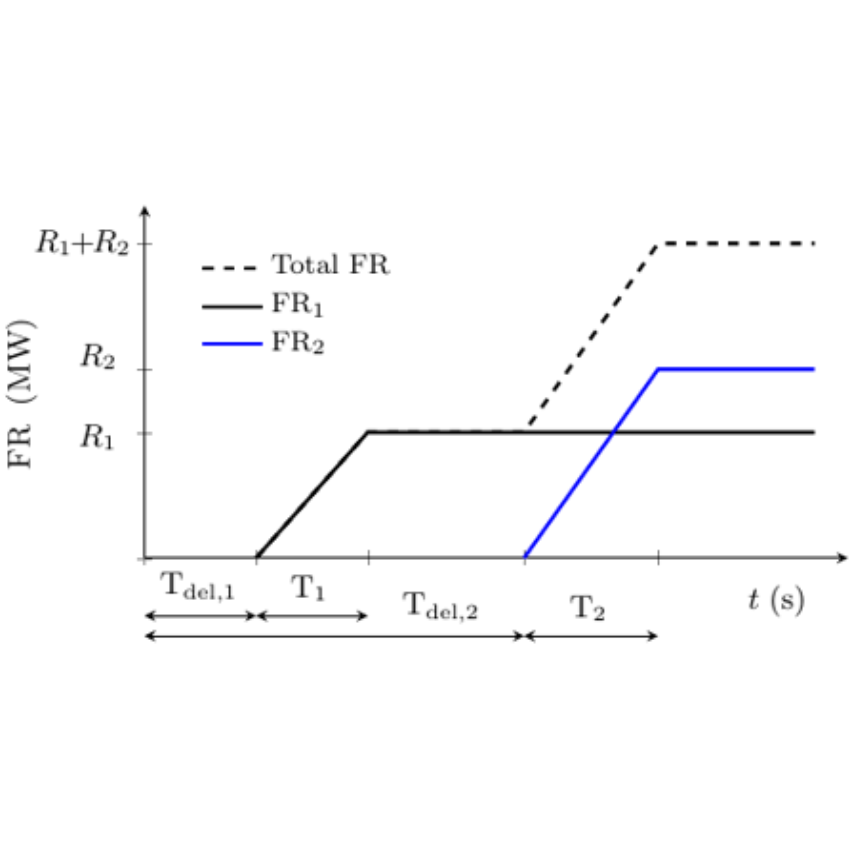}
      {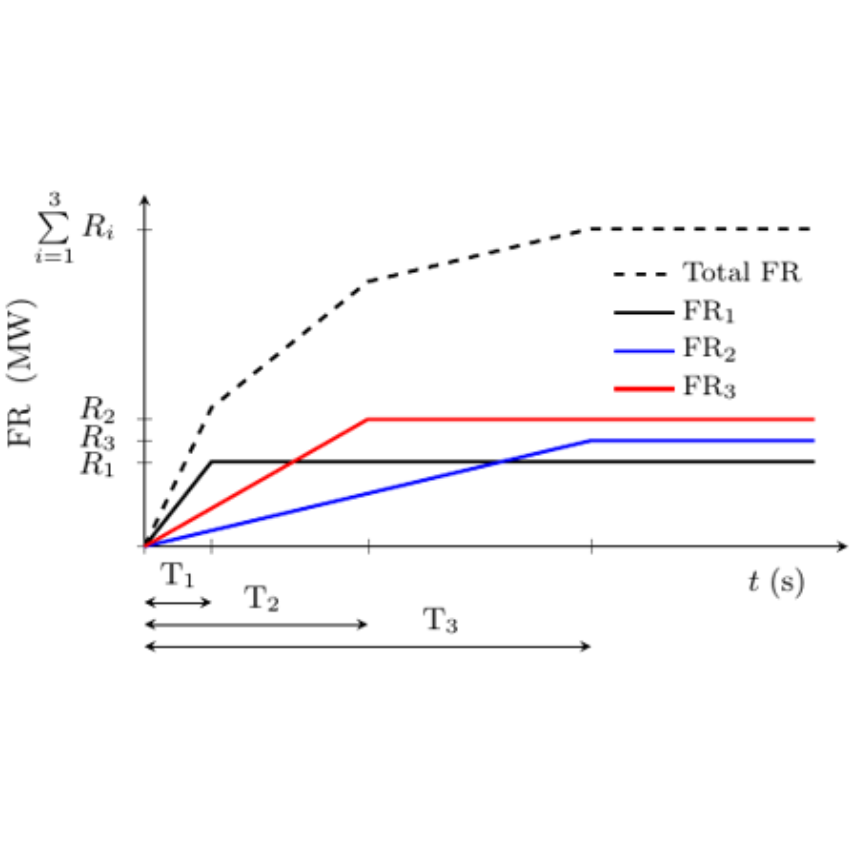}
    &
    \similarityblock
      {simtopfour}
      {white}
      {$\boldsymbol{0.98\text{--}0.99}$ $\boldsymbol{(2\%)}$}
      {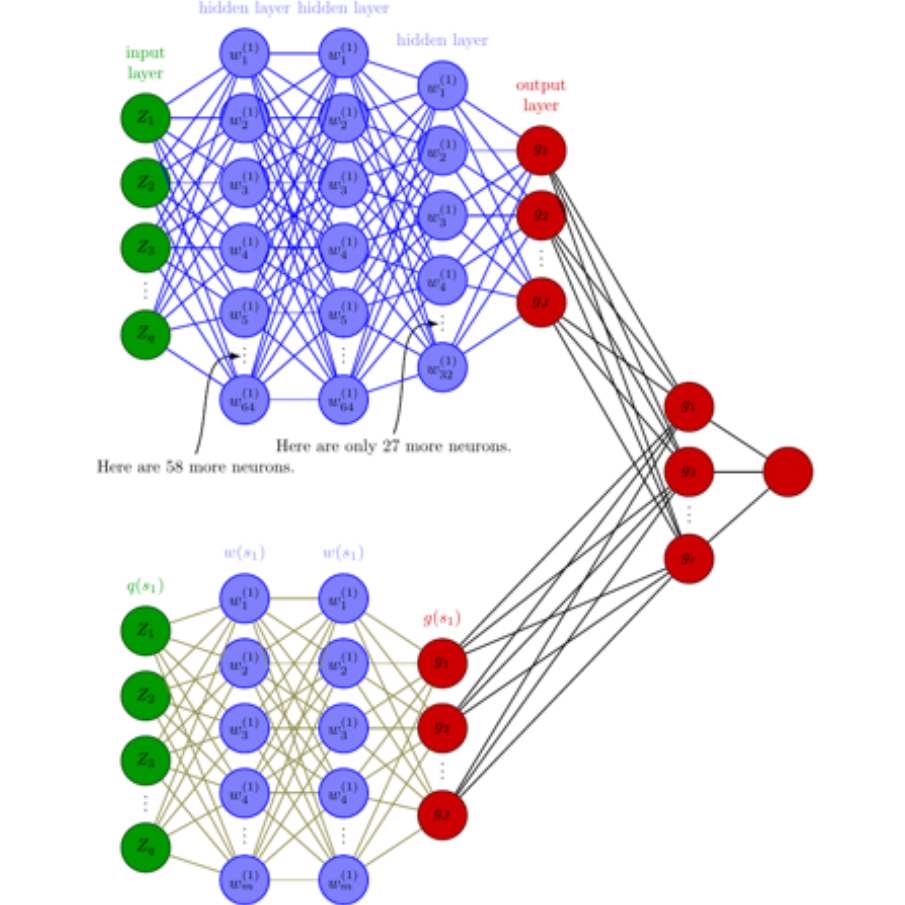}
      {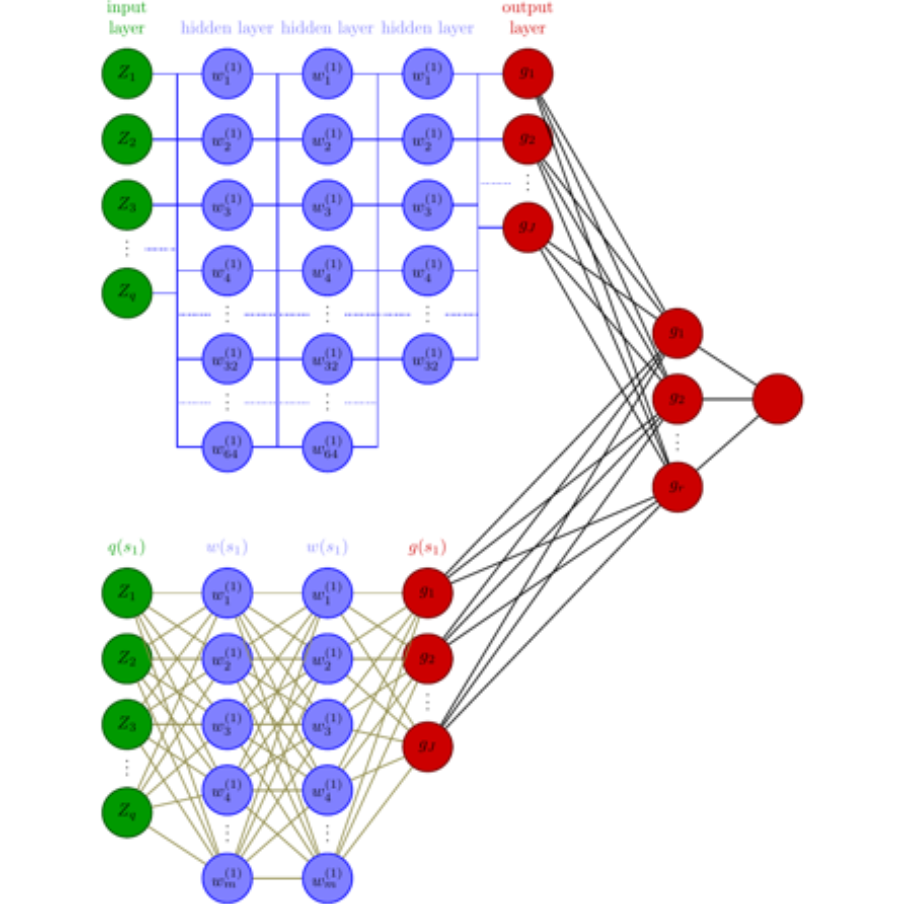}
    \\[4pt]
    \midrule
    \similarityblock
      {simbottomone}
      {black!10}
      {$\boldsymbol{0.93\text{--}0.94}$ $\boldsymbol{(22\%)}$}
      {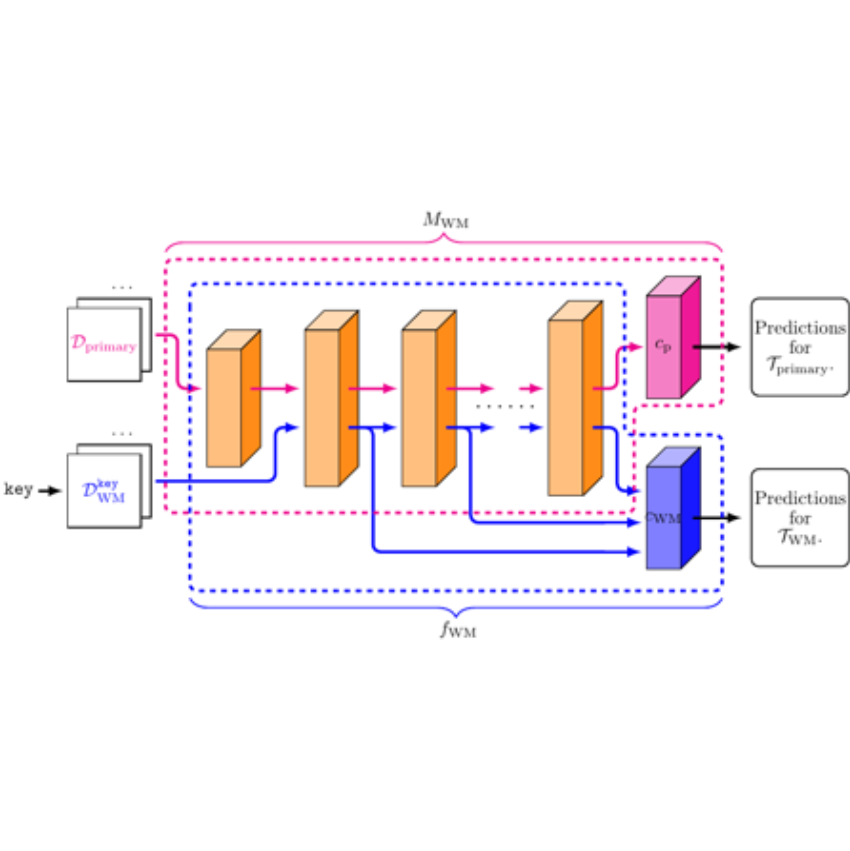}
      {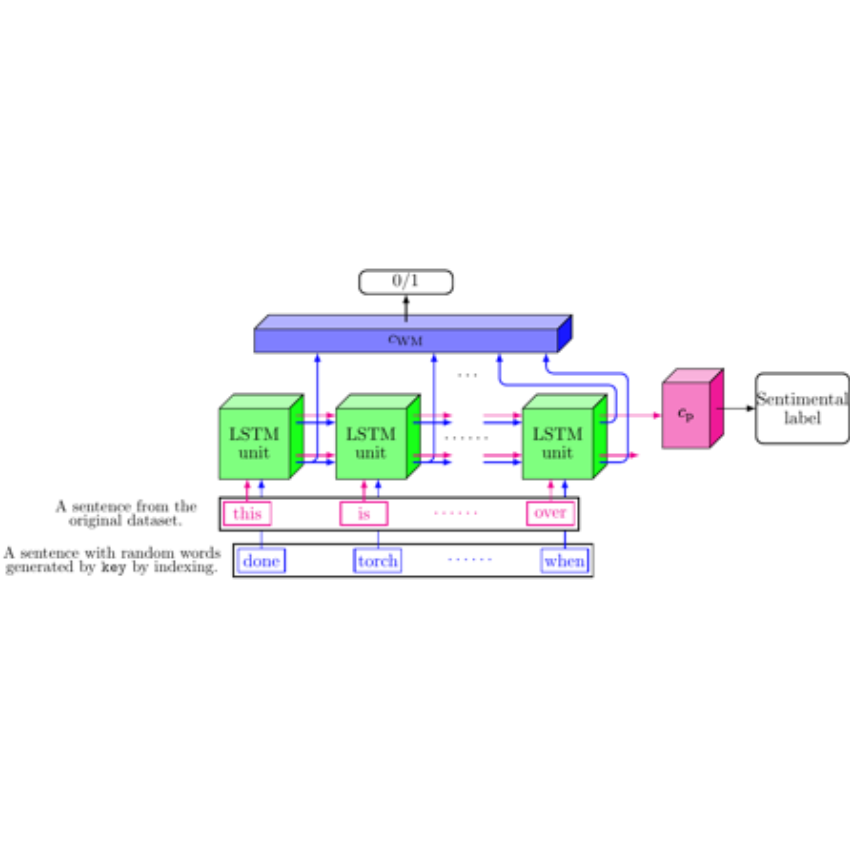}
    &
    \similarityblock
      {simbottomtwo}
      {white}
      {$\boldsymbol{0.95\text{--}0.96}$ $\boldsymbol{(10\%)}$}
      {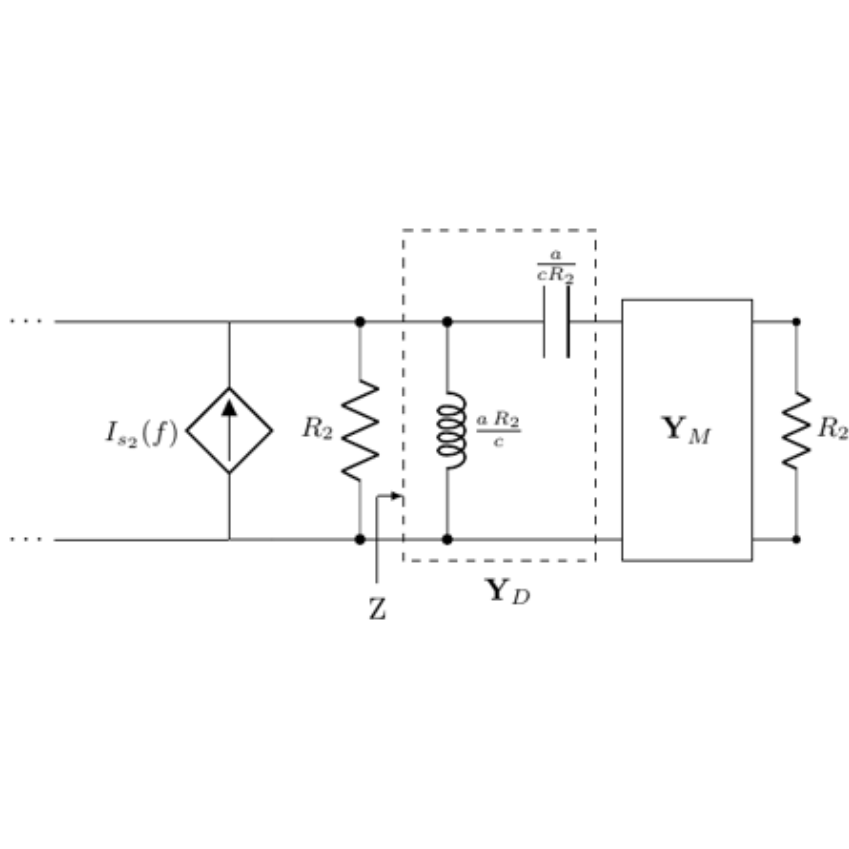}
      {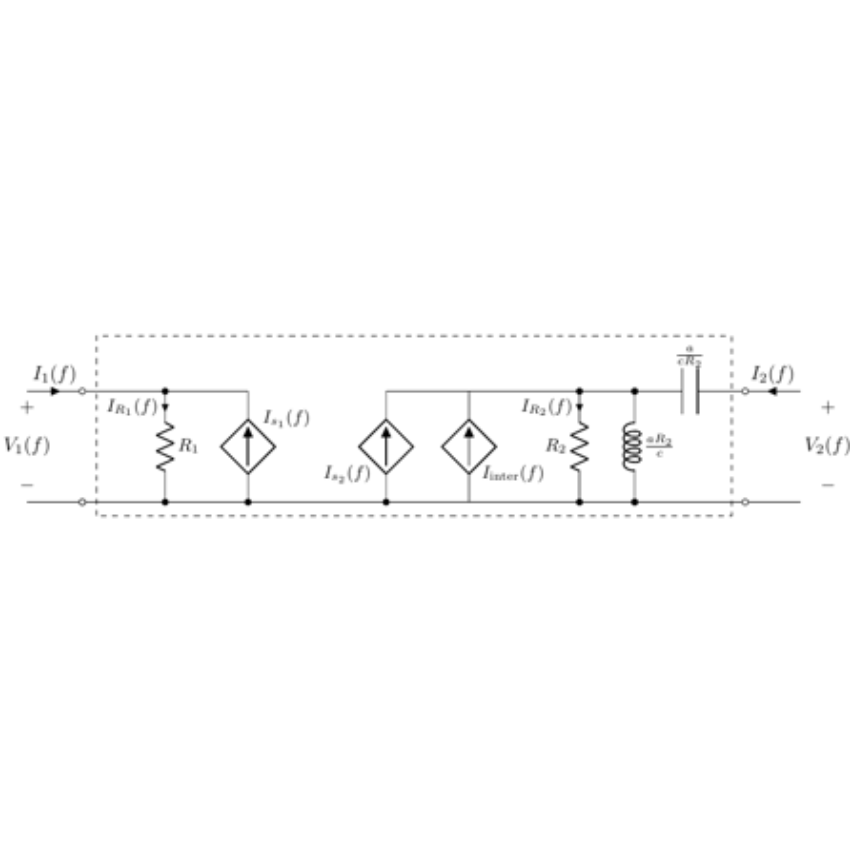}
    &
    \similarityblock
      {simbottomthree}
      {white}
      {$\boldsymbol{0.97\text{--}0.98}$ $\boldsymbol{(6\%)}$}
      {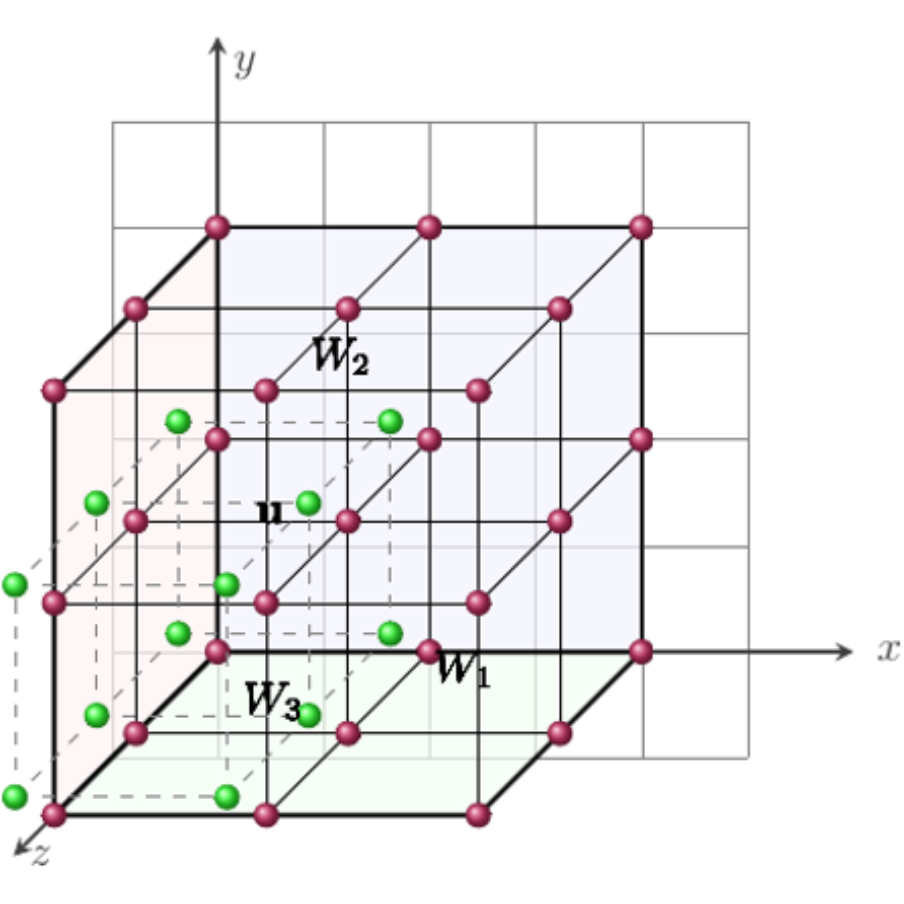}
      {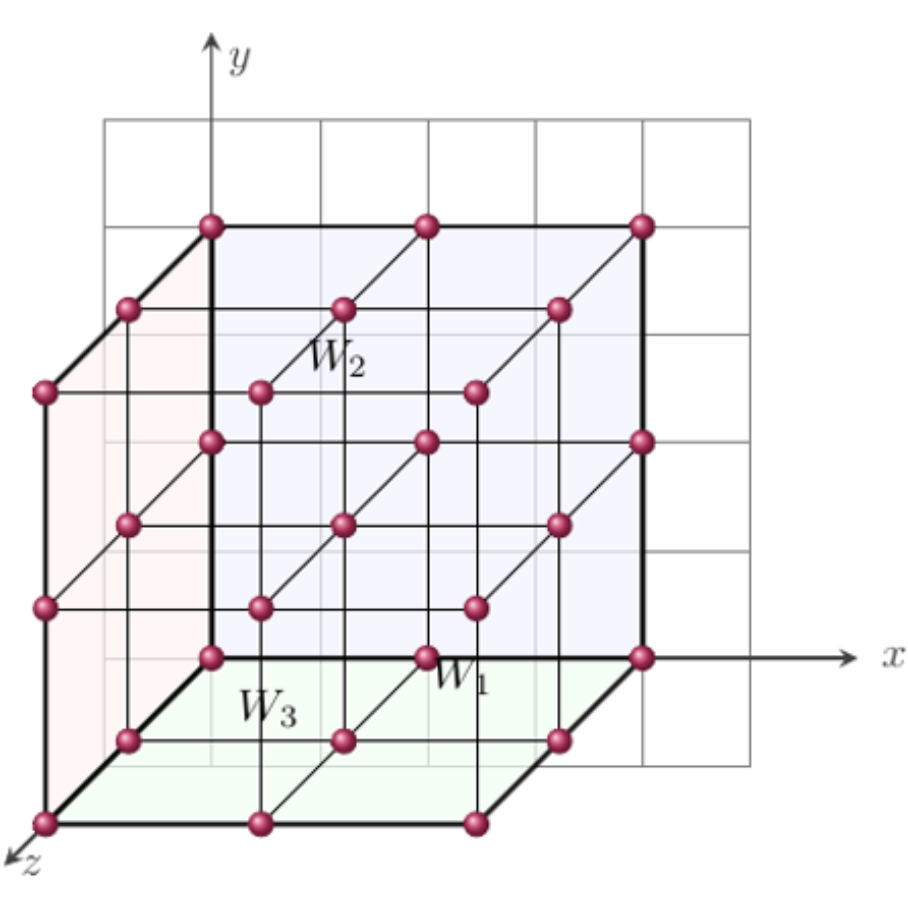}
    &
    \similarityblock
      {simbottomfour}
      {white}
      {$\boldsymbol{0.99\text{--}0.9999}$ $\boldsymbol{(12\%)}$}
      {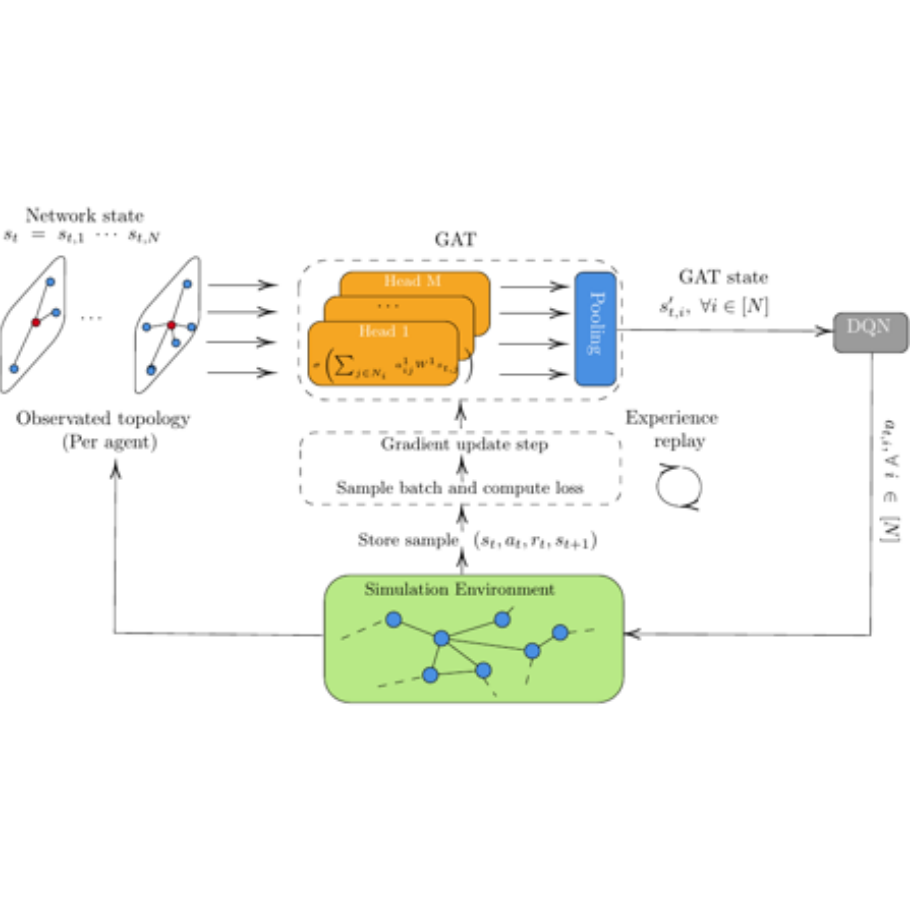}
      {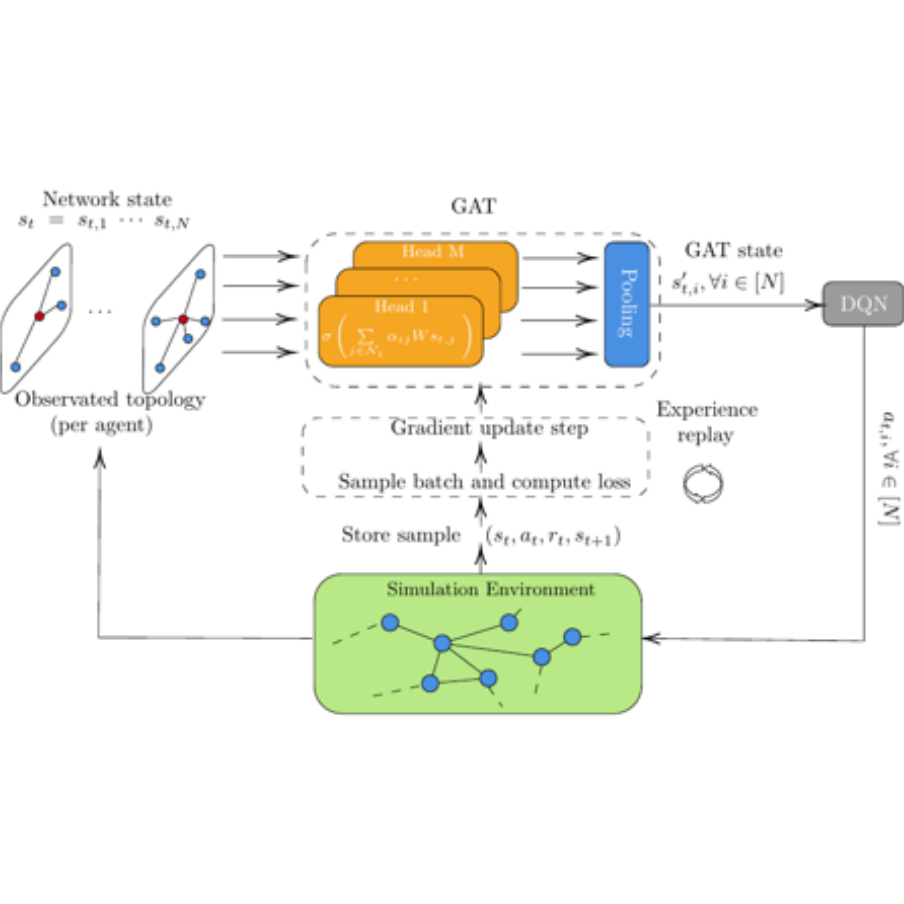} 
    \\[4pt]
    \bottomrule
  \end{tabular}
\end{table*}

\paragraph{Inferring Edit Instructions}
Because both endpoint figures are human-authored, we synthesize only the missing edit instruction using Qwen3.6-27B conditioned jointly on their renders and TikZ code. For each of the 430,442 candidate pairs, we infer both directions ($A\!\rightarrow\!B$ and $B\!\rightarrow\!A$), producing 860,884 candidate directional trajectories. The VLM classifies each direction as \texttt{ok}, \texttt{invalid}, or \texttt{identical}. For accepted transformations, it decomposes the transformation into atomic edits with an intent (\texttt{add}, \texttt{remove}, or \texttt{modify}), operation (\texttt{text}, \texttt{annotation}, \texttt{geometry}, \texttt{data}, \texttt{style}, or \texttt{structure}), and natural-language description. Requiring both directions to be accepted yields DaEdiTikZ with 390,516 figure pairs and 781,032 directional editing trajectories. Each trajectory contains 4.2 atomic edits on average, with descriptions averaging 22.3 words per atomic edit. Detailed analysis of DaEdiTikZ is in the Appendix~\ref{subsec:Dataset and Benchmark}.

\newcommand{\smathn}[1]{\scalebox{0.45}{$#1$}}

\begin{figure*}[t]
\centering

\begin{adjustbox}{max width=\textwidth}
\begin{tikzpicture}[
  ampersand replacement=\&,
  >=Latex,
  node distance=0.45cm and 0.55cm,
  pipelinearrow/.style={
    -{Latex[length=2.5mm,width=1.8mm]},
    line width=0.9pt,
    draw=black
  },
  newlabel/.style={
    font=\sffamily\bfseries\LARGE,
    text=red,
    anchor=south,
    inner sep=1pt
  },
  arxivlabel/.style={
    font=\tiny,
    text=red,
    anchor=south,
    inner sep=1pt
  },
  pipelinethinarrow/.style={
    -{Latex[length=2.1mm,width=1.5mm]},
    line width=0.7pt,
    draw=black
  },
  process/.style={
    draw=black!75,
    line width=0.7pt,
    rounded corners=1pt,
    fill=orange!7,
    inner sep=5pt,
    align=left
  },
  logo/.style={
    draw=black!70,
    rectangle,
    rounded corners=3pt,
    fill=blue!4,
    minimum width=0.5cm,
    minimum height=0.8cm,
    inner sep=2pt,
    align=center
  },
  imagebox/.style={
    draw=black!65,
    line width=0.5pt,
    fill=white,
    inner sep=1.5pt
  },
  tablebox/.style={
    draw=none,
    fill=white,
    inner sep=0pt,
    outer sep=0pt,
    minimum width=0pt,
    minimum height=0pt,
    align=center
},
  pairbox/.style={
    draw=black!75,
    line width=0.7pt,
    fill=white,
    inner sep=1.5pt,
    outer sep=0pt,
    align=center
},
  llm/.style={
    draw=black!75,
    line width=0.7pt,
    fill=orange!7,
    inner sep=5pt,
    align=left
},
  instruction/.style={
    draw=black!75,
    line width=0.7pt,
    fill=orange!4,
    inner sep=3pt,
    align=left
},
  imageencoder/.style={
    draw=black!80,
    line width=0.8pt,
    trapezium,
    trapezium left angle=70,
    trapezium right angle=70,
    shape border rotate=270,
    minimum width=1.1cm,
    minimum height=0.8cm,
    fill=blue!10,
    align=center
}
]

\node[logo] (arxiv) at (0,0) {%
    \includegraphics[width=1.00cm]{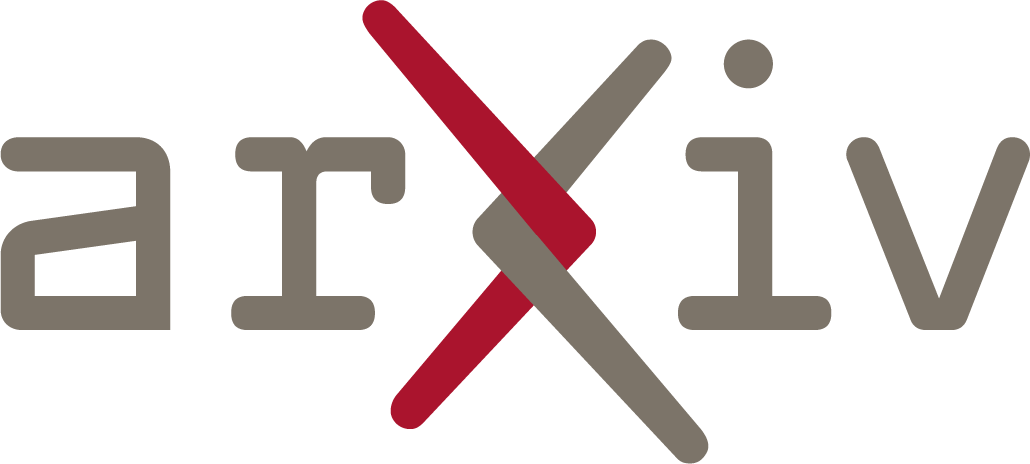}
};

\node[logo, right=0.055cm of arxiv] (github) {%
    \includegraphics[width=0.6cm]{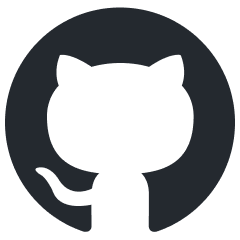}
};

\node[logo, right=0.055cm of github] (tex) {%
    \includegraphics[width=1.00cm]{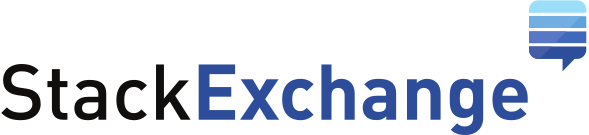}
};

\node[
    process,
    below=0.45cm of github,
    text width=2.66cm,
    align=left
] (filter) {
    \small
    \textbf{Data Preprocessing}\\
    \vspace{0.2em}
    \scriptsize
    \textbullet\ TikZilla Filtering \\
    \textbullet\ Rendering \\
    \textbullet\ Grouping \\
    \vspace{-0.55em}
    \textbullet\ Pruning
};

\node[
    anchor=north east,
    inner sep=0pt
] at ($(filter.north east)+(0.0cm,-1.3cm)$) {%
    \includegraphics[width=0.5cm]{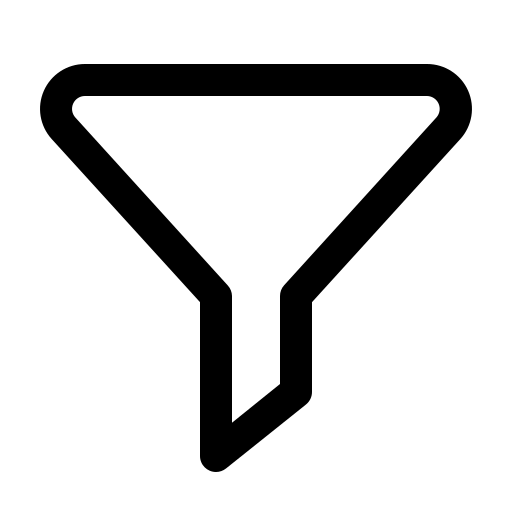}
};

\draw[pipelinethinarrow]
    (arxiv.south) -- ++(0,-0.20)
    -| ($(filter.north west)+(0.42cm,0)$);

\draw[pipelinethinarrow]
    (github.south) -- ($(filter.north)+(0,0)$);

\draw[pipelinethinarrow]
    (tex.south) -- ++(0,-0.20)
    -| ($(filter.north east)+(-0.42cm,0)$);

\node[imagebox, anchor=north west] (pathone)
    at (3.5,0.13) {%
        \includegraphics[
            width=1.5cm,
            height=0.5cm,
            keepaspectratio
        ]{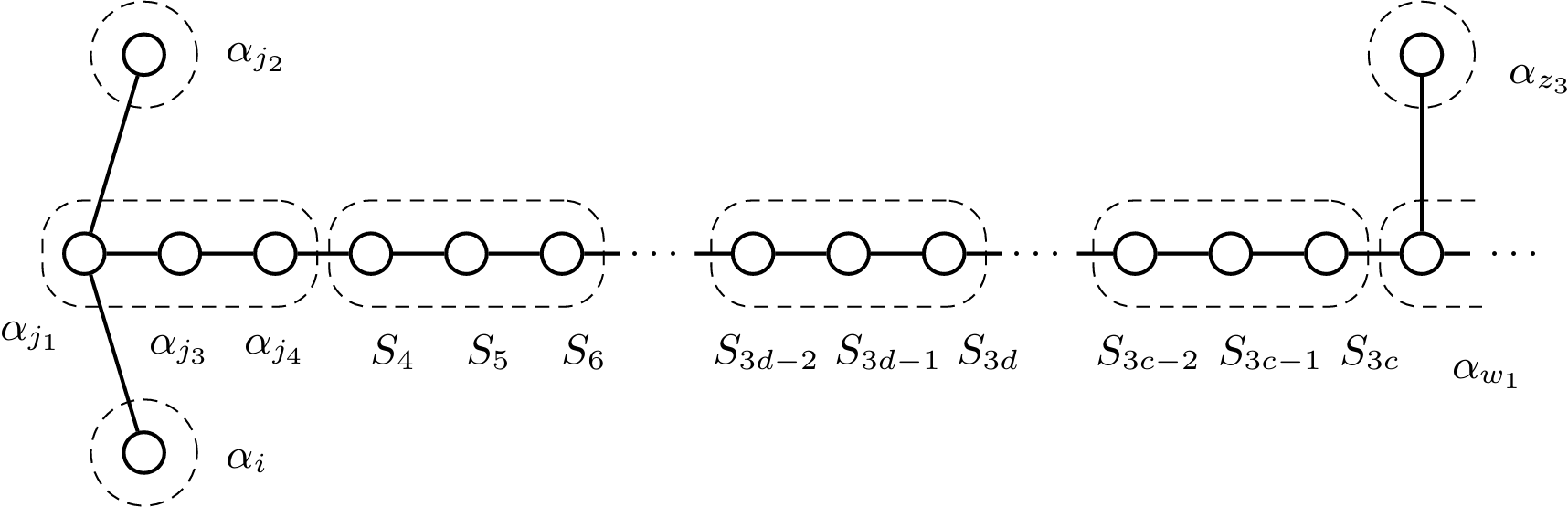}
    };

\node[imagebox, anchor=north west] (pathtwo)
    at (4.2,-0.5) {%
        \includegraphics[
            width=1.5cm,
            height=0.5cm,
            keepaspectratio
        ]{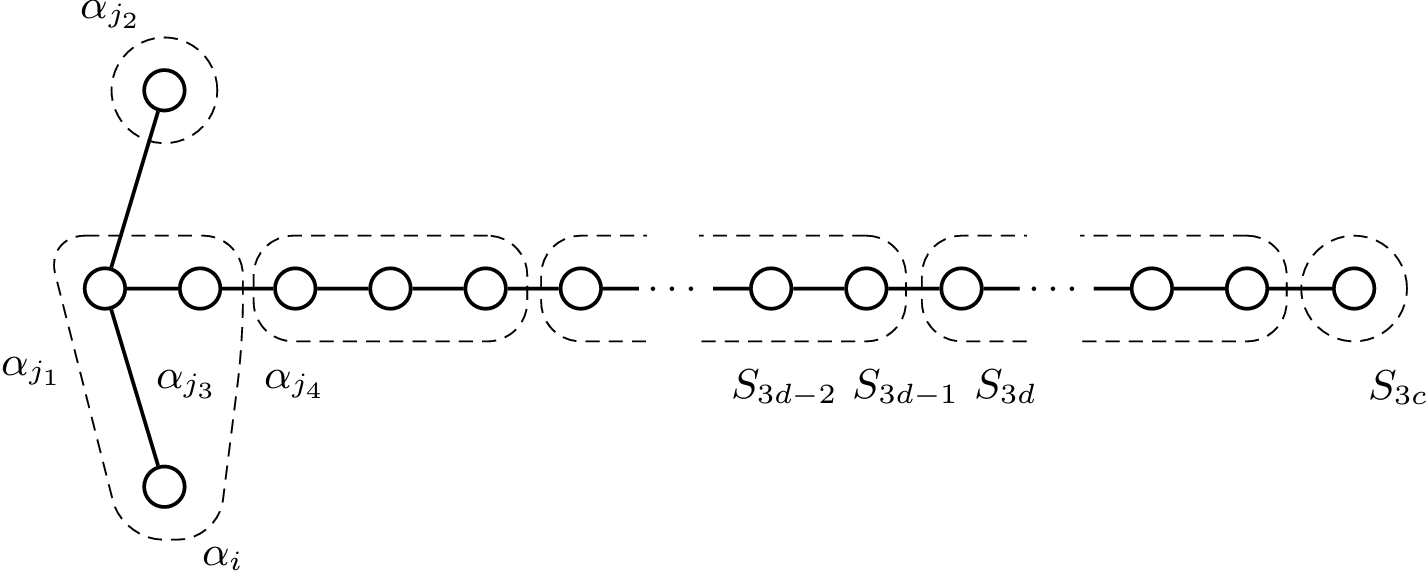}
    };

\node[imagebox, anchor=north west] (heatmap)
    at (4.2,-1.15) {%
        \includegraphics[
            width=0.95cm,
            height=0.95cm,
            keepaspectratio
        ]{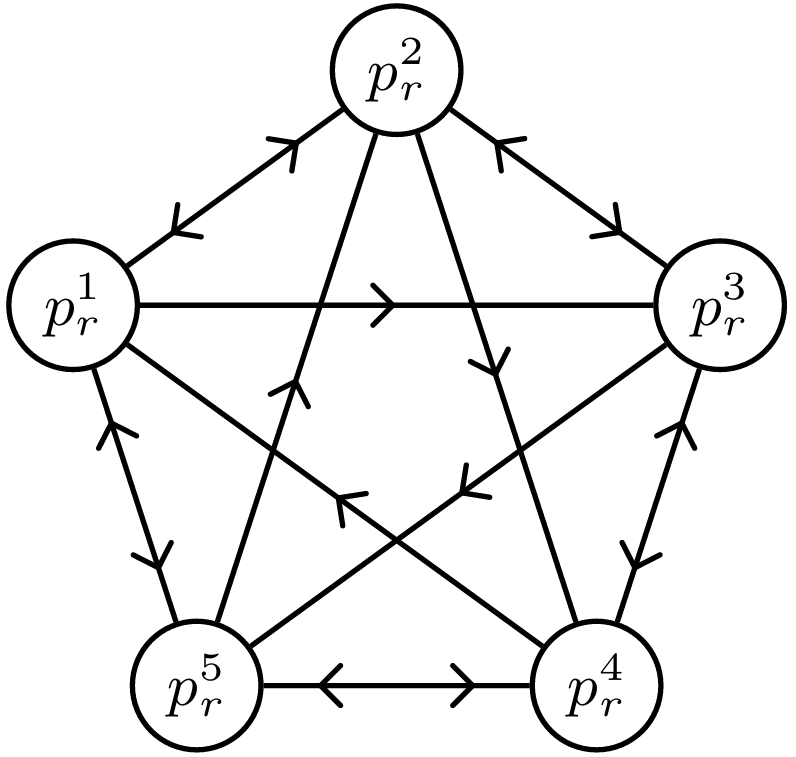}
    };

\node[imagebox, anchor=north west] (verticalchart)
    at (3.0,-0.5) {%
        \includegraphics[
            width=1.05cm,
            height=1.05cm,
            keepaspectratio
        ]{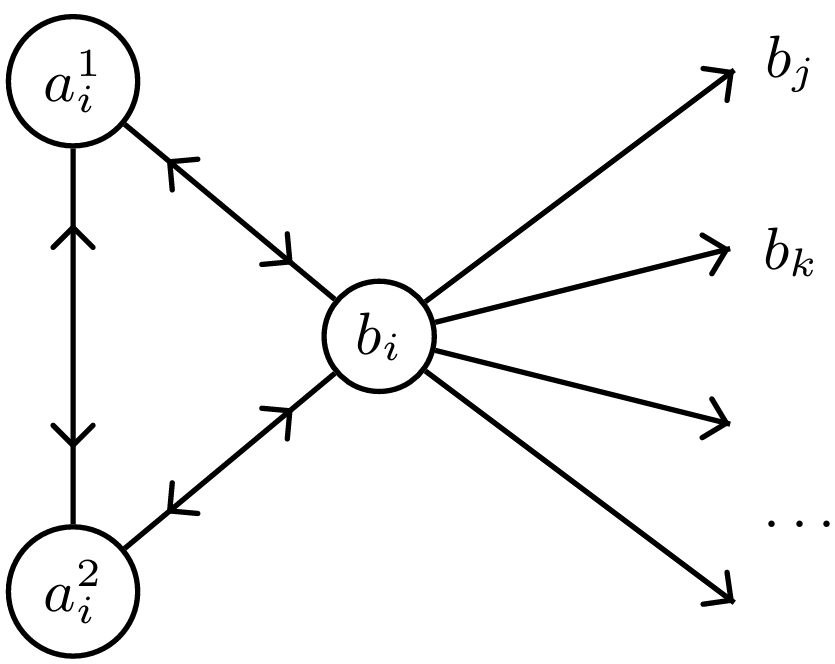}
    };

\node[newlabel] at (3.7,-2.0) {...};

\node[arxivlabel] at (4.3,-2.4) {arxiv/2107.04368};

\draw[pipelinethinarrow]
    (filter.east) to +($(0.33cm,0)$);

\draw[
  red,
  line width=1pt,
  dash pattern=on 5pt off 4pt
]
  (4.3,-1.10)
  ellipse [x radius=1.5cm,y radius=1.5cm];

\node[
    imageencoder,
    anchor=center
] (encoder) at (6.76,-0.2) {%
    \footnotesize IE
};

\node[
    tablebox,
    below=0.6cm of encoder
] (embmatrix) {%
{
\fontsize{3}{3.5}\selectfont
\setlength{\tabcolsep}{0.0pt}
\renewcommand{\arraystretch}{1.5}
\begin{tabular}{ccc}
\toprule
$\mathbf{e_1}$ & $\cdots$ & $\mathbf{e_N}$\\
\midrule
$-0.45$ & $\cdots$ & $0.52$\\
$0.03$ & $\cdots$ & $-0.01$\\
$0.27$ & $\cdots$ & $-0.79$\\[-2em]
$\vdots$ & $\ddots$ & $\vdots$\\
$0.82$ & $\cdots$ & $0.21$\\
\bottomrule
\end{tabular}
}
};

\draw[pipelinearrow]
    (encoder.south) -- ++(0.0,-0.6cm);

\draw[pipelinearrow]
    (5.5cm,-0.2cm) -- (encoder.west);

\node[
    tablebox,
    right=0.15cm of embmatrix,
    yshift=-0.1cm
] (simmatrix) {%
{
\fontsize{3}{3.5}\selectfont
\setlength{\tabcolsep}{0.7pt}
\renewcommand{\arraystretch}{1.15}
\begin{tabular}{c|cccc}
    & $\mathbf{i_1}$ & $\mathbf{i_2}$ & $\cdots$ & $\mathbf{i_N}$\\
\midrule
$\mathbf{i_1}$
    & $1.00$ & \textcolor{blue}{$0.98$} & $\cdots$ & \textcolor{red}{$0.92$}\\
$\mathbf{i_2}$
    & \textcolor{blue}{$0.98$} & $1.00$ & $\cdots$ & \textcolor{red}{$0.86$}\\[-2em]
$\vdots$
    & $\vdots$ & $\vdots$ & $\ddots$ & $\vdots$\\
$\mathbf{i_N}$
    & \textcolor{red}{$0.92$} & \textcolor{red}{$0.86$} & $\cdots$ & $1.00$\\
\end{tabular}
}
};

\draw[pipelinearrow]
    ($(embmatrix.east)+(-0.1,0)$)
    -- ++(0.4,0);

\node[
    pairbox,
    above=0.35cm of simmatrix
] (retainedpair) {%
    \begin{minipage}{1.9cm}
        \centering
        \includegraphics[
            width=2cm,
            height=0.75cm,
            keepaspectratio
        ]{structure/figures_dataset/structure_1.png}
        \\[-1.0em]
        \hspace*{0.6mm}\rule{1.85cm}{0.45pt}
        \\%[-0.15em]
        \includegraphics[
            width=2cm,
            height=0.75cm,
            keepaspectratio
        ]{structure/figures_dataset/structure_2.png}
    \end{minipage}
};

\draw[pipelinearrow]
    (simmatrix.north)
    -- (retainedpair.south);

\node[
    llm,
    right=0.35cm of retainedpair,
    yshift=0.11cm,
    minimum width=2cm,
    minimum height=0.7cm
] (vlmblock) {%
    \begin{minipage}{2.1cm}
    \small
        \textbf{VLM Editing}\\
        \tiny
        Analyze differences between Image X + TikZ X and Image Y + TikZ Y...
    \end{minipage}
};

\node[
    anchor=north east,
    inner sep=0pt
] at ($(vlmblock.north east)+(-0.02cm,-0.04cm)$) {%
    \includegraphics[width=0.38cm]{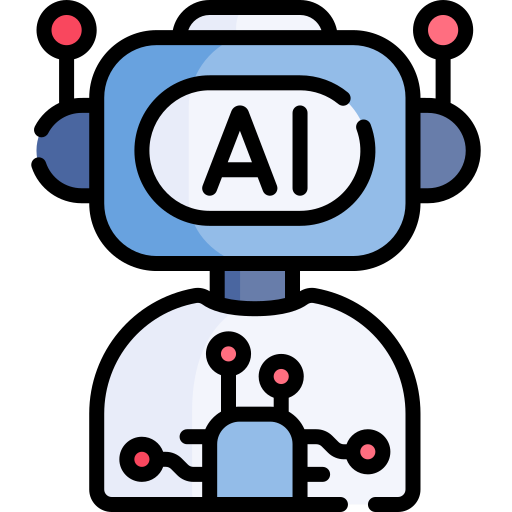}
};

\draw[pipelinearrow]
    (retainedpair.east)
    -- (vlmblock.west |- retainedpair.east);

\node[
    below=0.35cm of vlmblock,
    inner sep=0pt,
    outer sep=0pt,
    draw=none,
    fill=none,
    align=center
] (instructionblock) {%
    {
    \setlength{\tabcolsep}{0.08cm}
    \begin{tabular}{
        @{}
        p{1.25cm}
        @{\hspace{0.08cm}
        }
        p{1.25cm}
        @{}
    }
    \specialrule{0.55pt}{0pt}{-0.4em}
    \centering
    \fontsize{4.2}{4.5}\selectfont
    \textbf{Forward}\\[-0.5em]
    \rule{1.0cm}{0.4pt}\\[0.12em]

    \raggedright
    \fontsize{2.9}{3.2}\selectfont
    ``Delete the continuation to the right of the node labeled
    \smathn{S_{3c}}: remove the adjacent node labeled
    \smathn{\alpha_{w_1}}, the following horizontal ellipsis, the upper
    node labeled \smathn{\alpha_{z_3}}...''
    
    \vspace{0.2em}
    \textbullet\ Delete node labels\\
    \textbullet\ Regroup dashed regions\\
    \textbullet\ Shift annotations

    &
    
    \centering
    \fontsize{4.2}{4.5}\selectfont
    \textbf{Backward}\\[-0.5em]
    \rule{1.0cm}{0.4pt}\\[0.12em]

    \raggedright
    \fontsize{2.9}{3.2}\selectfont
    ``Extend the horizontal chain to the right of \smathn{S_{3c}} by
    adding a new adjacent node labeled \smathn{\alpha_{w_1}}, followed
    by a horizontal ellipsis. Add an upper node labeled \smathn{\alpha_{z_3}}...''
    
    \vspace{0.2em}
    \textbullet\ Add node labels\\
    \textbullet\ Regroup dashed regions\\
    \textbullet\ Shift annotations

    \end{tabular}
    }
};

\draw[pipelinearrow]
    (vlmblock.south)
    -- ++(0,-0.35cm);

\node[
    inner sep=5pt,
    fit=(arxiv)
        (github)
        (tex)
        (filter)
        (pathone)
        (pathtwo)
        (heatmap)
        (verticalchart)
        (encoder)
        (embmatrix)
        (simmatrix)
        (retainedpair)
        (vlmblock)
        (instructionblock)
] (pipelinebounds) {};

\begin{scope}[on background layer]
\node[
    draw=gray!65,
    dashed,
    rounded corners=4pt,
    fill=gray!5,
    line width=0.3pt,
    inner sep=0pt,
    fit={
        ($(pipelinebounds.north west)+(0.05cm,-0.05cm)$)
        ($(pipelinebounds.south east)+(-0.25cm,0.10cm)$)
    }
] (pipelinebox) {};
\end{scope}

\end{tikzpicture}
\end{adjustbox}

\caption{Construction pipeline for DaEdiTikZ. Scientific figures are collected and standardized, grouped by their scientific context, embedded with a scientific image encoder, paired according to cosine similarity, and passed to a VLM to produce bidirectional editing instructions.}
\label{fig:daeditikz-construction}
\end{figure*}

\begin{figure*}[t]
\centering
    \includegraphics[width=\linewidth]{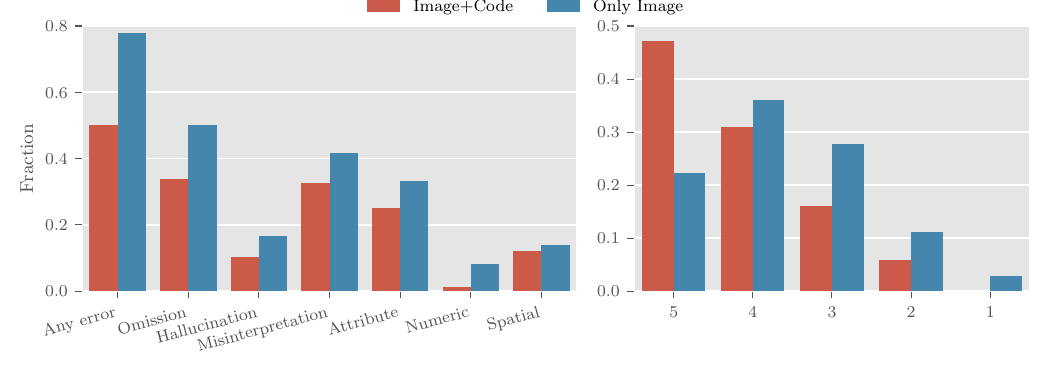}
\caption{Human evaluation of inferred edit instructions. Left: error rates with and without TikZ-code grounding, decomposed into omissions, hallucinations, and attribute, numeric, and spatial misinterpretations. Right: overall instruction quality rated on a 1--5 Likert scale.}
\label{fig:annotation_analysis}
\end{figure*}

\paragraph{Dataset Quality Analysis}
To validate instruction inference, two annotators evaluate 125 revision pairs, including 35 overlapping samples for agreement (Figure~\ref{fig:annotation_analysis}). They identify edit plausibility, omissions, hallucinations, and attribute, numeric, or spatial misinterpretations ($\kappa=0.82$), and rate overall quality on a 1--5 Likert scale (weighted $\kappa=0.79$). Overall, 98\% of retained transformations are plausible and 82.9\% of instructions are rated good (4) or very good (5). While 50\% contain at least one error, these are predominantly omissions (34\%) and misinterpretations (33\%), whereas hallucinations are rare (10\%). To quantify the benefit of code grounding, we repeat the analysis without TikZ code on 90 annotations. The error rate increases from 50\% to 80\%, with omissions increasing by 16 percentage points and numeric misinterpretations from 1\% to 8.5\%, indicating that code provides complementary grounding.

\paragraph{DaEdiTikZ-Bench}
To reduce data contamination, we construct DaEdiTikZ-Bench from arXiv submissions published between March and June 2026. For diversity, one pair per submission is retained with 100 pairs sampled from each similarity interval (0.95--0.96, ..., 0.99--1.00), and 50 pairs spanning group sizes from one to ten. We manually inspect all 500 candidates and remove quality issues, trivial edits, and rendering artifacts, leaving 345 revision pairs and 690 editing instances. Six annotators manually correct every VLM-generated instruction by removing hallucinations, correcting misinterpretations, and adding omissions (Figure~\ref{fig:instruction-annotation-pipeline}).

\newlength{\pipelinerowheight}
\setlength{\pipelinerowheight}{1.60cm}

\newcommand{\figimg}[1]{%
  \parbox[c][\pipelinerowheight][c]{1.78cm}{%
    \centering
    \includegraphics[
      width=1.72cm,
      height=1.48cm,
      keepaspectratio
    ]{#1}%
  }%
}

\newcommand{\instructioncell}[1]{%
  \parbox[t][\pipelinerowheight][t]{3.72cm}{%
    \vspace{-7.3pt}%
    \hspace*{0.4mm}%
    \parbox[t]{3.59cm}{%
      \raggedright
      \fontsize{3.15}{3.45}\selectfont
      #1\par
    }%
  }%
}

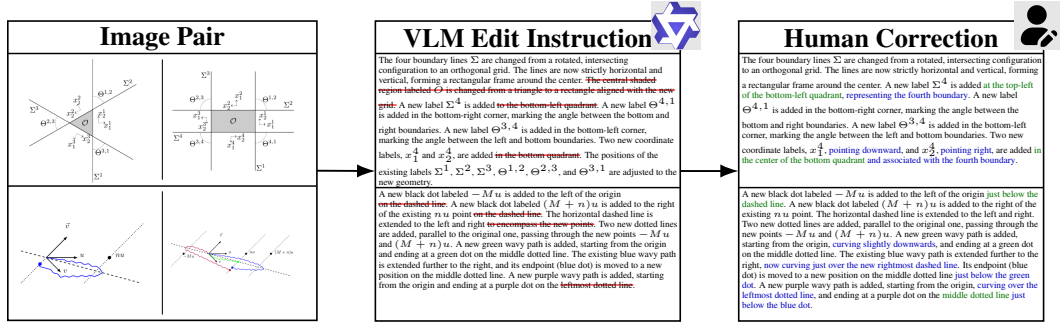
\begin{figure*}[t]
\centering

\begin{adjustbox}{width=\textwidth}
\begin{tikzpicture}[
    stage/.style={
        draw,
        line width=0.55pt,
        inner sep=0pt,
        outer sep=0pt,
        anchor=north west
    },
    pipeline arrow/.style={
        -{Latex[length=2.2mm,width=1.6mm]},
        line width=0.65pt
    }
]

\node[stage] (pairs) {%
\begin{tabular}{
    @{}
    >{\centering\arraybackslash}p{1.78cm}
    @{\hspace{0.25mm}}
    >{\centering\arraybackslash}p{1.78cm}
    @{}
}
\multicolumn{2}{c}{%
    \parbox[c][0.42cm][c]{3.56cm}{%
        \centering\bfseries\small Image Pair
    }%
}
\\[-0.4mm]
\midrule
\figimg{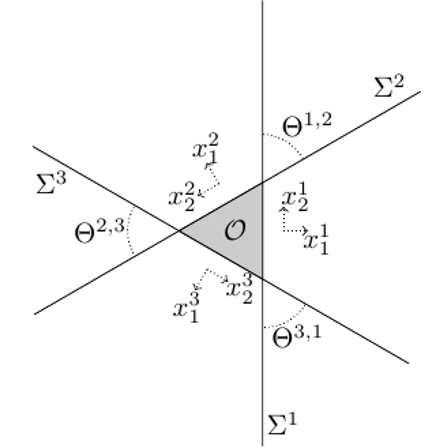}
&
\figimg{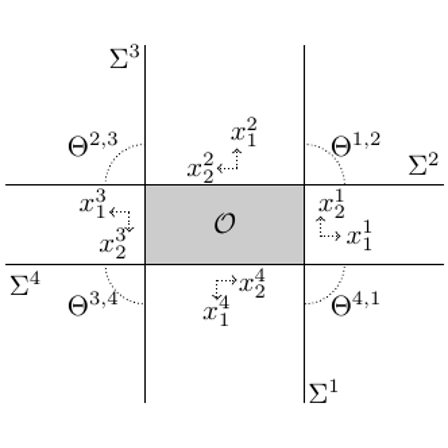}
\\[-0.3mm]
\midrule
\figimg{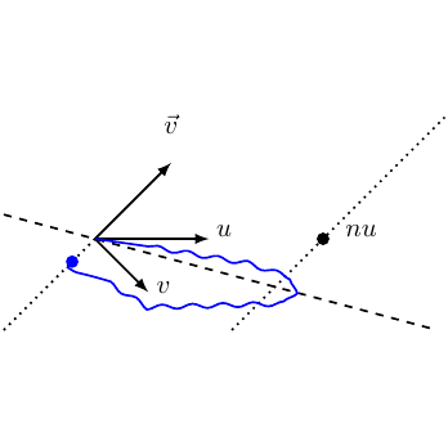}
&
\figimg{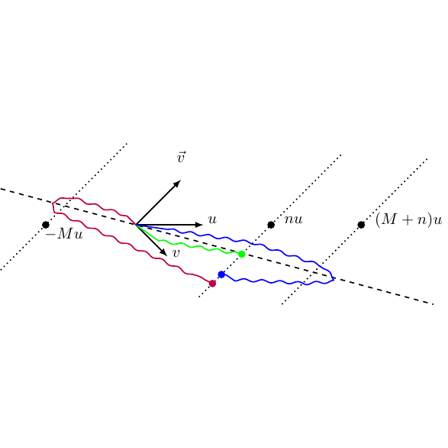}
\\
\end{tabular}
};

\coordinate (pairsepTop) at ($(pairs.north)+(-0.0,-0.46cm)$);
\coordinate (pairsepMidTop) at ($(pairs.north)+(-0.0,-1.9cm)$);
\coordinate (pairsepMidBottom) at ($(pairs.north)+(-0.0,-2.04cm)$);
\coordinate (pairsepBottom) at ($(pairs.south)+(-0.0,0.07cm)$);

\draw[line width=0.35pt]
    (pairsepTop)
    --
    (pairsepMidTop);

\draw[line width=0.35pt]
    (pairsepMidBottom)
    --
    (pairsepBottom);

\node[
    stage,
    right=7mm of pairs.north east,
    anchor=north west
] (generated) {%
\begin{tabular}{@{}p{3.65cm}@{}}
\parbox[c][0.42cm][c]{3.65cm}{%
    \centering\bfseries\small VLM Edit Instruction
}
\\[-0.4mm]
\midrule
\instructioncell{The four boundary lines \smath{\Sigma} are changed from a rotated, intersecting configuration to an orthogonal grid. The lines are now strictly horizontal and vertical, forming a rectangular frame around the center. \redstrike{The central shaded} \redstrike{region labeled \smath{O} is changed from a triangle to a rectangle aligned with the new} \redstrike{grid.} A new label \smath{\Sigma^{4}} is added \redstrike{to the bottom-left quadrant}. A new label \smath{\Theta^{4,1}} is added in the bottom-right corner, marking the angle between the bottom and right boundaries. A new label \smath{\Theta^{3,4}} is added in the bottom-left corner, marking the angle between the left and bottom boundaries. Two new coordinate labels, \smath{x^{4}_{1}} and \smath{x^{4}_{2}}, are added \redstrike{in the bottom quadrant}. The positions of the existing labels
\smath{\Sigma^{1}}, \smath{\Sigma^{2}}, \smath{\Sigma^{3}}, \smath{\Theta^{1,2}}, \smath{\Theta^{2,3}}, and \smath{\Theta^{3,1}} are adjusted to the new geometry.}
\\[-2.9mm]
\midrule
\instructioncell{A new black dot labeled \smath{-Mu} is added to the left of the origin \redstrike{on the dashed line}. A new black dot labeled \smath{(M+n)u} is added to the right of the existing \smath{nu} point \redstrike{on the dashed line}. The horizontal dashed line is extended to the left and right \redstrike{to encompass the new points}. Two new dotted lines are added, parallel to the original one, passing through the new points \smath{-Mu} and \smath{(M+n)u}. A new green wavy path is added, starting from the origin and ending at a green dot on the middle dotted line. The existing blue wavy path is extended further to the right, and its endpoint (blue dot) is moved to a new position on the middle dotted line. A new purple wavy path is added, starting from the origin and ending at a purple dot on the \redstrike{leftmost dotted line}.}
\\[-2.9mm]
\end{tabular}
};

\draw[pipeline arrow]
    ($(pairs.east)$)
    --
    ($(generated.west)$);

\node[
    anchor=center,
    inner sep=0pt,
    fill=gray!20
] at ($(generated.north east)+(-0.5mm,-0.5mm)$) {%
    \includegraphics[
        width=5.8mm,
        height=5.8mm,
        keepaspectratio
    ]{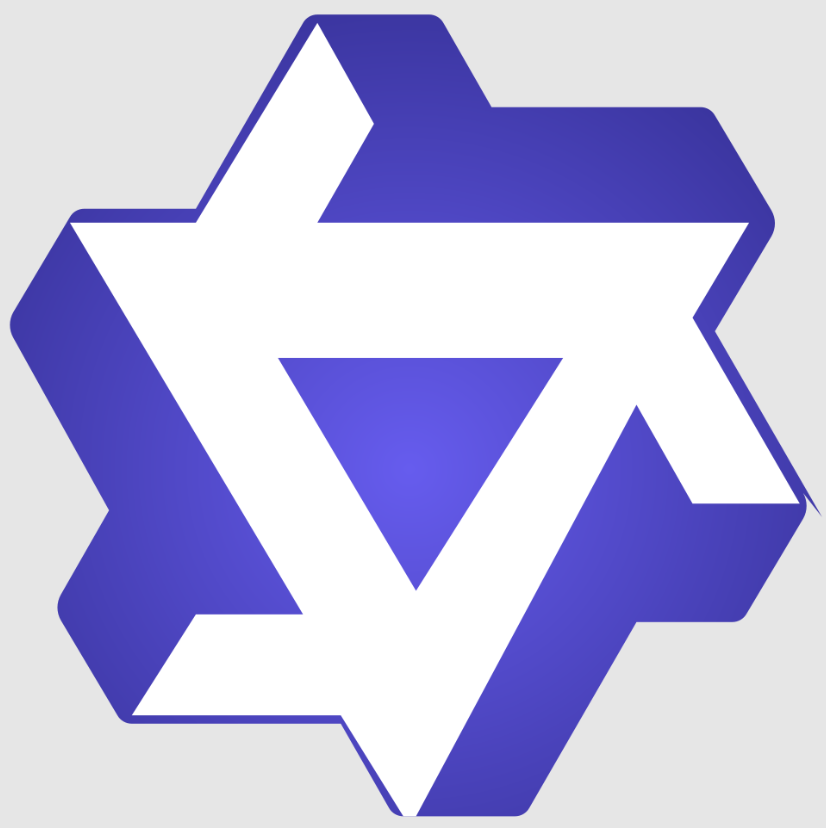}%
};

\node[
    stage,
    right=7mm of generated.north east,
    anchor=north west
] (corrected) {%
\begin{tabular}{@{}p{3.65cm}@{}}
\parbox[c][0.42cm][c]{3.65cm}{%
    \centering\bfseries\small Human Correction
}
\\[-0.4mm]
\midrule
\instructioncell{The four boundary lines \smath{\Sigma} are changed from a rotated, intersecting configuration to an orthogonal grid. The lines are now strictly horizontal and vertical, forming a rectangular frame around the center. A new label \smath{\Sigma^{4}} is added
\greencorr{at the top-left of the bottom-left quadrant}, \blueadd{representing the fourth boundary}. A new label \smath{\Theta^{4,1}} is added in the bottom-right
corner, marking the angle between the bottom and right boundaries. A new label \smath{\Theta^{3,4}} is added in the bottom-left corner, marking the angle between the left and bottom boundaries. Two new coordinate labels,
\smath{x^{4}_{1}}, \blueadd{pointing downward}, and \smath{x^{4}_{2}}, \blueadd{pointing right}, are added \greencorr{in the center of the bottom quadrant} \blueadd{and associated with the fourth boundary}.}
\\[-2.9mm]
\midrule
\instructioncell{A new black dot labeled \smath{-Mu} is added to the left of the origin \greencorr{just below the dashed line}. A new black dot labeled \smath{(M+n)u} is added to the right of the existing \smath{nu} point. The horizontal dashed line is extended to the left and right. Two new dotted lines are added, parallel to the original one, passing through the new points \smath{-Mu} and \smath{(M+n)u}. A new green wavy path is added, starting from the origin, \blueadd{curving slightly downwards}, and ending at a green dot on the middle dotted line. The existing blue wavy path is extended further to the right,
\blueadd{now curving just over the new rightmost dashed line}. Its endpoint (blue dot) is moved to a new position on the middle dotted line \blueadd{just below the green dot}. A new purple wavy path is added, starting from the origin, \blueadd{curving over the leftmost dotted line}, and ending at a purple dot on the \greencorr{middle dotted line} \blueadd{just below the blue dot}.}
\\[-2.9mm]
\end{tabular}
};

\draw[pipeline arrow]
    ($(generated.east)$)
    --
    ($(corrected.west)$);

\node[
    anchor=north east,
    inner sep=0pt,
    fill=gray!20
] at ($(corrected.north east)+(2.4mm,2.4mm)$) {%
    \includegraphics[
        width=5.8mm,
        height=5.8mm,
        keepaspectratio
    ]{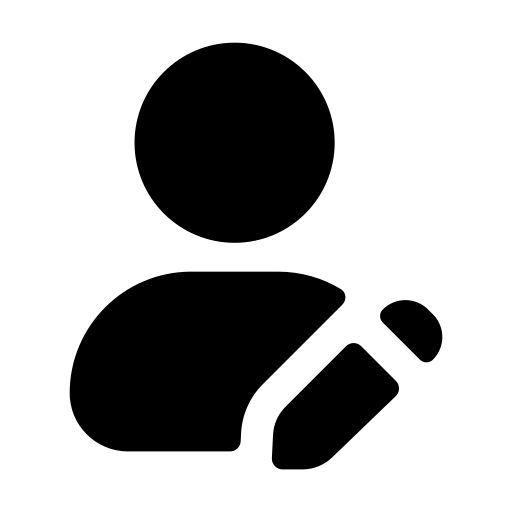}%
};

\end{tikzpicture}
\end{adjustbox}

\caption{Examples of human-refined benchmark instructions. \redstrike{Red strikethrough} marks removed errors, \greencorr{green} marks corrections, and \blueadd{blue} marks added omissions.}
\label{fig:instruction-annotation-pipeline}

\end{figure*}

\section{Editing-Specific Post-Training}
\label{sec:Editing-Specific Post-Training}

\paragraph{Joint Reconstruction and Editing SFT}
DaEdiTikZ provides 752K source figure--instruction--TikZ target triplets $(I_s,u,y)$, where $y=(y_1,\ldots,y_T)$ and $I_t$ denotes the rendered target figure. We minimize:
\begin{equation}
\mathcal{L}_{\mathrm{edit}}(\theta)=
\mathbb{E}_{(I_s,u,y)\sim\mathcal{D}_{\mathrm{edit}}}
\left[-\sum_{t=1}^{T}\log p_\theta(y_t\mid y_{<t},I_s,u)\right]
\label{eq:sft_edit}
\end{equation}
Since editing requires reconstructing the source figure while selectively modifying it, we jointly train with 752K image-to-TikZ reconstruction samples from DaTikZ-V4. Reconstruction uses the same objective over $(I_t,y)\sim\mathcal{D}_{\mathrm{rec}}$, conditioned only on $I_t$, strengthening the shared image-to-TikZ mapping while exposing the model to a broader distribution of scientific figures and TikZ programs.

\paragraph{Editing-Specific Rewards}
We further optimize the resulting SFT model using rewards computed from sampled TikZ rollouts $\hat y$ and their renderings $\hat I$. Unlike TikZilla, which trains a separate scientific image encoder~\citep{greisinger2026tikzilla}, we reuse a frozen copy of the SFT model's vision encoder. SFT already adapts this encoder to scientific figures on 1.5M editing and reconstruction samples. We freeze it during RL to prevent reward hacking. Given patch embeddings $\mathbf{x}=\{x_i\}_{i=1}^{N}$ and $\mathbf{z}=\{z_j\}_{j=1}^{M}$ of $I_t$ and $\hat I$, respectively, we compute:
\begin{equation}
D_{ij}
=
1-\cos(x_i,z_j),
\qquad
d_{\mathrm{EMD}}(\mathbf{x},\mathbf{z})
=
\min_{F\geq 0}
\sum_{i=1}^{N}\sum_{j=1}^{M}F_{ij}D_{ij}
\label{eq:selfsim_emd}
\end{equation}
subject to uniform marginals
$\sum_jF_{ij}=1/N$ and $\sum_iF_{ij}=1/M$.
The SelfSim reward is:
\begin{equation}
\mathcal{R}_{\mathrm{SSim}}
=
\operatorname{clip}
\left(
1+2\tanh[-d_{\mathrm{EMD}}(\mathbf{x},\mathbf{z})],
0,1
\right)
\label{eq:selfsim_reward}
\end{equation}
However, target similarity alone is insufficient for editing. First, DaEdiTikZ contains similar source--target pairs, allowing high $\mathcal{R}_{\mathrm{SSim}}$ from preserving unchanged content without applying the requested edits. Second, VLM-inferred instructions may contain omissions or inaccuracies, such that the target may not exactly realize the instruction and can penalize valid instruction-following outputs. We therefore introduce a complementary reference-free instruction-following reward $\mathcal{R}_{\mathrm{IF}}$. A VLM judge (Qwen3.6-27B) receives $(I_s,u,\hat I)$ and verifies each of the $K$ atomic edits with a binary score $v_k\in\{0,1\}$. We set $\mathcal{R}_{\mathrm{IF}}=\frac{1}{K}\sum_k v_k(I_s, u, \hat I)$, giving proportional credit for partially applied instructions. Finally, we define compilation and format validity as $\mathcal{R}_{\mathrm{Comp}}=\mathbbm{1}[\operatorname{compile}(\hat y)]$ and $\mathcal{R}_{\mathrm{Fmt}}=\mathbbm{1}[\operatorname{valid\_format}(\hat y)]$, where the latter requires the expected standalone TikZ structure (\texttt{\textbackslash documentclass[tikz]\{standalone\}}, \texttt{\textbackslash begin\{document\}}, ..., \texttt{\textbackslash end\{document\}}). Compilation and format validity gate both rewards: $r_m=\mathcal{R}_{\mathrm{Comp}}\mathcal{R}_{\mathrm{Fmt}}\mathcal{R}_m$ for $m\in\mathcal{M}=\{\mathrm{SSim},\mathrm{IF}\}$, assigning failed rollouts zero reward. Figure~\ref{fig:method_figure} summarizes the two-stage pipeline.

\setkeys{Gin}{keepaspectratio}

\newcommand{\smathm}[1]{\scalebox{0.35}{$#1$}}

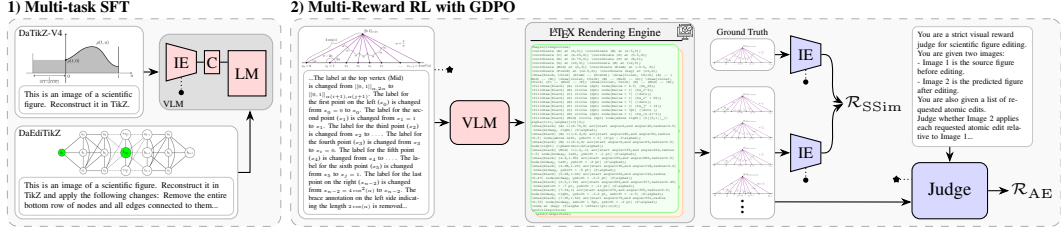
\begin{figure*}[t]
\centering

\begin{adjustbox}{max width=\textwidth}
\begin{tikzpicture}[
    x=1cm,
    y=1cm,
    stage/.style={
    draw=gray!65,
    dashed,
    rounded corners=4pt,
    fill=gray!5,
    line width=0.3pt,
    inner sep=0pt
},
branch/.style={
    draw=gray!55,
    rounded corners=3pt,
    fill=white,
    line width=0.3pt,
    inner sep=0pt
},
branchtitle/.style={
    font=\fontsize{3.5}{4}\selectfont,
    anchor=west,
    inner sep=0pt
},
prompt/.style={
    draw=gray!45,
    rounded corners=2pt,
    fill=white,
    line width=0.3pt,
    align=left,
    inner sep=2pt
},
img/.style={
    inner sep=0pt,
    outer sep=0pt,
    anchor=center
},
forward/.style={
    -{Stealth[length=1.4mm]},
    line width=0.4pt
},
connector/.style={
    line width=0.4pt
},
modelbox/.style={
    draw=gray!60,
    rounded corners=4pt,
    fill=gray!25,
    line width=0.4pt,
    inner sep=0pt
},
trainmodule/.style={
    draw=black,
    fill=red!15,
    line width=0.4pt,
    align=center,
    font=\tiny,
},
ie/.style={
    trainmodule,
    trapezium,
    trapezium left angle=75,
    trapezium right angle=75,
    shape border rotate=270,
    minimum width=0.3cm,
    minimum height=0.4cm,
    inner sep=0pt
},
mlp/.style={
    trainmodule,
    rectangle,
    minimum width=0.2cm,
    minimum height=0.305cm,
    inner sep=0pt
},
llm/.style={
    trainmodule,
    rectangle,
    minimum width=0.5cm,
    minimum height=0.6cm,
    inner sep=0pt
}
]

\node[
    stage,
    anchor=south west,
    minimum width=3.62cm,
    minimum height=2.75cm
] (sftstage) at (0,0) {};

\node[
    font=\bfseries\tiny,
    anchor=south west,
    inner sep=0pt
] at ($(sftstage.north west)+(0,-0.00cm)+(0,0.04cm)$)
{1) Multi-task SFT};

\node[
    branch,
    anchor=north west,
    minimum width=1.7cm,
    minimum height=1.21cm
] (reconstructionbranch)
at ($(sftstage.north west)+(0.10cm,-0.10cm)$) {};

\node[
    img,
    anchor=north
] (reconstructionimage)
at ($(reconstructionbranch.north)+(0,-0.10cm)$)
{%
    \includegraphics[
        width=1.4cm,
        height=1.4cm
    ]{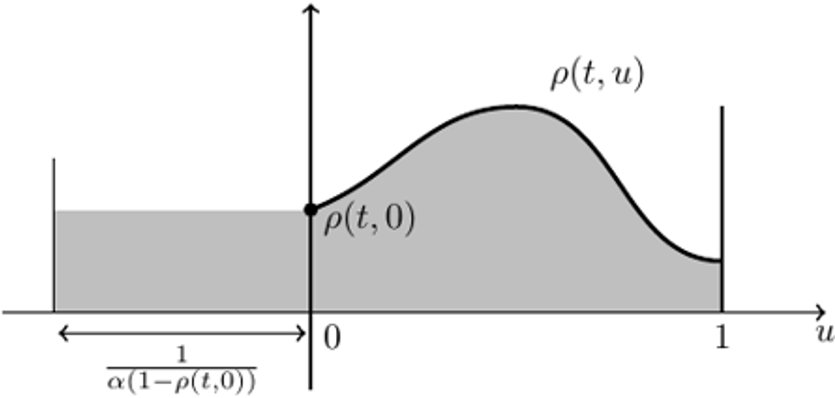}%
};

\node[
    branchtitle,
    anchor=north west
] (reconstructiontitle)
at ($(reconstructionbranch.north west)+(0.05cm,-0.04cm)$)
{DaTikZ-V4};

\node[
    prompt,
    anchor=north,
    text width=1.45cm,
    minimum height=0.35cm,
    font=\fontsize{3.2}{3.7}\selectfont
] (reconstructionprompt)
at ($(reconstructionimage.south)+(0,-0.02cm)$)
{%
    This is an image of a scientific figure.
    Reconstruct it in TikZ.
};

\node[
    branch,
    anchor=north west,
    minimum width=2.95cm,
    minimum height=1.23cm
] (editingbranch)
at ($(reconstructionbranch.south west)+(0,-0.10cm)$) {};

\node[
    img,
    anchor=north
] (editingimage)
at ($(editingbranch.north)+(0,-0.05cm)$)
{%
    \includegraphics[
        width=1.8cm,
        height=1.8cm
    ]{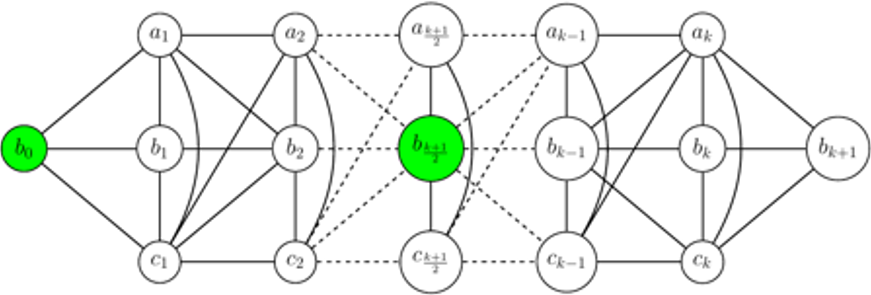}%
};

\node[
    branchtitle,
    anchor=north west
] (editingtitle)
at ($(editingbranch.north west)+(0.05cm,-0.04cm)$)
{DaEdiTikZ};

\node[
    prompt,
    anchor=north,
    text width=2.70cm,
    minimum height=0.50cm,
    font=\fontsize{3.2}{3.7}\selectfont
] (editingprompt)
at ($(editingimage.south)+(0,-0.02cm)$)
{%
    This is an image of a scientific figure.
    Reconstruct it in TikZ and apply the following changes:
    Remove the entire bottom row of nodes and all edges
    connected to them...
};

\node[
    modelbox,
    anchor=west,
    minimum height=1.0cm,
    minimum width=1.5cm,
] (vlmbox)
at ($(reconstructionbranch.east)+(0.2cm,0.0cm)$) {};

\node[
    branchtitle,
    anchor=south west
] (vlmtitle)
at ($(vlmbox.south west)+(0.05cm,0.04cm)$)
{VLM};

\node[
    ie,
    anchor=north west
] (imageencoder)
at ($(vlmbox.north west)+(0.09cm,-0.17cm)$)
{IE};

\node[
    circle,
    fill=black,
    inner sep=0.7pt
] (Portal)
at ($(imageencoder.south)+(0,-0.2cm)$) {};

\draw[
    connector,
    densely dotted
]
(imageencoder.south) -- (Portal.north);

\node[
    mlp,
    anchor=north west
] (mlpconnector)
at ($(imageencoder.north east)+(0.14cm,0)$)
{C};

\node[
    llm,
    anchor=north west
] (llmdecoder)
at ($(mlpconnector.north east)+(0.10cm,0)$)
{LM};

\draw[connector]
    (imageencoder.east)
    --
    (mlpconnector.west);

\draw[connector]
    (mlpconnector.east)
    --
    (llmdecoder.west |- mlpconnector.east);

\draw[forward]
    (editingbranch.east)
    -|
    ($(vlmbox.south)+(0.45cm,0)$);

\draw[forward]
    (reconstructionbranch.east)
    --
    (vlmbox.west);

\tikzset{
    rlvlm/.style={
        draw=black,
        fill=red!15,
        rounded corners=3pt,
        line width=0.4pt,
        minimum width=0.78cm,
        minimum height=0.58cm,
        align=center,
        font=\tiny
    },
    judgevlm/.style={
        draw=black,
        fill=blue!15,
        rounded corners=3pt,
        line width=0.4pt,
        minimum width=1.0cm,
        minimum height=0.744cm,
        align=center,
        font=\scriptsize
    },
    compilerbox/.style={
        draw=gray!60,
        fill=gray!10,
        rounded corners=3pt,
        line width=0.4pt,
        inner sep=2pt,
        minimum width=1.78cm,
        minimum height=1.65cm,
        align=left
    },
    rolloutbox/.style={
        draw=gray!55,
        fill=white,
        rounded corners=3pt,
        line width=0.3pt,
        inner sep=0pt,
        minimum width=2.90cm,
        minimum height=0.83cm
    },
    frozenie/.style={
        ie,
        fill=blue!15
    }
}

\node[
    stage,
    anchor=north west,
    minimum width=10.2cm,
    minimum height=2.75cm
] (rlstage)
at ($(sftstage.north east)+(0.12cm,0)$) {};

\node[
    font=\bfseries\tiny,
    anchor=south west,
    inner sep=0pt
] (rltitle)
at ($(rlstage.north west)+(0,0.04cm)$)
{2) Multi-Reward RL with GDPO};

\node[
    branch,
    anchor=north west,
    minimum width=1.80cm,
    minimum height=2.55cm
] (rlinput)
at ($(rlstage.north west)+(0.10cm,-0.10cm)$) {};

\node[
    img,
    anchor=north
] (rlsource)
at ($(rlinput.north)+(0,-0.05cm)$)
{%
    \includegraphics[
        width=1.71cm,
        height=1.20cm
    ]{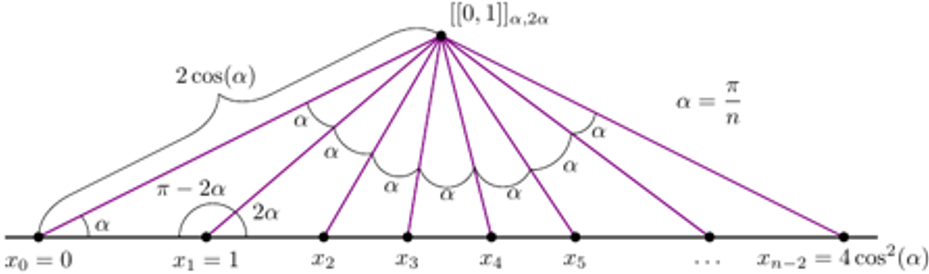}%
};

\node[
    prompt,
    anchor=north,
    text width=1.52cm,
    minimum height=0.85cm,
    font=\fontsize{2.5}{3}\selectfont
] (rlprompt)
at ($(rlsource.south)+(0,-0.03cm)$)
{%
...The label at the top vertex (Mid) is changed from \smathm{[[0,1]]_{\alpha,2\alpha}} to \smathm{[[0,1]]_{\alpha(i+1),\alpha(j+1)}}.
The label for the first point on the left (\smathm{x_0}) is changed from \smathm{x_0 = 0} to \smathm{x_0}.
The label for the second point (\smathm{x_1}) is changed from \smathm{x_1 = 1} to \smathm{x_1}.
The label for the third point (\smathm{x_2}) is changed from \smathm{x_2} to \smathm{\dots}.
The label for the fourth point (\smathm{x_3}) is changed from \smathm{x_3} to \smathm{x_i = 0}.
The label for the fifth point (\smathm{x_4}) is changed from \smathm{x_4} to \smathm{\dots}.
The label for the sixth point (\smathm{x_5}) is changed from \smathm{x_5} to \smathm{x_j = 1}.
The label for the last point on the right (\smathm{x_{n-2}}) is changed from \smathm{x_{n-2} = 4\cos^2(\alpha)} to \smathm{x_{n-2}}.
The brace annotation on the left side indicating the length \smathm{2\cos(\alpha)} is removed...
};

\node[
    star,
    fill=black,
    inner sep=0.7pt
] (RLPortalStar)
at ($(rlinput.east)+(0.2,0.7cm)$) {};

\draw[
    connector,
    densely dotted
]
($(rlinput.east)+(0.0,0.7cm)$) -- (RLPortalStar.west);

\node[
    rlvlm,
    anchor=west
] (rlvlm)
at ($(rlinput.east)+(0.20cm,0)$)
{VLM};

\draw[forward]
    (rlinput.east)
    --
    (rlvlm.west);

\node[
    compilerbox,
    anchor=west
] (compiler)
at ($(rlvlm.east)+(0.20cm,0)$)
{%
    \begin{minipage}{2.01cm}
        \centering

        {\fontsize{4.3}{4.8}\selectfont
        \LaTeX\ Rendering Engine
        }

        \vspace{-9pt}
        \rule{\linewidth}{0.25pt}
        \vspace{-21pt}

        \begin{tikzpicture}[baseline]
            \node[
                anchor=north west,
                inner sep=0pt
            ] (backcode) at (0.06cm,-0.06cm) {%
                \begin{tcolorbox}[
                    enhanced,
                    colback=orange!10,
                    colframe=orange!45,
                    boxrule=0.3pt,
                    arc=2pt,
                    left=0pt,
                    right=0pt,
                    top=0pt,
                    bottom=0pt,
                    boxsep=0pt,
                    width=1.96cm
                ]
\begin{lstlisting}[
                        language=,
                        breaklines=true,
                        breakautoindent=false,
                        breakindent=0pt,
                        columns=fullflexible,
                        keepspaces=true,
                        tabsize=1,
                        showstringspaces=false,
                        aboveskip=0pt,
                        belowskip=0pt,
                        xleftmargin=0pt,
                        basewidth=0.34em,
                        basicstyle=\fontsize{1.3}{1.5}\selectfont\ttfamily
                    ]
\begin{tikzpicture}
\coordinate (A) at (0,0); \coordinate (B) at (2.5,0);
\coordinate (C) at (4.25,0); \coordinate (D) at (5.5,0);
\coordinate (E) at (6.75,0); \coordinate (F) at (8,0);
\coordinate (G) at (10,0); \coordinate (H) at (12,0);
\coordinate (Mid) at (6,3); \coordinate (PreA) at (-0.5, 0);
\coordinate (PostH) at (12.5,0); \coordinate (key) at (10,2);
\draw[black, thick] (PreA) -- (PostH); \draw[violet, thick] (A) -- (Mid) -- (H); \draw[violet, thick] (B) -- (Mid) -- (G); \draw[violet, thick] (C) -- (Mid) -- (F); \draw[violet, thick] (D) -- (Mid) -- (E);
\filldraw[black] (A) circle (2pt) node[below = 2.5] {$x_0$};
\filldraw[black] (B) circle (2pt) node[below = 1] {$x_1'$};
\filldraw[black] (C) circle (2pt) node[below = 7] {\dots};
\filldraw[black] (D) circle (2pt) node[below = 1] {$x_i' = 0$};
\filldraw[black] (E) circle (2pt) node[below = 7] {\dots};
\filldraw[black] (F) circle (2pt) node[below = 1] {$x_j' = 1$};
\filldraw[black] (G) circle (2pt) node[below = 7pt] {\dots };
\filldraw[black] (H) circle (2pt) node[below = 1] {$x_{n-2}'$};
\filldraw[black] (Mid) circle (2pt) node[above right] {$[[0,1]]_{\alpha(i+1),\alpha(j+1)}$};
\draw[black] (A) ++(0.75,0) arc[start angle=0,end angle=40,radius=0.5] node[midway, right] {$\alpha$};
\draw[black] (B) ++(-0.4,0) arc[start angle=180,end angle=50,radius=0.5] node[above left, yshift = 2] {$\pi - 2\alpha$};
\draw[black] (B) ++(0.6,0) arc[start angle=0,end angle=49,radius=0.5] node[right] {\phantom{s}$2\alpha$};
\draw[black] (Mid) ++(-2,-1) arc[start angle=195,end angle=262,radius=.5] node[midway, left, yshift = -1 pt] {$\alpha$};
\draw[black] (4.4,1.65) arc[start angle=190,end angle=280,radius=0.5] node[midway, left, yshift = -2 pt] {$\alpha$};
\draw[black] (4.98,1.23) arc[start angle=190,end angle=318,radius=0.4] node[midway, yshift = -5 pt] {$\alpha$};
\draw[black] (5.68,1.02) arc[start angle=200,end angle=343,radius=0.43] node[midway, yshift = -3.5 pt] {$\alpha$};
\draw[black] (6.5,1.04) arc[start angle=200,end angle=337,radius=0.45] node[xshift = -7 pt, yshift = -11 pt] {$\alpha$};
\draw[black] (7.34,1) arc[start angle=270,end angle=355,radius=0.6] node[midway, right, yshift = -3.2 pt, xshift = -1.5] {$\alpha$};
\draw[black] (7.95,1.54) arc[start angle=270,end angle=351,radius=0.35] node[midway, xshift = 5pt, yshift = -2 pt] {$\alpha$};
\node at (key) {$\alpha = \dfrac{\pi}{n}$};
\end{tikzpicture}
\end{lstlisting}
                \end{tcolorbox}
            };

            \node[
                anchor=north west,
                inner sep=0pt
            ] (frontcode) at (0,0) {%
                \begin{tcolorbox}[
                    enhanced,
                    colback=green!5,
                    colframe=green!50,
                    boxrule=0.3pt,
                    arc=2pt,
                    left=0pt,
                    right=0pt,
                    top=0pt,
                    bottom=0pt,
                    boxsep=0pt,
                    width=1.96cm
                ]
\begin{lstlisting}[
                        language=,
                        breaklines=true,
                        breakautoindent=false,
                        breakindent=0pt,
                        columns=fullflexible,
                        keepspaces=true,
                        tabsize=1,
                        showstringspaces=false,
                        aboveskip=0pt,
                        belowskip=0pt,
                        xleftmargin=0pt,
                        basewidth=0.34em,
                        basicstyle=\fontsize{1.3}{1.5}\selectfont\ttfamily
                    ]
\begin{tikzpicture}
\coordinate (A) at (0,0); \coordinate (B) at (2.5,0);
\coordinate (C) at (4.25,0); \coordinate (D) at (5.5,0);
\coordinate (E) at (6.75,0); \coordinate (F) at (8,0);
\coordinate (G) at (10,0); \coordinate (H) at (12,0);
\coordinate (Mid) at (6,3); \coordinate (PreA) at (-0.5, 0);
\coordinate (PostH) at (12.5,0); \coordinate (key) at (10,2);
\draw[black, thick] (PreA) -- (PostH); \draw[violet, thick] (A) -- (Mid) -- (H); \draw[violet, thick] (B) -- (Mid) -- (G); \draw[violet, thick] (C) -- (Mid) -- (F); \draw[violet, thick] (D) -- (Mid) -- (E);
\filldraw[black] (A) circle (2pt) node[below = 2.5] {$x_0$};
\filldraw[black] (B) circle (2pt) node[below = 1] {$x_1'$};
\filldraw[black] (C) circle (2pt) node[below = 7] {\dots};
\filldraw[black] (D) circle (2pt) node[below = 1] {$x_i' = 0$};
\filldraw[black] (E) circle (2pt) node[below = 7] {\dots};
\filldraw[black] (F) circle (2pt) node[below = 1] {$x_j' = 1$};
\filldraw[black] (G) circle (2pt) node[below = 7pt] {\dots };
\filldraw[black] (H) circle (2pt) node[below = 1] {$x_{n-2}'$};
\filldraw[black] (Mid) circle (2pt) node[above right] {$[[0,1]]_{\alpha(i+1),\alpha(j+1)}$};
\draw[black] (A) ++(0.75,0) arc[start angle=0,end angle=40,radius=0.5] node[midway, right] {$\alpha$};
\draw[black] (B) ++(-0.4,0) arc[start angle=180,end angle=50,radius=0.5] node[above left, yshift = 2] {$\pi - 2\alpha$};
\draw[black] (B) ++(0.6,0) arc[start angle=0,end angle=49,radius=0.5] node[right] {\phantom{s}$2\alpha$};
\draw[black] (Mid) ++(-2,-1) arc[start angle=195,end angle=262,radius=.5] node[midway, left, yshift = -1 pt] {$\alpha$};
\draw[black] (4.4,1.65) arc[start angle=190,end angle=280,radius=0.5] node[midway, left, yshift = -2 pt] {$\alpha$};
\draw[black] (4.98,1.23) arc[start angle=190,end angle=318,radius=0.4] node[midway, yshift = -5 pt] {$\alpha$};
\draw[black] (5.68,1.02) arc[start angle=200,end angle=343,radius=0.43] node[midway, yshift = -3.5 pt] {$\alpha$};
\draw[black] (6.5,1.04) arc[start angle=200,end angle=337,radius=0.45] node[xshift = -7 pt, yshift = -11 pt] {$\alpha$};
\draw[black] (7.34,1) arc[start angle=270,end angle=355,radius=0.6] node[midway, right, yshift = -3.2 pt, xshift = -1.5] {$\alpha$};
\draw[black] (7.95,1.54) arc[start angle=270,end angle=351,radius=0.35] node[midway, xshift = 5pt, yshift = -2 pt] {$\alpha$};
\node at (key) {$\alpha = \dfrac{\pi}{n}$};
\end{tikzpicture}
\end{lstlisting}
                \end{tcolorbox}
            };
        \end{tikzpicture}
    \end{minipage}
};

\node[
    anchor=north east,
    inner sep=0pt
] at ($(compiler.north east)+(-0.025cm,-0.025cm)$) {%
    \includegraphics[width=0.2cm]{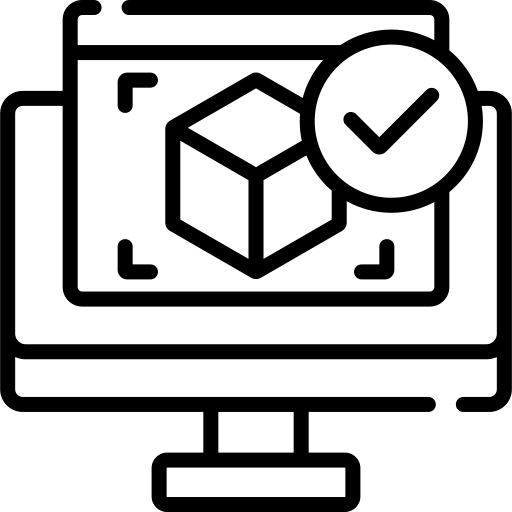}
};

\draw[forward]
    (rlvlm.east)
    --
    (compiler.west);

\node[
    rolloutbox,
    anchor=west,
    minimum width=0.85cm,
    minimum height=1.75cm
] (rolloutbox)
at ($(compiler.east)+(0.20cm,-0.40cm)$) {};

\node[
    img,
    anchor=north
] (rlpredone)
at ($(rolloutbox.north)+(0,-0.1cm)$)
{%
    \includegraphics[
        width=0.8cm,
        height=0.8cm
    ]{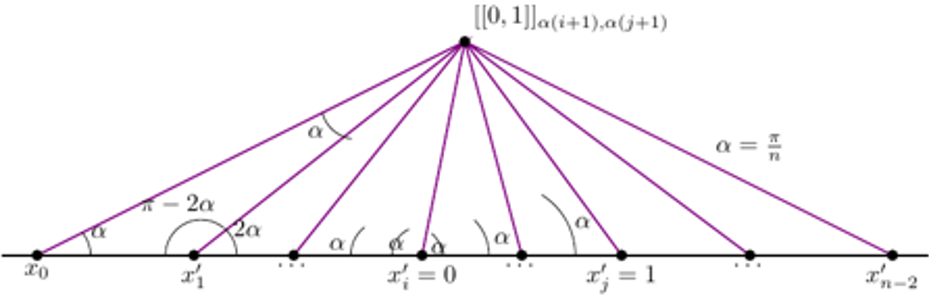}%
};

\node[
    img,
    anchor=north
] (rlpredtwo)
at ($(rlpredone.south)+(0,-0.1cm)$)
{%
    \includegraphics[
        width=0.8cm,
        height=0.8cm
    ]{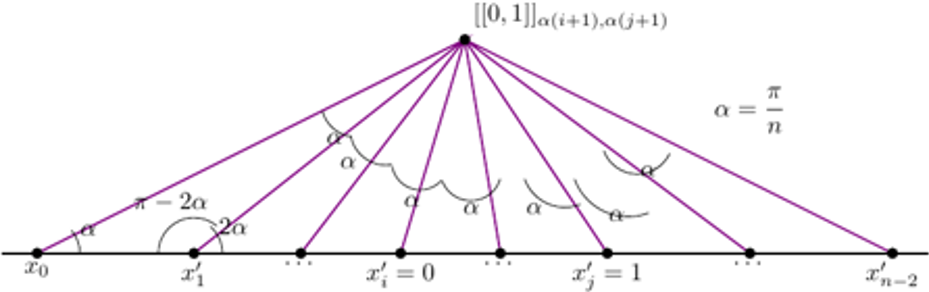}%
};

\node[
    img,
    anchor=north
] (rlpredthree)
at ($(rlpredtwo.south)+(0,-0.1cm)$)
{%
    \includegraphics[
        width=0.8cm,
        height=0.8cm
    ]{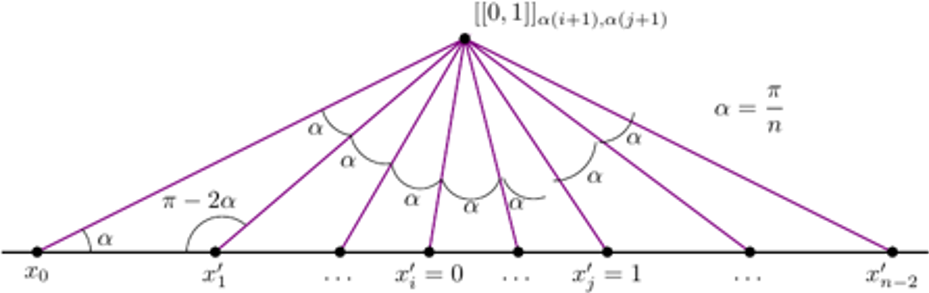}%
};

\node[
    img,
    anchor=north
] (rlpredfour)
at ($(rlpredthree.south)+(0,-0.10cm)$)
{%
    \includegraphics[
        width=0.8cm,
        height=0.8cm
    ]{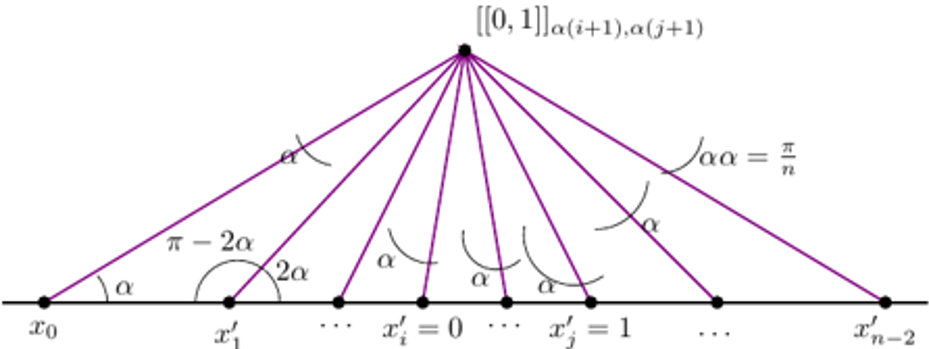}%
};

\node[
    anchor=north,
    font=\small
] at ($(rlpredfour.south)+(0.0,0.12cm)$)
{\shortstack{$\cdot$\\[-5pt]$\cdot$\\[-5pt]$\cdot$}};

\draw[forward]
    ($(compiler.east)+(0.0,-0.4cm)$)
    -- 
    (rolloutbox.west);

\node[
    frozenie,
    anchor=west
] (IEbottom)
at ($(rolloutbox.east)+(0.20cm,0)$)
{IE};

\draw[forward]
    (rolloutbox.east)
    --
    (IEbottom.west);

\node[
    circle,
    fill=black,
    inner sep=0.7pt
] (RLPortalLower)
at ($(IEbottom.south)+(0,-0.16cm)$) {};

\draw[
    connector,
    densely dotted
]
(IEbottom.south) -- (RLPortalLower.north);

\node[
    frozenie,
    anchor=west
] (IEtop)
at ($(IEbottom.west)+(0,1.3cm)$)
{IE};

\node[
    rolloutbox,
    anchor=east,
    minimum width=0.85cm,
    minimum height=0.42cm
] (targetbox)
at ($(IEtop.west)+(-0.20cm,0)$) {};

\node[
    branchtitle,
    anchor=south
]
at ($(targetbox.north)+(0,0.03cm)$)
{Ground Truth};

\node[
    img,
    anchor=center
] (rltarget)
at (targetbox.center)
{%
    \includegraphics[
        width=0.8cm,
        height=0.8cm
    ]{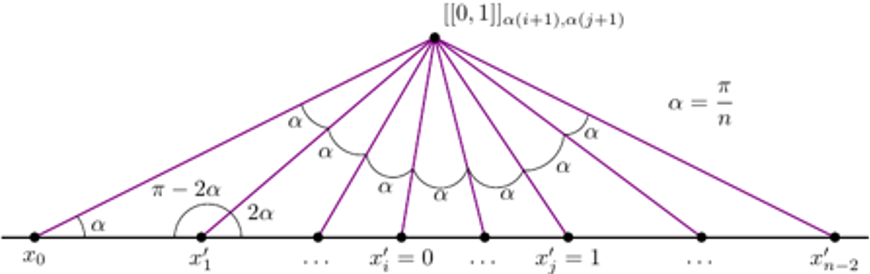}%
};

\draw[forward]
    (targetbox.east)
    --
    (IEtop.west);

\node[
    circle,
    fill=black,
    inner sep=0.7pt
] (RLPortalUpper)
at ($(IEtop.south)+(0,-0.16cm)$) {};

\draw[
    connector,
    densely dotted
]
(IEtop.south) -- (RLPortalUpper.north);

\node[
    font=\bfseries\tiny
] (Rsim)
at ($(IEtop.east)!0.5!(IEbottom.east)+(0.7cm,0)$)
{\(\mathcal{R}_{\mathrm{SSim}}\)};

\draw[forward]
    (IEtop.east)
    to[out=10,in=150]
    ($(Rsim.west)+(0.1cm,0)$);

\draw[forward]
    (IEbottom.east)
    to[out=-10,in=210]
    ($(Rsim.west)+(0.1cm,0)$);

\node[
    judgevlm,
    anchor=west
] (VLMJudge)
at ($(IEbottom.south)+(1.4,-0.3cm)$) {Judge};

\draw[forward]
    ($(rolloutbox.east)+(0.0,-0.65cm)$)
    -- 
    ($(VLMJudge.west)+(0.0,-0.15cm)$);

\node[
    star,
    fill=black,
    inner sep=0.7pt
] (JudgePortalTwo)
at ($(VLMJudge.west)+(-0.2,0.15cm)$) {};

\draw[
    connector,
    densely dotted
]
(JudgePortalTwo.east) -- ($(VLMJudge.west)+(0.0,0.15cm)$);

\node[
    prompt,
    anchor=north,
    text width=1.45cm,
    minimum height=0.85cm,
    font=\fontsize{3}{3.5}\selectfont
] (judgeprompt)
at ($(VLMJudge.north)+(0.3,1.8cm)$)
{You are a strict visual reward judge for scientific figure editing. \\
You are given two images: \\
- Image 1 is the source figure before editing. \\
- Image 2 is the predicted figure after editing. \\
You are also given a list of requested atomic edits. \\
Judge whether Image 2 applies each requested atomic edit relative to Image 1...};

\draw[forward]
    ($(judgeprompt.south)+(-0.3cm,0)$)
    to
    (VLMJudge.north);

\node[
    font=\bfseries\tiny
] (Rjudge)
at ($(VLMJudge.east)+(0.59cm,0)$)
{\(\mathcal{R}_{\mathrm{AE}}\)};

\draw[forward]
    (VLMJudge.east)
    to
    ($(Rjudge.west)+(0.1cm,0)$);

\end{tikzpicture}
\end{adjustbox}

\caption{Two-stage training pipeline. Left: Multi-task SFT jointly trains on equal amounts of editing (DaEdiTikZ) and reconstruction (DaTikZ-V4) data. Right: GDPO optimizes on a disjoint DaEdiTikZ subset using SelfSim from the frozen SFT vision encoder and instruction-following from a VLM judge.}
\label{fig:method_figure}

\end{figure*}

\paragraph{Multi-Reward Optimization with GDPO} 
$\mathcal{R}_{\mathrm{SSim}}$ provides dense target-similarity feedback, whereas $\mathcal{R}_{\mathrm{IF}}$ measures discrete atomic edit application. Since standard multi-reward GRPO aggregates rewards before group normalization, its learning signal is sensitive to their distributions. We instead use Group reward-Decoupled Normalization Policy Optimization (GDPO)~\citep{liu2026gdpogrouprewarddecouplednormalization}, which normalizes each reward independently before aggregation. For $G$ rollouts, GDPO computes:
\begin{equation}
A_m^{(i,j)}
=
\frac{
r_m^{(i,j)}-\operatorname{mean}_{j'}[r_m^{(i,j')}]
}{
\operatorname{std}_{j'}[r_m^{(i,j')}]+\varepsilon
},
\qquad
A_{\mathrm{sum}}^{(i,j)}
=
\sum_{m\in\mathcal{M}} w_m A_m^{(i,j)}
\label{eq:gdpo_reward_adv}
\end{equation}
Following GDPO, we normalize the aggregated advantages across the batch and optimize the clipped policy objective:
\begin{align}
\mathcal{J}_{\mathrm{GDPO}}(\theta)
=
\mathbb{E}_{x_i \sim \mathcal{D}_{\mathrm{edit}}}
\Bigg[
\frac{1}{G}
\sum_{j=1}^{G}
\frac{1}{L}
\sum_{t=1}^{|o_{i,j}|}
\min\Bigg(
&\frac{
\pi_{\theta}
\!\left(\hat{y}_{i,j,t}\mid x_i,\hat{y}_{i,j}^{<t}\right)
}{
\pi_{\theta_{\mathrm{old}}}
\!\left(\hat{y}_{i,j,t}\mid x_i,\hat{y}_{i,j}^{<t}\right)
}
 \widehat{A}_{\mathrm{sum}}^{(i,j)}, \nonumber\\
&\hspace{-15em}\operatorname{clip}\!\Bigl(
\frac{
\pi_{\theta}
\!\left(\hat{y}_{i,j,t}\mid x_i,\hat{y}_{i,j}^{<t}\right)
}{
\pi_{\theta_{\mathrm{old}}}
\!\left(\hat{y}_{i,j,t}\mid x_i,\hat{y}_{i,j}^{<t}\right)
},
1-\epsilon_{\mathrm{low}},
1+\epsilon_{\mathrm{high}}
\Bigr)
\widehat{A}_{\mathrm{sum}}^{(i,j)}
\Bigg) \nonumber 
  - \beta\,D_{\text{KL}}\!\big(p_\theta \,\|\, p_{\theta_{\text{SFT}}}\big)
\Bigg]
\label{eq:gdpo_objective}
\end{align}
Implementation details are provided in the Appendix~\ref{subsec:Method}.

\section{Experiments}
\label{sec:Experiments}

\paragraph{Setup}
We use disjoint group-level splits, reserving 27K DaEdiTikZ trajectories for RL and using the remaining 754K editing trajectories together with 754K DaTikZ-V4 reconstruction samples for SFT (1.51M instances total). Thus, figures from the same group never occur across training stages. SFT updates all parameters, whereas RL updates only the language model while freezing the vision encoder and embeddings. Unless stated otherwise, evaluation uses the 690 human-refined DaEdiTikZ-Bench instances, which are disjoint from all training groups.

\paragraph{Models}
We evaluate six proprietary VLMs---GPT-5.6-Sol, GPT-5.5, GPT-5.4, Gemini-3.1-Pro, Gemini-3.6-Flash, and Gemini-3.5-Flash---and eight open-source VLMs: Qwen3.6-27B\footnote{\href{https://openai.com/index/gpt-5-6/}{GPT-5.6-Sol}, \href{https://openai.com/index/introducing-gpt-5-5/}{GPT-5.5}, \href{https://openai.com/index/introducing-gpt-5-4/}{GPT-5.4}, \href{https://deepmind.google/models/model-cards/gemini-3-1-pro/}{Gemini 3.1 Pro}, \href{https://deepmind.google/models/model-cards/gemini-3-6-flash/}{Gemini 3.6 Flash}, \href{https://deepmind.google/models/model-cards/gemini-3-5-flash/}{Gemini 3.5 Flash}, \href{https://qwen.ai/blog?id=qwen3.6-27b}{Qwen3.6-27B}}, Qwen3.5 (27B, 9B, and 4B)~\citep{qwen3.5}, Qwen3-VL (8B and 4B)~\citep{bai2025qwen3vltechnicalreport}, and Qwen2.5-VL (7B and 3B)~\citep{bai2025qwen25vltechnicalreport}. We apply SFT to all models up to 9B parameters except Qwen2.5-VL-7B, yielding our EdiTikZ family. Subscripts distinguish earlier Qwen generations. RL is applied to EdiTikZ-4B and EdiTikZ-9B, denoted EdiTikZ-4B-RL and EdiTikZ-9B-RL.

\paragraph{Metrics}
We evaluate code similarity with TeX Edit Distance (TED)~\citep{kusner2015icml-word} and perceptual similarity with DreamSim (DSim)~\citep{NEURIPS2023_DreamSim}. Following VLM-based evaluation~\citep{ku-etal-2024-viescore}, GPT-5.5 scores three editing-specific criteria: (i) Edit Application (EA), measuring correct application of requested edits; (ii) Source Preservation (SP), measuring preservation of unaffected content; and (iii) Visual Quality (VQ), measuring legibility and publication readiness. Scores are produced on a 0--10 scale and normalized to $[0,1]$. We also report compilation rate (CR) and average output tokens (AT). The aggregate score (Avg) averages $1-\mathrm{TED}$, DSim, EA, SP, and VQ.

\section{Results}
\label{sec:Results}

\paragraph{Automatic Evaluation}
Across all architectures, SFT improves Avg by 0.186--0.363 and compilation rate by 19.0--39.3 percentage points. RL further improves EdiTikZ-4B/9B to 0.674/0.726 Avg. EdiTikZ-4B-RL reaches proprietary-level performance, while EdiTikZ-9B-RL achieves the highest overall score (Table~\ref{tab:editikz_main}). Additional results are in Appendix~\ref{subsec:Results}.

\subparagraph{Model Rankings Reverse after SFT}
Qwen3.5-4B/9B initially underperform Qwen3-VL-4B/8B (0.249/0.345 vs.\ 0.314/0.354 Avg), but surpass them after SFT (0.612/0.643 vs.\ 0.538/0.540), showing that base editing performance does not necessarily reflect task-specific adaptation potential.

\subparagraph{Visual Correctness vs.\ Code Similarity}
Unlike prior TikZ-generation RL, where TED improves after RL~\citep{greisinger2026tikzilla,zeng2026davinci}, ours worsens despite consistent gains across rendered metrics. We hypothesize that editing weakens visual--code coupling, as visually equivalent edits may differ at the code level.

\begin{table*}[t]
\centering
\caption{Results on DaEdiTikZ-Bench. \textbf{Bold} is best while \underline{underline} is second-best.}
\label{tab:editikz_main}
\small
\setlength{\tabcolsep}{5pt}
\renewcommand{\arraystretch}{0.95}

\begin{tabular}{lccccc:c:cc}
\toprule

\textbf{Model} &
\textbf{TED}$\downarrow$ &
\textbf{DSim}$\uparrow$ &
\textbf{EA}$\uparrow$ &
\textbf{SP}$\uparrow$ &
\textbf{VQ}$\uparrow$ &
\textbf{Avg}$\uparrow$ &
\textbf{CR}$\uparrow$ &
\textbf{AT}$\downarrow$ \\
\midrule

GPT-5.6-Sol
& 0.764 & 0.796 & 0.735 & 0.775 & 0.823
& \cellcolor{gray!32}0.673
& 88.3\% & 485 \\

GPT-5.5
& 0.765 & 0.829 & \underline{0.744} & 0.790 & \underline{0.849}
& \cellcolor{gray!33}0.689
& 92.2\% & 488 \\

GPT-5.4
& 0.763 & 0.741 & 0.674 & 0.706 & 0.762
& \cellcolor{gray!25}0.624
& 84.6\% & 487 \\

Gemini-3.1-Pro
& 0.716 & 0.795 & \textbf{0.761} & \underline{0.798} & 0.828
& \cellcolor{gray!34}\underline{0.693}
& 86.5\% & \textbf{384} \\

Gemini-3.6-Flash
& 0.740 & 0.665 & 0.654 & 0.677 & 0.698
& \cellcolor{gray!24}0.591
& 72.2\% & \underline{399} \\

Gemini-3.5-Flash
& 0.737 & 0.676 & 0.656 & 0.678 & 0.718
& \cellcolor{gray!24}0.598
& 74.0\% & 427 \\

\midrule

Qwen3.6-27B
& 0.768 & 0.675 & 0.470 & 0.521 & 0.635
& \cellcolor{gray!20}0.507
& 79.0\% & 547 \\

Qwen3.5-27B
& 0.757 & 0.677 & 0.482 & 0.524 & 0.636
& \cellcolor{gray!20}0.512
& 79.0\% & 490 \\

Qwen2.5-VL-7B
& 0.797 & 0.388 & 0.139 & 0.158 & 0.296
& \cellcolor{gray!7}0.237
& 50.6\% & 689 \\

\midrule

Qwen2.5-VL-3B
& 0.810 & 0.329 & 0.062 & 0.062 & 0.213
& \cellcolor{gray!3}0.171
& 45.9\% & 747 \\

EdiTikZ-3B
& 0.707 & 0.700 & 0.309 & 0.332 & 0.525
& \cellcolor{gray!19}0.432
& 82.2\% & 623 \\

\midrule

Qwen3-VL-4B
& 0.788 & 0.494 & 0.232 & 0.245 & 0.386
& \cellcolor{gray!12}0.314
& 61.9\% & 651 \\

EdiTikZ-4B\textsubscript{Qwen3}
& 0.644 & 0.790 & 0.445 & 0.474 & 0.627
& \cellcolor{gray!24}0.538
& 89.3\% & 567 \\

\midrule

Qwen3-VL-8B
& 0.772 & 0.543 & 0.281 & 0.293 & 0.427
& \cellcolor{gray!13}0.354
& 67.2\% & 509 \\

EdiTikZ-8B
& 0.676 & 0.765 & 0.462 & 0.509 & 0.639
& \cellcolor{gray!22}0.540
& 86.2\% & 579 \\

\midrule

Qwen3.5-4B
& 0.806 & 0.411 & 0.146 & 0.188 & 0.304
& \cellcolor{gray!9}0.249
& 51.4\% & 726 \\

EdiTikZ-4B
& \underline{0.629} & 0.813 & 0.552 & 0.609 & 0.714
& \cellcolor{gray!28}0.612
& 90.7\% & 542 \\

EdiTikZ-4B-RL
& 0.642 & \underline{0.871} & 0.633 & 0.706 & 0.803
& \cellcolor{gray!28}0.674
& \underline{95.2\%} & 494 \\

\midrule

Qwen3.5-9B
& 0.781 & 0.523 & 0.241 & 0.311 & 0.430
& \cellcolor{gray!13}0.345
& 64.1\% & 599 \\

EdiTikZ-9B
& \textbf{0.628} & 0.834 & 0.598 & 0.658 & 0.753
& \cellcolor{gray!30}0.643
& 92.0\% & 545 \\

EdiTikZ-9B-RL
& 0.676 & \textbf{0.893} & 0.734 & \textbf{0.815} & \textbf{0.865}
& \cellcolor{gray!35}\textbf{0.726}
& \textbf{96.8\%} & 488 \\

\bottomrule
\end{tabular}
\end{table*}

\begin{figure}[h]
  \centering
  \includegraphics[width=0.85\textwidth]{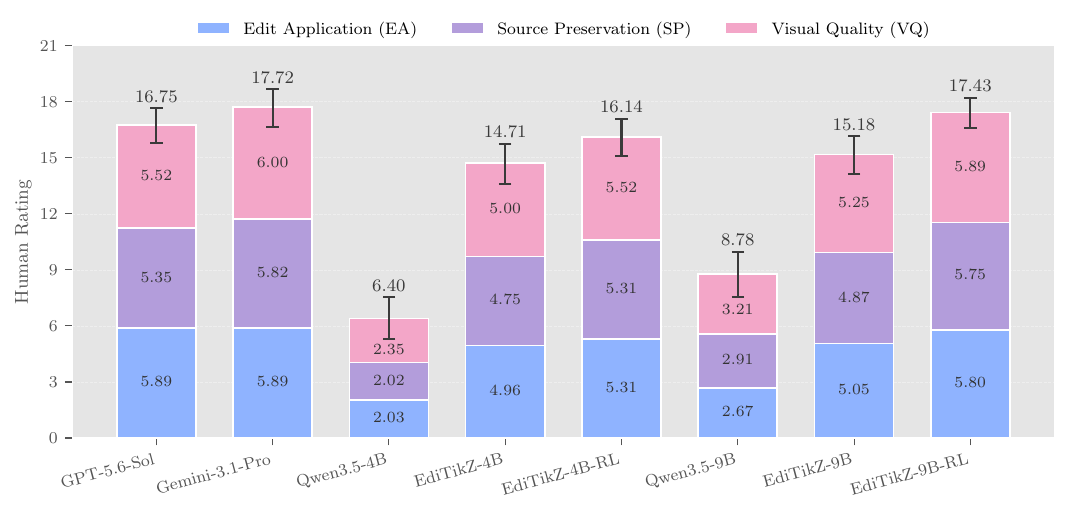}
  \caption{Likert-scale (1-7) across three evaluation criteria (EA, SP, and VQ) with 95\% confidence intervals for eight VLMs (4 baseline, 4 fine-tuned).}
  \label{fig:human_evaluation}
\end{figure}

\paragraph{Human Evaluation}
We conduct a human evaluation with 9 qualified annotators, who rate predictions from eight models on EA, SP, and VQ using a 1--7 Likert scale (Figure~\ref{fig:human_evaluation}). Each annotator evaluates 20 randomized figure groups with 10\% overlap, yielding 4,320 ratings. Quadratic-weighted agreement is high ($\kappa_{\mathrm{EA}}=0.801$, $\kappa_{\mathrm{SP}}=0.786$, $\kappa_{\mathrm{VQ}}=0.832$).

\subparagraph{Human Evaluation Confirms Post-Training Gains}
SFT raises the combined score of Qwen3.5-4B/9B from 6.40/8.78 to 14.71/15.18, with RL further improving it to 16.14/17.43, with gains across all three criteria. EdiTikZ-9B-RL nearly matches Gemini-3.1-Pro (17.43 vs.\ 17.72) and performs above GPT-5.6-Sol (16.75). SFT narrows the 4B--9B gap from 2.38 to 0.47 points, whereas RL widens it to 1.29 points.

\subparagraph{Automatic Metrics Align with Humans}
Our aggregate metric correlates strongly with combined human judgments ($\rho=0.823$). While TED correlates poorly ($\rho=0.374$), DSim and the criterion-specific EA, SP, and VQ metrics each reach $\rho\approx0.77$. $\mathcal{R}_{\mathrm{SSim}}$ correlates more strongly with human SP/VQ, whereas adding $\mathcal{R}_{\mathrm{IF}}$ improves EA correlation by 0.045 and raises overall correlation from 0.812 to 0.827, showing the intended complementarity.

\paragraph{Ablations: Data Mixtures}
Table~\ref{tab:editikz_ablation_data_mixture_new} compares reconstruction and editing mixtures on two VLMs. Joint training performs best for 3B (0.432 Avg vs.\ 0.392 editing-only) and matches sequential training for 8B (0.539/0.540), with both exceeding editing-only (0.531). Thus, reconstruction consistently improves editing, while joint training additionally retains both capabilities.

\paragraph{Ablations: Rewards and GDPO}
Table~\ref{tab:editikz_ablation_rl_new} ablates our rewards and optimization algorithm. $\mathcal{R}_{\mathrm{IF}}$ outperforms $\mathcal{R}_{\mathrm{SSim}}$, by +0.031 EA. Combining both with GRPO adds +0.009 Avg, while GDPO increases this gain to +0.038, supporting independent normalization of the complementary rewards.

\begin{table*}[t]
\centering
\tiny
\setlength{\tabcolsep}{1.2pt}
\renewcommand{\arraystretch}{0.92}

\begin{minipage}[t]{0.53\textwidth}
\centering
\caption{Ablation on DaEdiTikZ-Bench for data-mixing strategies on two VLMs.}
\label{tab:editikz_ablation_data_mixture_new}

\begin{tabular}{clccccc:c:cc}
\toprule
\textbf{VLM} &
\textbf{Variant} &
\textbf{TED}$\downarrow$ &
\textbf{DSim}$\uparrow$ &
\textbf{EA}$\uparrow$ &
\textbf{SP}$\uparrow$ &
\textbf{VQ}$\uparrow$ &
\textbf{Avg}$\uparrow$ &
\textbf{CR}$\uparrow$ &
\textbf{AT}$\downarrow$ \\
\midrule

\multirow{5}{*}{%
\rotatebox{90}{%
\fontsize{4.5}{5.1}\selectfont
\textbf{Qwen2.5-VL-3B}%
}%
}
& Base
& 0.810 & 0.329 & 0.062 & 0.062 & 0.213
& \cellcolor{gray!3}0.171
& 45.9\% & 747 \\

& Only Recon
& 0.751 & 0.615 & 0.016 & 0.277 & 0.426
& \cellcolor{gray!16}0.317
& 68.4\% & 800 \\

& Only Edit
& 0.741 & 0.626 & \underline{0.296} & \underline{0.309}
& 0.472
& \cellcolor{gray!22}0.392
& 73.9\% & 723 \\

& Recon$\rightarrow$Edit
& \underline{0.720} & \underline{0.647} & 0.272 & 0.297
& \underline{0.480}
& \cellcolor{gray!22}\underline{0.395}
& \underline{76.2\%} & \underline{686} \\

& Recon+Edit
& \textbf{0.707} & \textbf{0.700} & \textbf{0.309} & \textbf{0.332}
& \textbf{0.525}
& \cellcolor{gray!26}\textbf{0.432}
& \textbf{82.2\%} & \textbf{623} \\

\midrule

\multirow{5}{*}{%
\rotatebox{90}{%
\fontsize{4.5}{5.1}\selectfont
\textbf{Qwen3-VL-8B}%
}%
}
& Base
& 0.772 & 0.543 & 0.281 & 0.293 & 0.427
& \cellcolor{gray!19}0.354
& 67.2\% & \textbf{509} \\

& Only Recon
& 0.709 & 0.737 & 0.157 & 0.495 & 0.592
& \cellcolor{gray!28}0.454
& 85.1\% & 646 \\

& Only Edit
& 0.713 & 0.785 & 0.451 & 0.493 & 0.638
& \cellcolor{gray!34}0.531
& \textbf{90.6\%} & \underline{557} \\

& Recon$\rightarrow$Edit
& \underline{0.685} & \textbf{0.769} & \underline{0.459} & \textbf{0.512}
& \textbf{0.640}
& \cellcolor{gray!35}\underline{0.539}
& \underline{87.1\%} & 593 \\

& Recon+Edit
& \textbf{0.676} & \underline{0.765} & \textbf{0.462} & \underline{0.509}
& \underline{0.639}
& \cellcolor{gray!35}\textbf{0.540}
& 86.2\% & 579 \\

\bottomrule
\end{tabular}
\end{minipage}
\hfill
\begin{minipage}[t]{0.45\textwidth}
\centering
\caption{Ablation on DaEdiTikZ-Bench for reward functions and algorithms using EdiTikZ-9B.}
\label{tab:editikz_ablation_rl_new}

\setlength{\tabcolsep}{1.2pt}
\begin{tabular}{lccccc:c:cc}
\toprule
\textbf{Variant} &
\textbf{TED}$\downarrow$ &
\textbf{DSim}$\uparrow$ &
\textbf{EA}$\uparrow$ &
\textbf{SP}$\uparrow$ &
\textbf{VQ}$\uparrow$ &
\textbf{Avg}$\uparrow$ &
\textbf{CR}$\uparrow$ &
\textbf{AT}$\downarrow$ \\
\midrule

Post-SFT
& \textbf{0.628} & 0.834 & 0.598 & 0.658
& 0.753
& \cellcolor{gray!3}0.643
& 92.0\% & 545 \\

\midrule

$\mathcal{R}_{\mathrm{SSim}}$
& 0.667 & 0.872 & 0.633 & 0.706
& 0.805
& \cellcolor{gray!13}0.670
& 96.1\% & 489 \\

$\mathcal{R}_{\mathrm{IF}}$
& 0.671 & 0.866 & 0.664 & 0.721
& 0.814
& \cellcolor{gray!17}0.679
& 95.2\% & \underline{488} \\

Both (w. GRPO)
& \underline{0.662} & \underline{0.879}
& \underline{0.671} & \underline{0.728}
& \underline{0.822}
& \cellcolor{gray!20}\underline{0.688}
& \underline{96.4\%} & \textbf{476} \\

Both (w. GDPO)
& 0.676 & \textbf{0.893}
& \textbf{0.734} & \textbf{0.815}
& \textbf{0.865}
& \cellcolor{gray!35}\textbf{0.726}
& \textbf{96.8\%} & \underline{488} \\

\bottomrule
\end{tabular}
\end{minipage}

\end{table*}

\begin{figure}[h]
  \centering
  \includegraphics[width=0.75\textwidth]{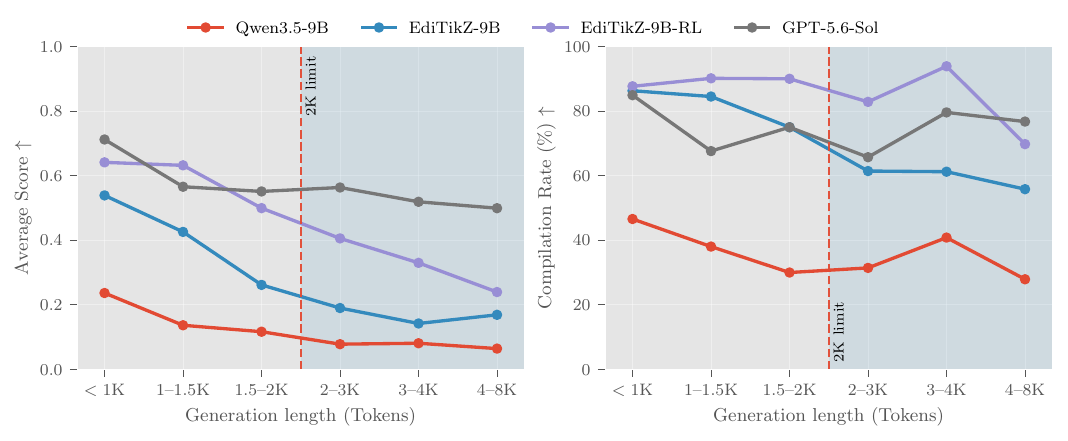}
  \caption{OOD performance on SPIQA and CharXiv combined by generation length. Average score is $(\mathrm{EA}+\mathrm{SP}+\mathrm{VQ})/3$. The red dashed line marks the 2K-token post-training limit.}
  \label{fig:ood_evaluation}
\end{figure}

\paragraph{Generalization under Severe Distribution Shift}
We stress-test EdiTikZ on SPIQA~\cite{NEURIPS2024_SPIQA} and CharXiv~\cite{NEURIPS2024_CharXiv}, which contain complex architectural diagrams, multi-panel plots, tables, schematics, and charts across diverse scientific domains, produced with tools such as Matplotlib, MATLAB, DrawIO, ggplot, and Plotly rather than TikZ. We sample 300 SPIQA and 600 CharXiv figures and manually remove those requiring external data, leaving 190 and 497 instances, respectively. Since neither dataset provides editing pairs, we use GPT-5.6-Sol to generate synthetic edit instructions. We then evaluate model predictions reference-free using EA, SP, VQ, CR, and AT. OOD generations require $3$--$5\times$ more tokens than DaEdiTikZ-Bench and frequently exceed the 2K-token completion limit used during post-training. We stratify examples by mean generation length across all four models (Figure~\ref{fig:ood_evaluation}), ensuring identical examples within each bin.

\subparagraph{RL Gains Increase with Difficulty}
For short generations ($<$1K), SFT provides most of the gain over Qwen3.5-9B. With increasing length, SFT gains diminish while the additional benefit from RL grows, dominating from 1.5--4K tokens. RL also maintains $>$80\% compilation through 3--4K, consistently exceeding GPT-5.6-Sol, whereas SFT compilation degrades steadily.

\subparagraph{Competitive within the Training Horizon}
Within the trained $\leq$2K regime, EdiTikZ-9B-RL remains close to GPT-5.6-Sol, with $<$0.1 Avg difference across all bins. Beyond 2K, EdiTikZ degrades faster and the gap widens, although SFT and RL gains persist throughout the 2--8K regime.

\section{Conclusion, Limitations, and Future Work}
\label{sec:Conclusion, Limiatations, and Future Work}

We introduced a scalable framework for recovering naturally occurring scientific-figure revisions from arXiv, GitHub, and TeX SE, instantiated as DaEdiTikZ, a large-scale real-world TikZ editing dataset. We further introduced the human-refined DaEdiTikZ-Bench and EdiTikZ, a family of 4--9B models trained with multi-task SFT and multi-reward RL. EdiTikZ-9B-RL leads automatic evaluation and reaches comparable human ratings to the strongest proprietary system. Post-training gains transfer to substantially more complex SPIQA and CharXiv figures, even beyond the 2K-token training horizon. Overall, naturally occurring revision trajectories provide effective supervision for small, open scientific-figure editing models competitive with much larger proprietary systems.

DaEdiTikZ inherits noise from automatically inferred instructions, including omissions and misinterpretations despite filtering and code grounding. Performance also degrades for long OOD generations, motivating post-training on more complex figures in the future. Evaluation in this regime is itself limited by synthetic instructions and potentially less reliable reference-free judging. Beyond these limitations, our visualization-language-agnostic revision-mining framework could extend to Matplotlib or LaTeX tables, while access to source programs could enable localized editing without full reconstruction. Revision trajectories could further support comparative VQA, retrieval, and representation learning, while helping to unify generation and editing within general-purpose scientific visualization models.

\section*{AI Use Statement}
We did not use AI to design or refine the dataset and methodological approaches, to search for relevant literature, to provide feedback on experiments, or to analyze data or interpret results. Generative AI tools were used to create editing instructions for the construction of the dataset, the benchmark, and the out-of-distribution dataset, as explicitly described in Section~\ref{sec:Dataset and Benchmark}. In addition, we used generative AI tools to improve the grammar, wording, and readability of the manuscript (including figures, tables, and equations) as well as to assist with debugging and editing the code written by the authors. All AI-assisted work was reviewed and verified by the authors. We therefore take responsibility for the final content of this work, including text, statements, code, and artifacts created with the help of generative AI.

\bibliography{iclr2027_conference}
\bibliographystyle{plainnat}

\appendix

\section{Appendix}
\label{sec:Appendix}

\subsection{Related Work}
\label{subsec:Related Work}

\subsubsection{Image Editing}
\label{subsubsec:Image Editing}
Most image-editing progress has been driven by diffusion models~\citep{neurips2020-ddpm}. Early methods preserve source structure through attention manipulation~\citep{mokady2022null}, while InstructPix2Pix~\citep{Brooks2022InstructPix2PixLT} generates editing supervision by using GPT-3 for instructions and Stable Diffusion with Prompt-to-Prompt for paired before/after images. UltraEdit~\citep{zhao2024ultraeditinstructionbasedfinegrainedimage} instead anchors automatically generated edits in real photographs and artworks and adds region-level supervision. AnyEdit~\citep{Yu2024AnyEditMU} expands this setting to over 20 edit types with adaptive editing and automatic result selection. SeedEdit~\citep{wang2025seededit30fasthighquality} combines synthetic pairs with professional editing workflows, traditional operators, and video-derived image pairs, while ScaleEdit-12M~\citep{chen2026scaleedit12mscalingopensourceimage} scales data generation through an open-source multi-agent pipeline with adaptive synthesis and task-aware verification. These methods primarily target natural images rather than structurally constrained scientific graphics.

\subsection{Dataset and Benchmark}
\label{subsec:Dataset and Benchmark}

\subsubsection{Inferring Edit Instructions}
\label{subsubsec:Inferring Edit Instructions}
For large-scale edit instruction inference, we use Qwen3.6-27B (non-thinking) conditioned jointly on image pair and TikZ code. Temperature is 0.1, top\_p is 1.0, and output tokens are set to 1024. On $4\times$ NVIDIA H100 (94 GB) GPUs with the vLLM~\citep{kwon2023efficient} framework, this took 11 days. The prompt is in Figure~\ref{fig:prompt_edit_instructions}.
\refstepcounter{figure}
\begin{PromptBox}{VLM Instruction Generation}
Your task is to analyze the differences between two scientific figures given as: \\
- Image 1 + TikZ Code 1 \\
- Image 2 + TikZ Code 2 \\
Provide detailed descriptions of the edits needed to transform the first scientific figure into the second. \\

Requirements: \\

0) Pair quality: Before listing any edits, decide whether Image 2 is a plausible edited version of Image 1. \\
- \texttt{"ok"}: same underlying figure or scene with a plausible edit path. \\
- \texttt{"identical"}: no visible differences and no code changes implying visible differences. \\
- \texttt{"invalid"}: no plausible edit path or no shared figure identity (e.g., different figure type, different number of panels, different main subject, or no shared anchors). \\

1) Code-first, render-grounded: Use the TikZ code to discover candidate differences between the figures. Only keep differences that produce a perceptible change in the rendered images. Describe edits using human-visible anchors from the images (rendered text, mathematics, relative location, appearance, connectivity), not code-only identifiers or absolute coordinates. \\

2) Atomic and reconstructable: Each edit must describe exactly one logical change and include enough concrete before$\rightarrow$after detail that a human, given Image 1 alone, could plausibly recreate the corresponding part of Image 2. If many elements change in the same way, group them into one global edit, but enumerate the specific visible changes. \\

Output requirements: \\
- Report ONLY differences (no full image or code descriptions). \\
- Output ONLY valid JSON (no extra text and no trailing commas). \\

Return exactly the following JSON format: \\
{\ttfamily
\{ \\
\hspace*{1.5em}"pair\_quality": "ok|identical|invalid", \\
\hspace*{1.5em}"edits": [ \\
\hspace*{3em}\{ \\
\hspace*{4.5em}"intent": "add|remove|modify", \\
\hspace*{4.5em}"operation": "text|annotation|geometry|style|data|structure|other", \\
\hspace*{4.5em}"detailed\_change": "specific human-visible description" \\
\hspace*{3em}\} \\
\hspace*{1.5em}] \\
\} \\
}

If \texttt{pair\_quality} is \texttt{"identical"}, return exactly: \\
{\ttfamily
\{ "pair\_quality": "identical", "edits": [] \}
}

If \texttt{pair\_quality} is \texttt{"invalid"}, return exactly: \\
{\ttfamily
\{ "pair\_quality": "invalid", "edits": [] \}
} \\

Operation guidance: \\
- text: visible strings or mathematics (labels, titles, tick labels, axis names, legend entries). \\
- annotation: explanatory or highlighting elements (arrows, callouts, braces, highlight boxes, emphasis marks). \\
- geometry: position, shape, size, orientation, alignment, or spacing of visible elements. \\
- style: appearance changes that do not alter the encoded meaning (color, line style, thickness, opacity, font). \\
- data: changes to plotted or encoded values, including new points, curves, colormap normalization, or category-to-color mappings. \\
- structure: high-level organization while preserving the same underlying figure identity (e.g., added or removed panels, plot type changes of the same data, topology changes in the same diagram). \\
- other: visible changes not covered above (e.g., clipping, layer ordering, global transforms). \\

TikZ Code 1: \\
\{code\_1\} \\

TikZ Code 2: \\
\{code\_2\} \\

Your output JSON:
\label{fig:prompt_edit_instructions}
\end{PromptBox}

\clearpage

\begin{table*}[t]
\centering
\caption{Summary statistics for DaEdiTikZ. A retained pair has both directions with \texttt{pair\_quality=ok}. Each valid direction forms a separate editing trajectory.}
\label{tab:daeditikz_summary}
\setlength{\tabcolsep}{8pt}
\renewcommand{\arraystretch}{0.98}
\begin{tabular}{lr}
\toprule
\textbf{Statistic} & \textbf{Value} \\
\midrule
Total groups & 87,051 \\
Unique figures & 589,986 \\
Candidate figure pairs / trajectories & 430,442 / 860,884 \\
Bidirectional valid pairs / trajectories & 390,516 / 781,032 \\
Total atomic edits & 3,282,215 \\
Mean atomic edits per trajectory (median; P95 / P99) & 4.20 (4; 9 / 13) \\
Mean atomic edit length & 22.3 words (127.0 characters) \\
Mean trajectory length & 93.6 words (533.9 characters) \\
\bottomrule
\end{tabular}
\end{table*}
DaEdiTikZ connects 590K unique figures through 430K candidate pairs from 87K context groups. Pair validation retains 90.7\% of candidates, yielding 781K directional trajectories and 3.28M atomic edits. Figure reuse is limited. 69.3\% of figures occur in only one pair and 97.3\% in at most three (Table~\ref{tab:daeditikz_summary}).

\begin{table*}[t]
\centering
\caption{Directional response quality and pair-level retention of all 430,442 candidate pairs.}
\label{tab:daeditikz_retention}
\setlength{\tabcolsep}{7pt}
\renewcommand{\arraystretch}{0.98}
\begin{tabular}{llrr}
\toprule
\textbf{Level} & \textbf{Outcome} & \textbf{Count} & \textbf{Percentage} \\
\midrule
\multirow{4}{*}{Forward direction}
 & OK        & 401,357 & 93.24\% \\
 & Invalid   &  17,770 &  4.13\% \\
 & Identical &   4,151 &  0.96\% \\
 & Missing   &   7,164 &  1.66\% \\
\midrule
\multirow{4}{*}{Backward direction}
 & OK        & 400,892 & 93.13\% \\
 & Invalid   &  18,191 &  4.23\% \\
 & Identical &   4,243 &  0.99\% \\
 & Missing   &   7,116 &  1.65\% \\
\midrule
\multirow{2}{*}{Candidate pair}
 & Both directions valid  & 390,516 & 90.72\% \\
 & Excluded      &  39,926 &  9.28\% \\
\bottomrule
\end{tabular}
\end{table*}
Quality is nearly symmetric across directions, with 93.2\% of both forward and backward responses accepted. 390.5K pairs support supervision in both directions (Table~\ref{tab:daeditikz_retention}).

\clearpage

\begin{table*}[t]
\centering
\caption{Frequency and description length of atomic edits by intent and operation.}
\label{tab:daeditikz_edit_distribution}
\small
\setlength{\tabcolsep}{7pt}
\renewcommand{\arraystretch}{0.98}
\begin{tabular}{llrrrr}
\toprule
\textbf{Dimension} &
\textbf{Category} &
\textbf{Atomic edits} &
\textbf{Share} &
\textbf{Words/edit} &
\textbf{Characters/edit} \\
\midrule
\multirow{3}{*}{Intent}
& Add
& 418,547
& 12.75\%
& 19.31
& 107.47 \\
& Modify
& 2,515,332
& 76.64\%
& 23.53
& 134.14 \\
& Remove
& 348,167
& 10.61\%
& 16.65
& 99.19 \\
\midrule
\multirow{6}{*}{Operation}
& Annotation
& 372,974
& 11.36\%
& 20.24
& 117.69 \\
& Data
& 322,790
& 9.83\%
& 32.33
& 174.08 \\
& Geometry
& 657,043
& 20.02\%
& 25.90
& 147.39 \\
& Structure
& 246,587
& 7.51\%
& 28.38
& 164.97 \\
& Style
& 302,560
& 9.22\%
& 21.31
& 119.78 \\
& Text
& 1,380,250
& 42.05\%
& 17.84
& 103.68 \\
\bottomrule
\end{tabular}
\end{table*}
Most inferred atomic edits modify existing content (76.6\%), while additions and removals jointly account for 23.4\% (Table~\ref{tab:daeditikz_edit_distribution}). Text is the most frequent operation (42.1\%), followed by geometry (20.0\%), annotation (11.4\%), data (9.8\%), style (9.2\%), and structure (7.5\%). Description length also varies systematically with edit semantics. Data and structural changes require the longest descriptions, averaging 32.3 and 28.4 words per edit, whereas text edits average 17.8 words. Modifications are longer than additions and removals, consistent with the need to specify both an existing and a desired state. Manual inspection suggests that modifications are somewhat overrepresented, as the VLM occasionally describes additions or removals as replacements (e.g., replacing a label with nothing).

\begin{table*}[t]
\centering
\caption{Intent distribution conditioned on operation.}
\label{tab:daeditikz_intent_given_operation}
\setlength{\tabcolsep}{9pt}
\renewcommand{\arraystretch}{0.98}
\begin{tabular}{lrrr}
\toprule
\textbf{Operation} & \textbf{Add} & \textbf{Modify} & \textbf{Remove} \\
\midrule
Annotation & 41.21\% & 27.94\% & 30.85\% \\
Data       &  9.47\% & 81.18\% &  9.35\% \\
Geometry   & 10.05\% & 81.08\% &  8.88\% \\
Structure  & 23.52\% & 50.19\% & 26.28\% \\
Style      &  0.70\% & 98.49\% &  0.81\% \\
Text       &  7.84\% & 86.56\% &  5.61\% \\
\bottomrule
\end{tabular}
\end{table*}
Intent depends strongly on operation (Table~\ref{tab:daeditikz_intent_given_operation}). Text, geometry, data, and especially style edits predominantly modify existing content ($>$80\%). In contrast, 50.2\% of structure and 27.9\% of annotation edits modify an element. Structure is balanced across additions and removals (23.5\% vs. 26.3\%), whereas annotations include more additions (41.2\% vs. 30.9\%).

\clearpage

\begin{table*}[t]
\centering
\caption{Intent and Operation distribution across semantic similarity intervals.}
\label{tab:daeditikz_similarity_composition}
\scriptsize
\setlength{\tabcolsep}{3.2pt}
\renewcommand{\arraystretch}{0.95}
\resizebox{\textwidth}{!}{%
\begin{tabular}{lrrrrrrrrr}
\toprule
\multirow{2}{*}{\textbf{Similarity}} &
\multicolumn{3}{c}{\textbf{Intent}} &
\multicolumn{6}{c}{\textbf{Operation}} \\
\cmidrule(lr){2-4}\cmidrule(lr){5-10}
& \textbf{Add} & \textbf{Modify} & \textbf{Remove}
& \textbf{Annotation} & \textbf{Data} & \textbf{Geometry}
& \textbf{Structure} & \textbf{Style} & \textbf{Text} \\
\midrule
$[0.95,0.96)$ & 16.41 & 70.69 & 12.89 & 13.84 &  7.84 & 18.21 & 9.13 &  8.53 & 42.45 \\
$[0.96,0.97)$ & 14.97 & 72.94 & 12.09 & 12.79 &  8.60 & 19.06 & 8.82 &  8.55 & 42.17 \\
$[0.97,0.98)$ & 13.08 & 75.81 & 11.10 & 11.49 &  9.76 & 20.21 & 7.90 &  8.73 & 41.91 \\
$[0.98,0.99)$ & 10.04 & 80.90 &  9.06 &  9.37 & 11.75 & 21.82 & 6.19 &  9.29 & 41.58 \\
$[0.99,1.00]$ &  5.16 & 89.77 &  5.07 &  6.70 & 13.27 & 22.35 & 3.39 & 12.34 & 41.95 \\
\bottomrule
\end{tabular}%
}
\end{table*}
Lower-similarity pairs contain more additions and removals, whereas higher similarity pairs contain more modifications. Moreover, highly similar pairs primarily modify existing geometry, data, and style while annotation and structural changes correspond to lower similarity. Text remains stable at approximately 42\% across all intervals (Table~\ref{tab:daeditikz_similarity_composition}).

\begin{table*}[t]
\centering
\caption{Candidate distribution, quality, and edit magnitude across semantic similarity intervals.}
\label{tab:daeditikz_similarity_summary}
\small
\setlength{\tabcolsep}{5.2pt}
\renewcommand{\arraystretch}{0.98}
\begin{tabular}{lrrrrrrr}
\toprule
\textbf{Similarity} & \textbf{Pairs} & \textbf{Share} & \textbf{Retention}
& \textbf{Invalid} & \textbf{Identical} & \textbf{Edits/traj.} & \textbf{Words/edit} \\
\midrule
$[0.95,0.96)$ & 95,641 & 22.22\% & 81.91\% & 12.09\% & 0.04\% & 5.29 & 20.86 \\
$[0.96,0.97)$ & 83,942 & 19.50\% & 90.05\% &  5.24\% & 0.08\% & 4.96 & 21.39 \\
$[0.97,0.98)$ & 79,764 & 18.53\% & 94.47\% &  1.81\% & 0.12\% & 4.48 & 22.11 \\
$[0.98,0.99)$ & 80,295 & 18.65\% & 96.55\% & 0.47\% & 0.28\% & 3.82 & 23.23 \\
$[0.99,1.00]$ & 90,098 & 20.93\% & 92.91\% &  0.22\% & 4.19\% & 2.61 & 25.35 \\
\bottomrule
\end{tabular}
\end{table*}
The five intervals each contain between 18.5\% and 22.2\% of candidates in Table~\ref{tab:daeditikz_similarity_summary}. Mean edit count decreases monotonically from 5.29 to 2.61 as similarity increases, while description length rises from 20.9 to 25.4 words per edit. Bidirectional retention peaks at 96.6\% in $[0.98,0.99)$. Lower similarities increasingly produce non-plausible edit pairs, whereas the highest interval contains more identical pairs that differ only at the code level (e.g., through refactoring).

\clearpage

\begin{table*}[t]
\centering
\caption{Source-specific characteristics.}
\label{tab:daeditikz_source_characteristics}
\small
\setlength{\tabcolsep}{3.8pt}
\renewcommand{\arraystretch}{1.05}
\resizebox{\textwidth}{!}{%
\begin{tabular}{lrrrrrrp{3.7cm}p{6.8cm}}
\toprule
\textbf{Source} &
\textbf{Pairs} &
\textbf{Share} &
\textbf{Groups} &
\textbf{Retention} &
\textbf{Edits/traj.} &
\textbf{Words/edit} &
\textbf{Intent} &
\textbf{Operation} \\
\midrule
arXiv
& 402,411
& 93.49\%
& 79,246
& 90.90\%
& 4.23
& 22.26
& Modify 76.66\%, add 12.74\%, remove 10.59\%
& Text 42.34\%, geometry 20.02\%, annotation 11.29\%, data 9.78\%,
  style 9.09\%, structure 7.47\% \\
GitHub
& 21,891
& 5.09\%
& 3,032
& 87.98\%
& 4.12
& 21.69
& Modify 76.14\%, add 13.09\%, remove 10.77\%
& Text 40.38\%, geometry 17.51\%, annotation 12.64\%, data 11.31\%,
  style 9.49\%, structure 8.66\% \\
TeX SE
& 6,068
& 1.41\%
& 4,761
& 89.26\%
& 2.29
& 26.52
& Modify 77.11\%, add 11.49\%, remove 11.40\%
& Geometry 36.04\%, style 23.26\%, text 16.73\%, annotation 11.76\%,
  data 6.98\%, structure 5.23\% \\
\bottomrule
\end{tabular}%
}
\end{table*}
ArXiv supplies most trajectories and is primarily text-centered, whereas GitHub contains more annotation and data edits. TeX SE provides a distinct form of supervision. Its trajectories contain fewer atomic edits (2.3 versus 4.2) but require the most detailed instructions (26.5 words per edit versus 22). It predominantly involves geometric and stylistic over textual refinements (Table~\ref{tab:daeditikz_source_characteristics}).

\subsubsection{Dataset Quality Analysis}
\label{subsubsec:Dataset Quality Analysis}
\begin{figure}[h]
  \centering
  \includegraphics[width=\textwidth]{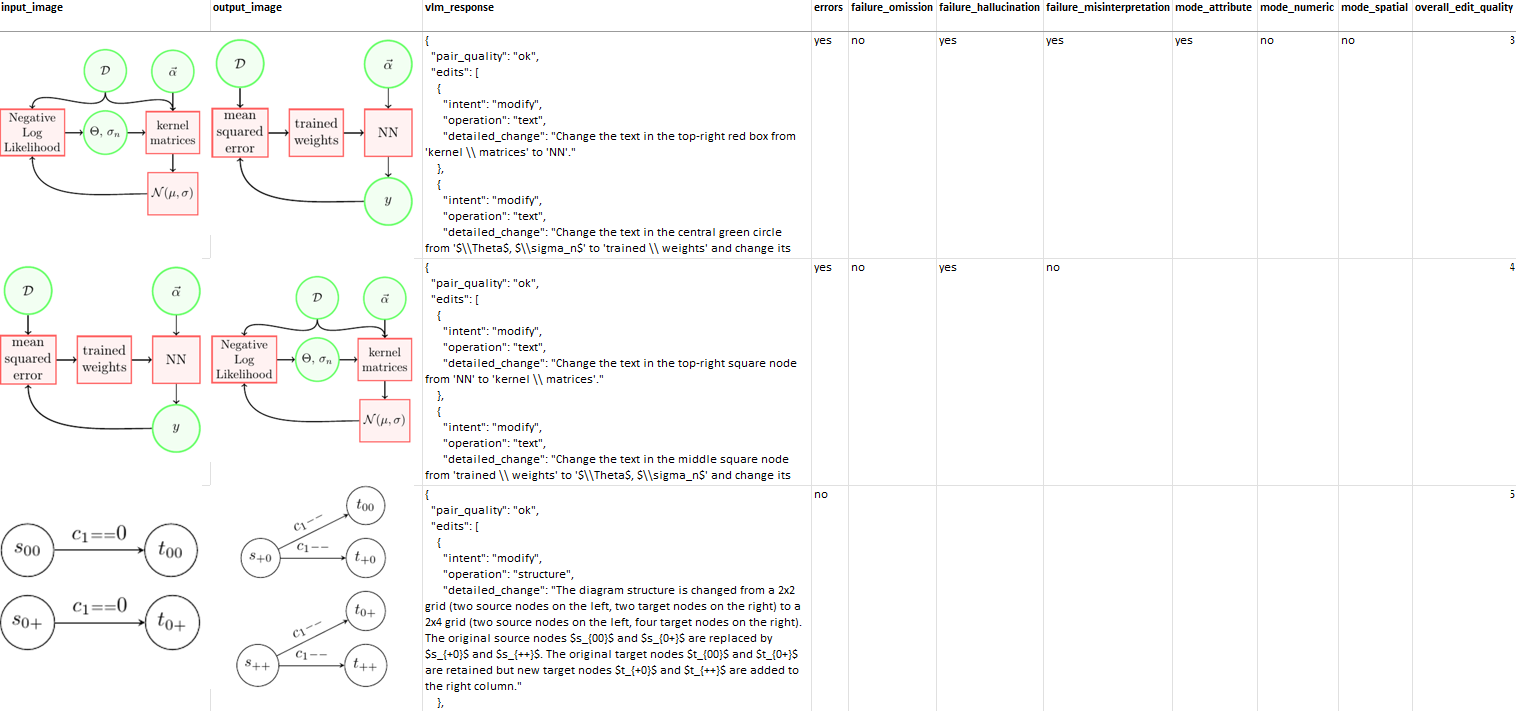}
  \caption{Screenshot of our excel sheet for evaluating the directional VLM-generated edit instructions.}
  \label{fig:vlm_instruction_sheet}
\end{figure}

Our dataset quality analysis involved one master's student and one PhD student. Both annotators completed the evaluation sheet in Figure~\ref{fig:vlm_instruction_sheet}.
The guidelines for completing our evaluation form are summarized in Table~\ref{tab:instruction_error_guidelines}.
\begin{table}[t]
\centering
\caption{Guidelines for annotating errors in VLM-generated edit instructions.}
\label{tab:instruction_error_guidelines}
\small
\setlength{\tabcolsep}{4pt}
\begin{tabular}{lp{0.72\linewidth}}
\toprule
\textbf{Category} & \textbf{Guideline} \\
\midrule
Error &
The instruction contains at least one of omission, hallucination, or misinterpretation. \\
\midrule
Omission &
A change is absent or only partially captured by the instruction. \\

Hallucination &
A change is specified for which no corresponding source-to-target change exists. \\

Misinterpretation &
A change is identified but describes
one or more of its properties incorrectly. \\
\midrule
Attribute &
An edited object's identity, appearance, style, text, shape, or other non-numeric property is described incorrectly. \\

Numeric &
A numerical value or quantitative change is described incorrectly (e.g., values, counts, dimensions, or magnitudes). \\

Spatial &
The spatial relation, position, orientation, direction, or arrangement of edited elements is described incorrectly. \\
\bottomrule
\end{tabular}
\end{table}

\subsubsection{DaEdiTikZ-Bench}
\label{subsubsec:DaEdiTikZ-Bench}
\begin{figure}[h]
  \centering
  \includegraphics[width=\textwidth]{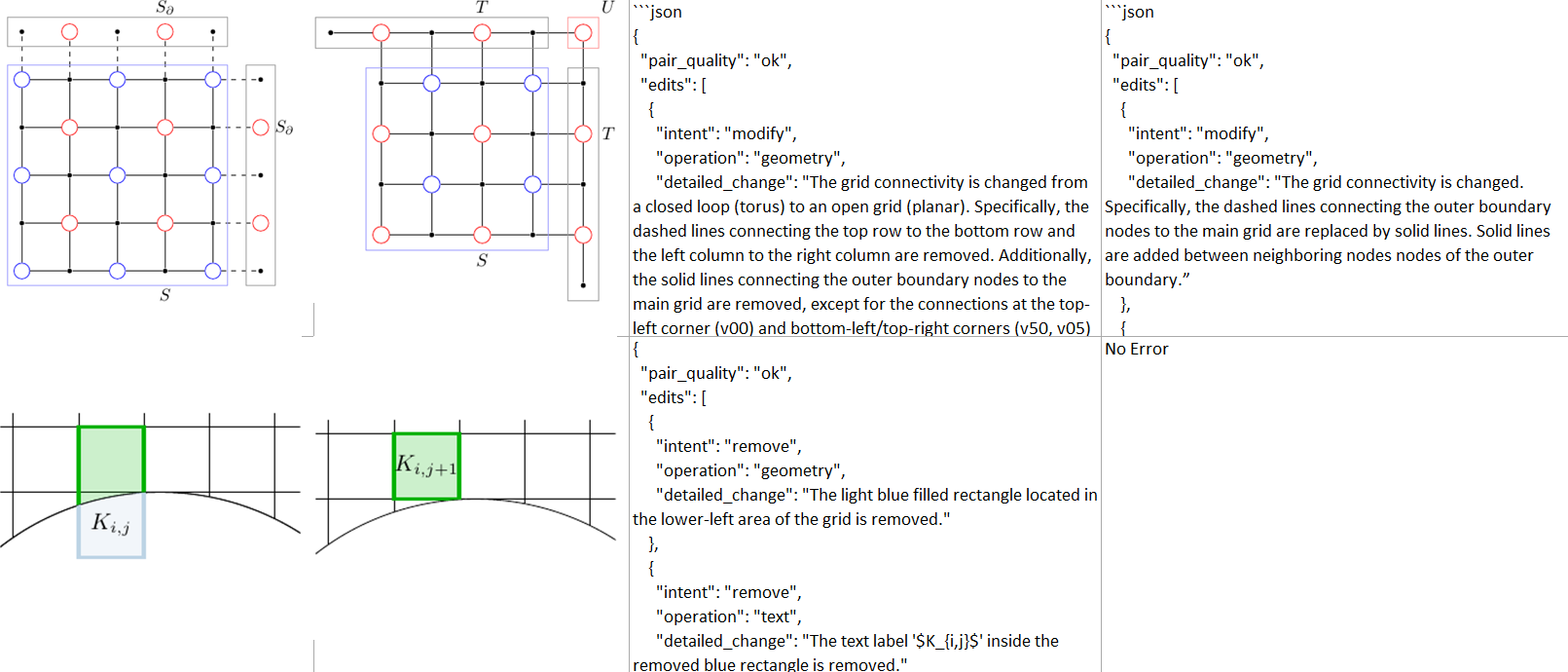}
  \caption{Screenshot of our excel sheet for correcting the directional VLM-generated edit instructions.}
  \label{fig:vlm_correction_sheet}
\end{figure}

Six annotators (four master’s students, one PhD student, one assistant professor) manually correct all 690 VLM-generated instructions from our benchmark. Similar to the dataset quality analysis, they were provided with the source image, target image, and the raw VLM response in JSON objects/entries. For omissions, they append another part of the JSON object (with \texttt{intent}, \texttt{operation}, and \texttt{detailed\_change}), where the missed change is described. For hallucination, the corresponding part of the JSON object is removed and misinterpretation keeps it but corrects the error. The correction sheet is in Figure~\ref{fig:vlm_correction_sheet}.

\subsection{Method}
\label{subsec:Method}

\subsubsection{Joint Reconstruction and Editing SFT}
\label{subsubsec:Joint Reconstruction and Editing SFT}
Figure~\ref{fig:prompt_editing} and~\ref{fig:prompt_reconstruction} present the prompts for joint editing and reconstruction SFT. The editing prompt is used across all training stages and evaluation of all models.
\refstepcounter{figure}
\begin{PromptBox}{Editing Prompt}
This is an image of a scientific figure. Reconstruct it in TikZ and apply the following changes: \\
\{edit\_instruction\} \\
Wrap your code using \texttt{\textbackslash documentclass[tikz]\{standalone\}}, and include \texttt{\textbackslash begin\{document\}}...\texttt{\textbackslash end\{document\}}. Only output valid LaTeX code with no extra text.
\label{fig:prompt_editing}
\end{PromptBox}
\refstepcounter{figure}
\begin{PromptBox}{Reconstruction Prompt}
This is an image of a scientific figure. Reconstruct it in TikZ. \\
Wrap your code using \texttt{\textbackslash documentclass[tikz]\{standalone\}}, and include \texttt{\textbackslash begin\{document\}}...\texttt{\textbackslash end\{document\}}. Only output valid LaTeX code with no extra text.
\label{fig:prompt_reconstruction}
\end{PromptBox}

\subsubsection{Editing-Specific Rewards}
\label{subsubsec:Editing-Specific Rewards}
The prompt template for our instruction-following reward $\mathcal{R}_{\mathrm{IF}}$ is shown in Figure~\ref{fig:prompt_instruction_following}. As our VLM-as-a-judge backbone, we use Qwen3.6-27B (thinking disabled). It uses greedy decoding (\texttt{temperature=0.0} and \texttt{top\_p=1.0}) and 128 output tokens. Judging is done with vLLM on 1 x Nvidia H100 (94 GB).
\refstepcounter{figure}
\begin{PromptBox}{VLM-as-a-Judge ($\mathcal{R}_{\mathrm{IF}}$)}
You are a strict visual reward judge for scientific figure editing. \\
You are given TWO images: \\
- Image 1 is the source figure before editing. \\
- Image 2 is the predicted figure after editing. \\
You are also given a list of requested atomic edits. \\
Judge whether Image 2 applies each requested atomic edit relative to Image 1. \\
For each atomic edit, decide whether the rendered image visibly and fully applies that edit. \\
Binary rating rubric: \\
1 = APPLIED. All essential visible requirements are satisfied in the predicted figure, the correct target was modified, requested text or mathematics is legible, and conflicting old content is absent. \\
0 = NOT APPLIED. The edit is absent, incomplete, incorrect, applied to the wrong target, contradicted by old content, insufficiently legible, or cannot be reliably verified. \\
Important judging rules: \\
- Score the predicted figure, not the source figure. \\
- Use the source figure only to identify original objects, positions, labels, shapes, connections, and content that should be added, removed, or modified. \\
- Judge only visible instruction faithfulness, not general similarity or visual beauty. \\
- Do not give credit for incomplete or ambiguous attempts. \\
- For an edit with multiple required parts, return 1 only if all essential parts are visibly satisfied. \\
- For text and mathematical edits, return 1 only if the requested content is legible. \\
- If an edit changes X to Y, return 1 only if Y is visible at the correct location and conflicting X is absent there. \\
- For removals, return 1 only if the specified original content is absent from its original location. \\
- Evaluate every edit independently and preserve the exact order. \\
- Return exactly one integer per atomic edit. \\
Return ONLY compact valid JSON in exactly this format:\\
{\ttfamily
\{ \\
\hspace*{1.5em}"ratings": [0, 0, ..., 0] \\
\} \\
}
The array length must exactly match the number of atomic edits. \\
Atomic edits:\\
\{atomic\_edits\} \\
Your output JSON:
\label{fig:prompt_instruction_following}
\end{PromptBox}

\subsubsection{Multi-Reward Optimization with GDPO}
\label{subsubsec:Multi-Reward Optimization with GDPO}
GDPO independently normalizes the advantages induced by $\mathcal{R}_{\mathrm{SSim}}$ and $\mathcal{R}_{\mathrm{IF}}$ across its rollout group before combining them with equal weights. For the policy loss, we adopt the constant-length normalization proposed by Dr.GRPO~\citep{liu2025understanding} where the summed token-level loss of each rollout is normalized by the fixed maximum completion length $L$ which avoids introducing a response-length-dependent optimization bias for TikZ programs. We further adopt DAPO's Clip-Higher strategy~\citep{yu2025dapo}, using asymmetric clipping with $\epsilon_{\mathrm{low}}=0.2$ and $\epsilon_{\mathrm{high}}=0.28$. The relaxed upper bound allows larger probability increases for low-probability exploratory tokens while the lower bound remains unchanged. Rollouts are sampled with \texttt{temperature=1.0} and \texttt{top\_p=0.99}, with a maximum completion length of 2048 tokens. Completions truncated at this limit are excluded from the policy loss. We disable KL regularization ($\beta=0$).

\subsection{Experiments}
\label{subsec:Experiments}

\subsubsection{Models}
\label{subsubsec:Models}
We use separate hyperparameter configurations for models in the 3--4B (Small) and 8--9B (Large) parameter ranges during both SFT and RL. The configurations are summarized in Table~\ref{tab:summary_hyperparameters}. Input images are resized to $448\times448$. We exclude samples whose TikZ code exceeds 4,000 characters or whose instruction exceeds 2,000 characters. Optimization uses AdamW~\citep{loshchilov2019decoupledweightdecayregularization}. We train with Deepspeed ZeRO-2~\citep{rajbhandari2020zero}.
\begin{table*}[t]
\centering
\caption{Training hyperparameters for the small (3--4B) and large (8--9B) models.}
\label{tab:summary_hyperparameters}
\small
\setlength{\tabcolsep}{8pt}
\renewcommand{\arraystretch}{1.02}
\begin{tabular}{lcc|cc}
\toprule
\multirow{2}{*}{\textbf{Hyperparameter}} &
\multicolumn{2}{c}{\textbf{SFT}} &
\multicolumn{2}{c}{\textbf{RL}} \\
\cmidrule(lr){2-3}\cmidrule(lr){4-5}
& \textbf{Small} & \textbf{Large}
& \textbf{Small} & \textbf{Large} \\
\midrule
Training duration (days)
& 6 & 13 & 8 & 10 \\
GPUs
& 4 x H100 & 4 x H100 & 3 x H100 & 3 x H100 \\
Epochs
& 2 & 2 & 1 & 1 \\
Per-device batch size
& 10 & 6 & 10 & 6 \\
Gradient accumulation steps
& 4 & 7 & 6 & 10 \\
Learning rate
& $1\times10^{-4}$ & $2\times10^{-5}$
& $2\times10^{-6}$ & $1\times10^{-6}$ \\
Learning-rate scheduler
& \texttt{cosine} & \texttt{cosine}
& \texttt{constant} & \texttt{constant} \\
Weight decay
& $0.0$ & $0.0$ & $0.01$ & $0.01$ \\
Generations per prompt
& -- & -- & 8 & 8 \\
\bottomrule
\end{tabular}
\end{table*}
All open-source baselines and trained models are evaluated on
DaEdiTikZ-Bench with a maximum of 2,048 output tokens, temperature $0.2$, top-$p$ $0.9$, and top-$k$ $50$. Proprietary GPT and Gemini models use their default reasoning settings and a maximum output budget of 10K tokens. For the out-of-distribution evaluation on CharXiv and SPIQA, we use temperature $0.1$ and a 10K-token output budget for both open-source and proprietary models to test extrapolation beyond the open models’ training output regime.

\subsubsection{Metrics}
\label{subsubsec:Metrics}
TeX Edit Distance (TED) uses Extended Edit Distance~\citep{kusner2015icml-word} with \texttt{TexLexer}. DreamSim (DSim) uses an ensemble of CLIP~\citep{radford2021learning}, DINO~\citep{caron2021emerging}, and OpenCLIP (\texttt{ViT-B/16}). Average tokens (AT) are measured with the \texttt{o200k\_base} tokenizer. Figure~\ref{fig:prompt_evaluation} presents our task specific VLM-as-a-Judge metric for Edit Application (EA), Source preservation (SP), and Visual Quality (VQ).
\refstepcounter{figure}
\begin{PromptBox}{VLM-as-a-Judge (EA, SP, and VQ)}
You are evaluating whether a predicted edited scientific figure correctly follows an edit instruction. \\\\
You are given: \\
1. The edit instruction \\
2. The original source figure image before editing \\
3. The predicted edited figure image \\
4. A reference target figure image showing one intended edited result \\\\
Important: \\
- The edit instruction is the primary specification. \\
- The reference target image is a helpful guide for the intended result, but it may contain minor artifacts, imperfect alignment, or details not fully described in the instruction. \\
- Do not require the prediction to copy harmless imperfections from the reference target. \\
- Reward predictions that correctly apply the instruction, preserve unrelated source content/avoid extra changes, and remain visually clean. \\\\
Score the prediction on three criteria from 0 to 10. \\\\
Criterion 1: edit\_application\_score \\
How completely and correctly are the requested edits applied? \\
0 = no requested edits are applied or the prediction is unrelated/unusable \\
1-2 = almost all requested edits are missing or wrong \\
3-4 = a few requested edits are attempted, but most are missing/wrong \\
5-6 = some requested edits are correct, but important edits are missing or inaccurate \\
7-8 = most requested edits are correct, with minor omissions or inaccuracies \\
9 = essentially all requested edits are correct, with only tiny issues \\
10 = all requested edits are applied correctly and completely \\\\
Criterion 2: source\_preservation\_score \\
How well does the prediction preserve all source content not required to change, and avoid adding/removing unrelated elements? \\
0 = unchanged source content is completely lost, corrupted, replaced, or dominated by unrelated additions \\
1-2 = most unchanged content is badly altered, removed, or many unrelated elements are added \\
3-4 = many unchanged elements are altered, missing, misplaced, or extra unrelated elements are present \\
5-6 = major unchanged structure is preserved, but several details change unnecessarily or some unrelated elements appear \\
7-8 = most unchanged content is preserved, with only minor/moderate unrelated changes \\
9 = nearly all unchanged content is preserved, with only tiny unrelated differences \\
10 = all unchanged source content is preserved very well and no unrelated elements are introduced \\\\
Criterion 3: visual\_quality\_score \\
How visually clean, legible, and publication-ready is the predicted figure? \\
0 = unusable rendering, blank image, or severe corruption \\
1-2 = severe layout/rendering problems, mostly unreadable \\
3-4 = many visual problems such as clipping, overlap, or unreadable labels \\
5-6 = usable but visibly flawed or messy \\
7-8 = mostly clean and legible, with minor/moderate visual issues \\
9 = very clean, with only tiny visual issues \\
10 = clean, legible, well-aligned, and publication-quality \\\\
Return valid JSON only in exactly this format: \\
{\ttfamily
\{ \\
\hspace*{1.5em}"edit\_application\_score": 0, \\
\hspace*{1.5em}"edit\_application\_reasoning": "", \\
\hspace*{1.5em}"source\_preservation\_score": 0, \\
\hspace*{1.5em}"source\_preservation\_reasoning": "", \\
\hspace*{1.5em}"visual\_quality\_score": 0, \\
\hspace*{1.5em}"visual\_quality\_reasoning": "", \\
\}
}\\\\
Use integer scores from 0 to 10. Keep each reasoning field to one concise sentence. \\\\
Edit instruction: \\
\{edit\_instruction\} \\\\
Your output JSON:
\label{fig:prompt_evaluation}
\end{PromptBox}

\subsection{Results}
\label{subsec:Results}

\subsubsection{Automatic Evaluation}
\label{subsubsec:Automatic Evaluation}
DaEdiTikZ-Bench enables evaluating inverse graphics by treating source and target figures of each editing pair as independent reconstruction examples. We evaluate all 690 figures using the same metrics where applicable, excluding EA and adapting SP (Table~\ref{tab:detikzify_evaluation_full}). Our EdiTikZ-4B and 9B models achieve 0.701 and 0.748 Avg, outperforming all evaluated baselines including GPT-5.6-Sol (0.652), Gemini-3.1-Pro (0.655), and improving substantially over their base models (+0.374/+0.430). Moreover, EdiTikZ-8B performs worse than DeTikZify-8B (0.624 vs.\ 0.672) suggesting that editing supervision does not improve reconstruction, whereas reconstruction supervision improves editing. We hypothesize that editing requires preserving large parts of the source figure while applying localized changes, so that additional reconstruction examples strengthen the capability needed for editing. Conversely, reconstruction does not require instruction following, and mixing its image-to-TikZ supervision with potentially noisy edit instructions may dilute its objective.

\begin{table*}[t]
\centering
\caption{Reconstruction performance on the 790 endpoint figures of DaEdiTikZ-Bench.}
\label{tab:detikzify_evaluation_full}
\small
\setlength{\tabcolsep}{5pt}
\renewcommand{\arraystretch}{0.95}

\begin{tabular}{lcccc:c:cc}
\toprule

\textbf{Model} &
\textbf{TED}$\downarrow$ &
\textbf{DSim}$\uparrow$ &
\textbf{SP\textsubscript{R}}$\uparrow$ &
\textbf{VQ}$\uparrow$ &
\textbf{Avg}$\uparrow$ &
\textbf{CR}$\uparrow$ &
\textbf{AT}$\downarrow$ \\
\midrule

GPT-5.6-Sol
& 0.798 & 0.777 & \textbf{0.802} & 0.828
& \cellcolor{gray!29}0.652
& 84.0\% & 568 \\

GPT-5.5
& 0.791 & 0.628 & 0.642 & 0.684
& \cellcolor{gray!22}0.541
& 70.0\% & 533 \\

Gemini-3.1-Pro
& 0.730 & 0.773 & 0.770 & 0.808
& \cellcolor{gray!29}0.655
& 82.0\% & \underline{459} \\

Gemini-3.6-Flash
& 0.742 & 0.573 & 0.598 & 0.614
& \cellcolor{gray!20}0.511
& 62.0\% & \textbf{340} \\

\midrule

Qwen3.6-27B
& 0.782 & 0.581 & 0.470 & 0.590
& \cellcolor{gray!17}0.465
& 66.4\% & 583 \\

Qwen3.5-27B
& 0.769 & 0.673 & 0.561 & 0.700
& \cellcolor{gray!22}0.541
& 76.8\% & 525 \\

Qwen2.5-VL-7B
& 0.778 & 0.482 & 0.237 & 0.495
& \cellcolor{gray!10}0.359
& 60.7\% & 516 \\

\midrule

Qwen2.5-VL-3B
& 0.810 & 0.354 & 0.122 & 0.363
& \cellcolor{gray!3}0.257
& 48.1\% & 748 \\

DeTikZify-3B
& 0.681 & 0.674 & 0.380 & 0.634
& \cellcolor{gray!19}0.502
& 76.1\% & 661 \\

EdiTikZ-3B
& 0.718 & 0.697 & 0.348 & 0.669
& \cellcolor{gray!19}0.499
& 80.3\% & 624 \\

\midrule

Qwen3-VL-4B
& 0.801 & 0.480 & 0.302 & 0.493
& \cellcolor{gray!10}0.369
& 58.6\% & 749 \\

EdiTikZ-4B\textsubscript{Qwen3}
& 0.651 & 0.810 & 0.538 & 0.782
& \cellcolor{gray!27}0.620
& 89.3\% & 535 \\

\midrule

Qwen3-VL-8B
& 0.784 & 0.555 & 0.360 & 0.559
& \cellcolor{gray!14}0.423
& 65.9\% & 625 \\

DeTikZify-8B
& 0.640 & 0.843 & 0.661 & 0.822
& \cellcolor{gray!30}0.672
& 91.9\% & 510 \\

EdiTikZ-8B
& 0.690 & 0.795 & 0.609 & 0.780
& \cellcolor{gray!27}0.624
& 87.1\% & 545 \\

\midrule

Qwen3.5-4B
& 0.826 & 0.352 & 0.220 & 0.338
& \cellcolor{gray!4}0.271
& 42.6\% & 773 \\

EdiTikZ-4B
& \underline{0.618} & \underline{0.850} & 0.727 & \underline{0.843}
& \cellcolor{gray!32}\underline{0.701}
& \underline{91.6\%} & 512 \\

\midrule

Qwen3.5-9B
& 0.810 & 0.480 & 0.344 & 0.481
& \cellcolor{gray!11}0.374
& 56.2\% & 750 \\

EdiTikZ-9B
& \textbf{0.590} & \textbf{0.894} & \underline{0.795} & \textbf{0.892}
& \cellcolor{gray!35}\textbf{0.748}
& \textbf{94.5\%} & 503 \\

\bottomrule
\end{tabular}
\end{table*}

\subsubsection{Human Evaluation}
\label{subsubsec:Human Evaluation}
Five master's students, three PhD students, and one faculty member (5 male, 4 female) participate in the human evaluation. Each annotator receives detailed guidelines and an Excel sheet containing 20 benchmark examples, yielding 240 example-level annotations and 4,320 individual criterion ratings. Each row presents the source figure, edit instruction, and randomly ordered, anonymized outputs from Gemini-3.1-Pro, GPT-5.6-Sol, Qwen3.5-4B, EdiTikZ-4B, EdiTikZ-4B-RL, Qwen3.5-9B, EdiTikZ-9B, and EdiTikZ-9B-RL. Successfully compiled outputs are rated on 1--7 Likert scales for Edit Application (EA), Source Preservation (SP), and Visual Quality (VQ). Non-compilable outputs receive a score of 0. The complete rating criteria are provided below, and representative examples are shown in Figure~\ref{fig:human_eval_1},~\ref{fig:human_eval_2}, and~\ref{fig:human_eval_3}. Likert scale definitions are shown below:
\begin{itemize}
\item \textbf{Edit Application (EA)}: 7) All requested edits are applied correctly and completely. 6) Essentially all requested edits are correct, with only tiny issues. 5) Most requested edits are correct, with minor omissions or inaccuracies. 4) Some requested edits are correct, but important edits are missing or inaccurate. 3) A few requested edits are attempted, but most are missing or wrong. 2) Almost all requested edits are missing or wrong. 1) No requested edits are applied, or the prediction is unrelated or unusable.
\item \textbf{Source Preservation (SP)}: 7) All unchanged source content is preserved very well, and no unrelated elements are introduced. 6) Nearly all unchanged content is preserved, with only tiny unrelated differences. 5) Most unchanged content is preserved, with only minor or moderate unrelated changes. 4) The major unchanged structure is preserved, but several details change unnecessarily or some unrelated elements appear. 3) Many unchanged elements are altered, missing, misplaced, or accompanied by extra unrelated elements. 2) Most unchanged content is badly altered or removed, or many unrelated elements are added. 1) Unchanged source content is completely lost, corrupted, replaced, or dominated by unrelated additions.
\item \textbf{Visual Quality (VQ)}: 7) Clean, legible, well-aligned, and publication-quality. 6) Very clean, with only tiny visual issues. 5) Mostly clean and legible, with minor or moderate visual issues. 4) Usable but visibly flawed or messy. 3) Many visual problems, such as clipping, overlap, or unreadable labels. 2) Severe layout or rendering problems; mostly unreadable. 1) Unusable rendering, blank image, or severe corruption.
\end{itemize}
\begin{figure}[h]
  \centering
  \includegraphics[width=\textwidth]{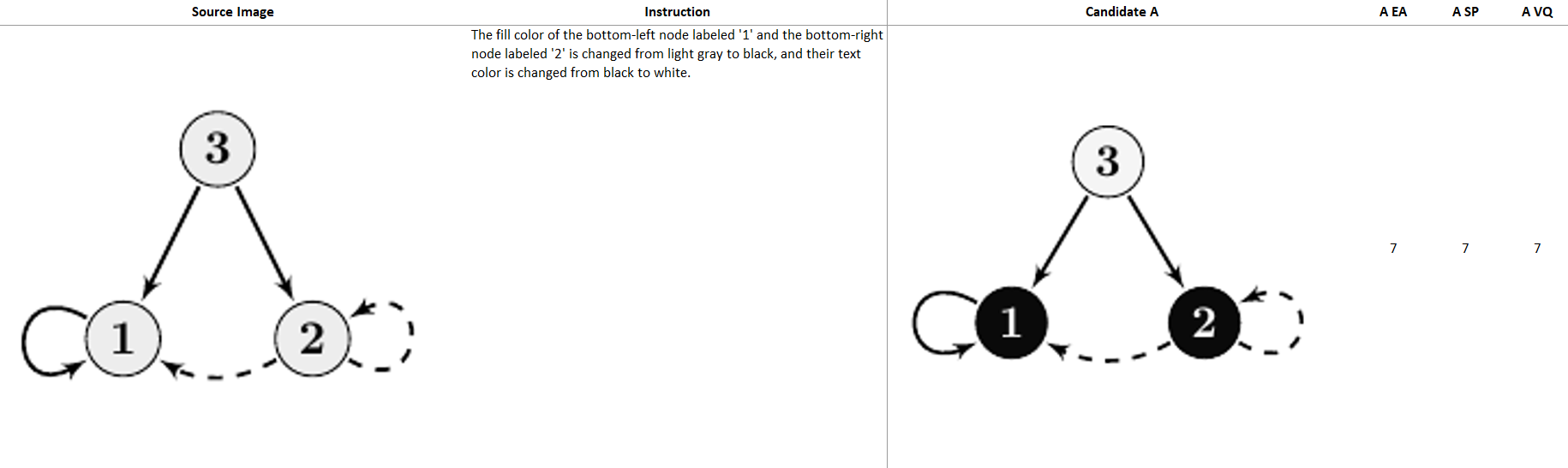}
  \caption{Example with perfect scores for EA, SP, and VQ.}
  \label{fig:human_eval_1}
\end{figure}
\begin{figure}[h]
  \centering
  \includegraphics[width=\textwidth]{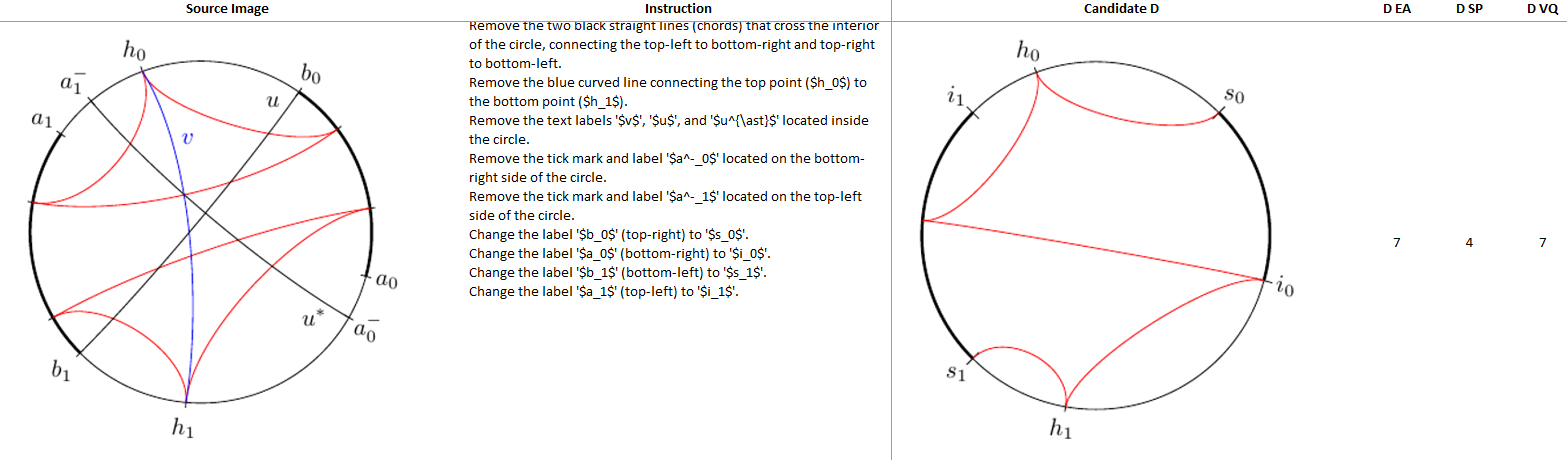}
  \caption{Example with lower source preservation but high edit application and visual quality.}
  \label{fig:human_eval_2}
\end{figure}
\begin{figure}[h]
  \centering
  \includegraphics[width=\textwidth]{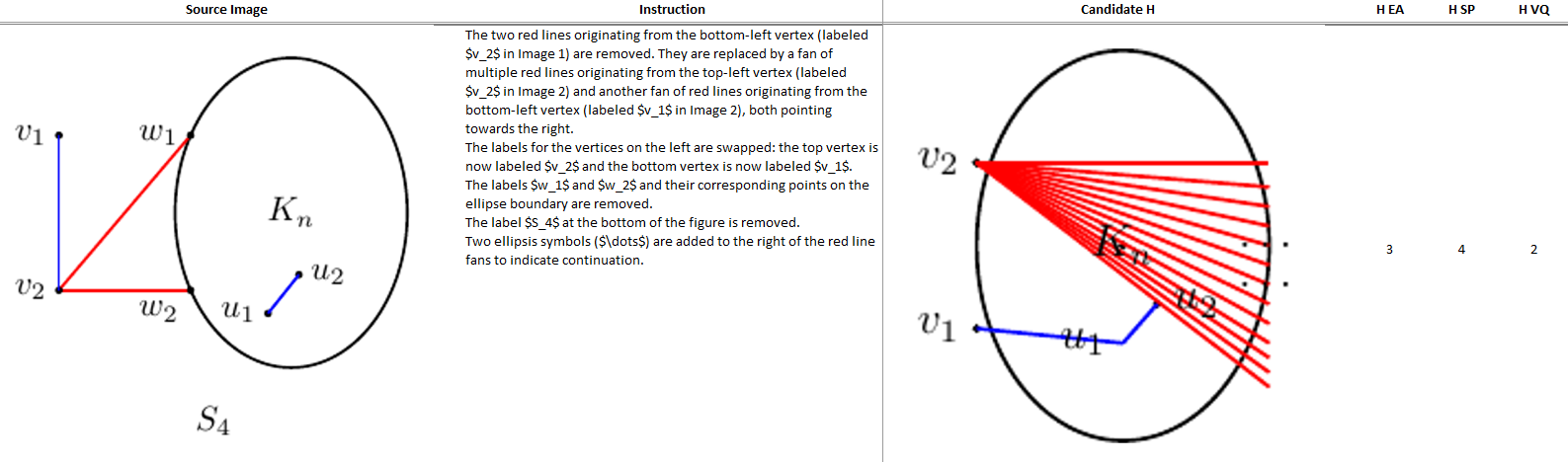}
  \caption{Example with lower scores for all three.}
  \label{fig:human_eval_3}
\end{figure}

\subsubsection{Generalization under Severe Distribution Shift}
\label{subsubsec:Generalization under Severe Distribution Shift}
During pilot generation, synthetic instructions frequently collapsed to repetitive edit types, specified only one or two shallow changes, or referred to elements that were not visibly grounded in the input figure. We therefore condition GPT-5.6-Sol on the desired number of atomic edits and the exact numbers of \texttt{modify}, \texttt{add}, and \texttt{remove} intents. The prompt additionally specifies admissible operation types, atomicity and visual-grounding constraints, and nine diverse human-written examples of plausible scientific-figure edits. To avoid a fixed synthetic edit profile, we sample the requested number of edits and intent composition for each figure from the empirical DaEdiTikZ distribution. The complete prompt for generating synthetic edit instructions for the SPIQA and CharXiv analyses is provided in Figure~\ref{fig:prompt_synthetic}.
\refstepcounter{figure}
\begin{PromptBox}{OOD Edit-Instruction Generation}
You are creating image-editing instructions for a scientific-figure editing evaluation benchmark. \\

Inspect the supplied scientific figure carefully. \\

Generate exactly \{num\_edits\} atomic edit instructions. \\

The instructions must contain exactly:

- \{num\_modify\} MODIFY operations \\
- \{num\_add\} ADD operations \\
- \{num\_remove\} REMOVE operations \\

OPERATION DEFINITIONS \\

MODIFY: \\
Modify an existing visible element while preserving or replacing its role, appearance, value, position, structure, or content. \\

ADD: \\
Introduce a new element that is not currently present in the figure. \\

REMOVE: \\
Delete an existing visible element without replacing it as part of the same atomic operation. \\

ATOMICITY \\
Each edit object must represent exactly one atomic edit. \\
Do not combine multiple independent edits into one instruction merely to satisfy the requested operation counts. \\
A replacement of one visible element with another may be treated as a MODIFY operation when it represents a single coherent replacement. \\

VALIDITY REQUIREMENTS \\
- Every MODIFY or REMOVE instruction must refer to an element that is clearly visible and unambiguously identifiable in the supplied image. \\
- Do not invent existing labels, values, curves, nodes, arrows, boxes, legends, colors, equations, annotations, or other elements. \\
- Never refer to an existing element unless you can visually verify that it is present. \\
- For ADD instructions, describe the new element's position or connection relative to clearly visible existing elements. \\
- All requested edits must be visually executable from the supplied image alone. \\
- Do not require information from the source paper, underlying numerical data, hidden metadata, or outside knowledge. \\
- The edits must be mutually compatible. \\
- Do not ask to modify, move, recolor, or relabel an element that another requested edit removes. \\
- Do not produce redundant edits that accomplish essentially the same modification twice. \\

DIVERSITY REQUIREMENTS \\
Prefer meaningful textual, annotation-level, geometric, structural, stylistic, semantic, or data-level modifications. \\

Possible edit targets include, but are not limited to: \\
- labels, equations, symbols, and numerical values \\
- axes, ticks, legends, titles, and plotted data \\
- curves, bars, markers, arrows, and paths \\
- nodes, blocks, connections, and graph topology \\
- scientific diagrams, architectures, circuits, and geometry \\
- annotations, boxes, braces, loops, regions, and grids \\
- positions, dimensions, orientations, and spatial relationships \\
- colors, line styles, marker styles, and fills when visually meaningful \\

Do not force any particular edit type if it does not naturally apply to the shown figure. \\

When multiple edits are requested, prefer edits that affect different meaningful aspects of the figure rather than repeatedly modifying nearly identical elements. \\

SPECIFICITY \\
Refer to visible elements using enough identifying information to make the edit unambiguous. \\

Good references include: \\
- the upper-right node \\
- the dashed rectangle around the encoder \\
- the blue curve labeled 'Method A' \\
- the y-axis tick at 0.5 \\
- the arrow connecting the first and second blocks \\

Avoid vague references such as 'the line', 'the box', 'the node', or 'the label' when multiple such elements exist. \\

STYLE EXAMPLES FROM REAL FIGURE-EDIT REQUESTS \\
These examples demonstrate the desired specificity and variety only. \\
Do not copy their content, entities, values, or sentence structures unless they naturally apply to the supplied figure. \\
1. Change the input label from
'$\hat{\boldsymbol{u}}^{\mathrm{sym}}$' to '$\boldsymbol{u}$'. \\
2. Change the fill color of the 'Policy' block from white to light red. \\
3. Add a new block labeled '$\mathbb{R}2\mathbb{C}$' at the end of the chain, after the Policy block. \\
4. Replace the rectangular block labeled 'CLK' and 'MEM' with a D-shaped AND gate symbol. \\
5. Adjust the y-axis tick marks to 0, 5, and 10. \\
6. Replace the current 3D surface with a Rosenbrock-style surface containing a long curved valley. \\
7. Move the rectangular path from the right side of the y-axis to the left side, with its vertical segment at $x=-1$ and its horizontal segments at $y=1$ and $y=-1$. \\
8. Add a blue curly brace annotation to the right of the legend spanning the top two entries, accompanied by the label 'Grouped'. \\
9. Remove the Greek letter labels '$\alpha$', '$\beta$', '$\gamma$', and '$\delta$' located above the first four nodes of the top horizontal line. \\

OUTPUT FORMAT \\
Return ONLY valid JSON. \\
Do not use Markdown code fences. \\
Do not include commentary before or after the JSON. \\
Do not explain your reasoning. \\

The JSON must have exactly this structure: \\
{\ttfamily
\{ \\
\hspace*{1.5em}"edits": [ \\
\hspace*{3em}\{ \\
\hspace*{4.5em}"operation": "modify|add|remove", \\
\hspace*{4.5em}"instruction": "..." \\
\hspace*{3em}\} \\
\hspace*{1.5em}] \\
\} \\
}\\
Requirements: \\
- "operation" must be exactly one of: "modify", "add", "remove". \\
- "instruction" must contain exactly one complete atomic edit instruction. \\
- Every instruction must be one sentence. \\
- The "edits" array must contain exactly \{num\_edits\} objects. \\
- There must be exactly \{num\_modify\} objects with operation "modify". \\
- There must be exactly \{num\_add\} objects with operation "add". \\
- There must be exactly \{num\_remove\} objects with operation "remove". \\

Now generate the edit instructions for the supplied scientific figure.
\label{fig:prompt_synthetic}
\end{PromptBox}
For the OOD evaluation, we adapt the prompt in Figure~\ref{fig:prompt_evaluation} to a reference free setting by omitting the target figure and all target-dependent instructions, and explicitly instructing the judge to evaluate the prediction from the source figure and edit instruction alone. The GPT-5.5 judge and decoding configuration remain unchanged. Table~\ref{tab:ood_table} provides the full results of our stress-tests on SPIQA and CharXiv. EdiTikZ generations are approximately $4$--$5\times$ longer than on DaEdiTikZ-Bench and exhibit substantially lower scores and compilation rates. GPT-5.6-Sol achieves the strongest overall performance on both datasets, while EdiTikZ-9B-RL remains competitive with higher compilation rates and slightly higher VQ on CharXiv. The improvement from SFT to RL is substantially larger across the OOD metrics than on DaEdiTikZ-Bench. Despite RL using only a small in-domain subset of DaEdiTikZ, its benefits transfer strongly to substantially more complex figures outside the training distribution. Across models, SP degrades most strongly, indicating that preserving unchanged content becomes particularly challenging as figure complexity increases.

\begin{table*}[t]
\centering
\caption{Performance of our EdiTikZ models on SPIQA and CharXiv against baselines.}
\label{tab:ood_table}
\setlength{\tabcolsep}{4pt}
\begin{tabular}{lccccc|ccccc}
\toprule
&
\multicolumn{5}{c}{\textbf{SPIQA}}
&
\multicolumn{5}{c}{\textbf{CharXiv}}
\\
\cmidrule(lr){2-6}
\cmidrule(lr){7-11}

\textbf{Model}
&
\textbf{EA}$\uparrow$
&
\textbf{SP}$\uparrow$
&
\textbf{VQ}$\uparrow$
&
\textbf{CR}$\uparrow$
&
\textbf{AT}$\downarrow$
&
\textbf{EA}$\uparrow$
&
\textbf{SP}$\uparrow$
&
\textbf{VQ}$\uparrow$
&
\textbf{CR}$\uparrow$
&
\textbf{AT}$\downarrow$
\\
\midrule

GPT-5.6-Sol
& \textbf{0.634}
& \textbf{0.598}
& \textbf{0.659}
& 80.6\%
& \textbf{1352}
& \textbf{0.524}
& \textbf{0.488}
& \underline{0.552}
& 68.3\%
& \textbf{1594}
\\

\midrule

Qwen3.5-27B
& 0.234
& 0.191
& 0.277
& 57.9\%
& 1787
& 0.206
& 0.164
& 0.247
& 42.6\%
& 2351
\\

\midrule

Qwen3.5-4B
& 0.063
& 0.037
& 0.083
& 24.2\%
& 2872
& 0.076
& 0.051
& 0.127
& 26.8\%
& 3164
\\

EdiTikZ-4B
& 0.158
& 0.112
& 0.309
& 63.2\%
& 2547
& 0.154
& 0.099
& 0.364
& 65.8\%
& 2550
\\

EdiTikZ-4B-RL
& 0.243
& 0.174
& 0.445
& 85.8\%
& \underline{1683}
& 0.210
& 0.145
& 0.439
& 81.1\%
& \underline{1917}
\\

\midrule

Qwen3.5-9B
& 0.088
& 0.066
& 0.152
& 34.2\%
& 1888
& 0.140
& 0.092
& 0.196
& 37.9\%
& 2428
\\

EdiTikZ-9B
& 0.319
& 0.238
& 0.428
& 74.2\%
& 2166
& 0.254
& 0.185
& 0.421
& 71.1\%
& 2883
\\

EdiTikZ-9B-RL
& \underline{0.520}
& \underline{0.466}
& \underline{0.622}
& \textbf{87.6\%}
& 1981
& \underline{0.399}
& \underline{0.323}
& \textbf{0.559}
& \textbf{85.0\%}
& 2541
\\

\bottomrule
\end{tabular}
\end{table*}

\subsubsection{Examples}
\label{subsubsec:Examples}

\begin{figure*}[p]
\centering

\newcommand{\codefile}[1]{%
    \lstinputlisting[
        style=mystyle,
        basicstyle=\fontsize{4.1}{4.5}\selectfont\ttfamily,
        frame=none,
        backgroundcolor=\color{white},
        aboveskip=1pt,
        belowskip=0pt,
        xleftmargin=1pt,
        xrightmargin=1pt
    ]{#1}%
}

\newcommand{\smalltitle}[1]{%
    {\fontsize{7}{7.5}\selectfont\bfseries #1}
}

\begin{tcolorbox}[
    width=\textwidth,
    colback=magenta!3,
    colframe=purple!40,
    title=Example with TikZ Code,
    enhanced,
    sharp corners=south,
    boxrule=0.5pt,
    left=2mm,
    right=2mm,
    top=1mm,
    bottom=1mm
]

\begin{minipage}[t]{0.29\linewidth}
\centering
\smalltitle{Source}

\vspace{1mm}

\includegraphics[
    width=\linewidth,
    height=2.9cm,
    keepaspectratio
]{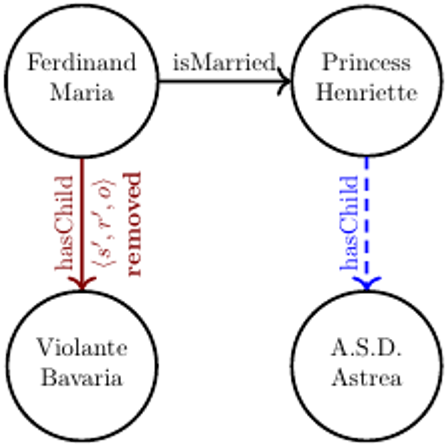}
\end{minipage}
\hfill
\begin{minipage}[t]{0.36\linewidth}
\centering
\smalltitle{Edit Instruction}

\vspace{-2mm}

\begin{tcolorbox}[
    colback=white,
    colframe=gray!45,
    boxrule=0.3pt,
    arc=1mm,
    left=1mm,right=1mm,
    top=1mm,bottom=1mm
]
\raggedright
\fontsize{5.4}{6.1}\selectfont
The node labeled 'A.S.D. Astrea' and the blue dashed arrow connecting 'Princess Henriette' to it are removed from the diagram. The arrow connecting 'Ferdinand Maria' to 'Violante Bavaria' is changed from a solid red line to a solid black line. The text annotation '$\langle s',r',o\rangle$ removed' located below the arrow between 'Ferdinand Maria' and 'Violante Bavaria' is removed. A new blue dashed arrow labeled 'hasChild' is added, connecting 'Princess Henriette' to 'Violante Bavaria'. A text block reading 'target prediction $\langle s,r,o\rangle$' is added to the bottom right, with an arrow pointing from it to the new blue dashed arrow.
\end{tcolorbox}
\end{minipage}
\hfill
\begin{minipage}[t]{0.29\linewidth}
\centering
\smalltitle{Ground Truth}

\vspace{1mm}

\includegraphics[
    width=\linewidth,
    height=2.9cm,
    keepaspectratio
]{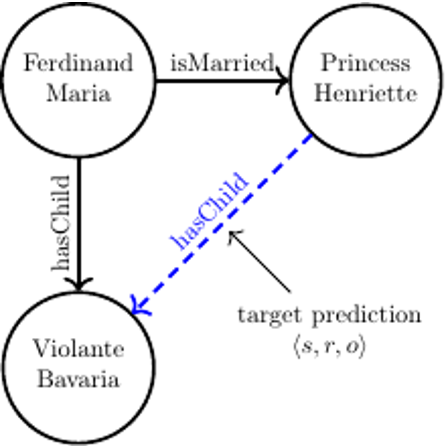}
\end{minipage}

\vspace{2mm}

\begin{minipage}[t]{0.44\linewidth}
\vspace{0pt}
\begin{tcolorbox}[
    width=\linewidth,
    colback=white,
    colframe=gray!45,
    boxrule=0.3pt,
    arc=1mm,
    left=0.5mm,right=0.5mm,
    top=0.5mm,bottom=0.5mm
]
\codefile{structure/examples_new/sample_0066_source.tex}
\end{tcolorbox}
\end{minipage}
\hfill
\begin{minipage}[t]{0.44\linewidth}
\vspace{0pt}
\begin{tcolorbox}[
    width=\linewidth,
    colback=white,
    colframe=gray!45,
    boxrule=0.3pt,
    arc=1mm,
    left=0.5mm,right=0.5mm,
    top=0.5mm,bottom=0.5mm
]
\codefile{structure/examples_new/sample_0066_gt.tex}
\end{tcolorbox}
\end{minipage}

\vspace{1mm}

\tcblower

\begin{minipage}[t]{0.485\linewidth}
\centering
\smalltitle{GPT-5.6-Sol}

\vspace{1mm}

\begin{tikzpicture}
\node[inner sep=0] (img) {%
    \includegraphics[
        width=0.72\linewidth,
        height=2.9cm,
        keepaspectratio
    ]{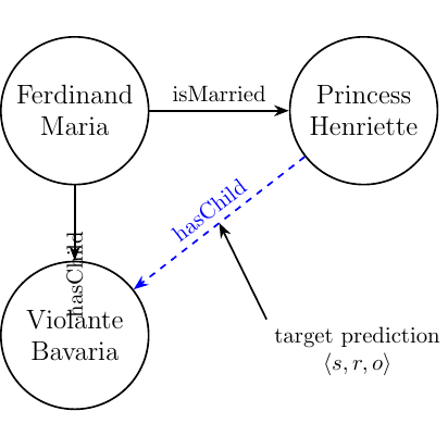}%
};

\node[
    anchor=west,
    right=1.5mm of img.east,
    align=left,
    fill=white,
    draw=gray!45,
    line width=0.3pt,
    rounded corners=1mm,
    inner xsep=1mm,
    inner ysep=0.8mm
] {%
    \fontsize{5.0}{5.8}\selectfont
    \begin{tabular}{@{}lr@{}}
    \textbf{TED}  & 0.716 \\
    \textbf{DSim} & 0.922 \\
    \textbf{EA}   & 1.000 \\
    \textbf{SP}   & 1.000 \\
    \textbf{VQ}   & 0.800 \\
    \textbf{AT}   & 334
    \end{tabular}
};
\end{tikzpicture}

\vspace{1mm}

\begin{tcolorbox}[
    colback=white,
    colframe=gray!45,
    boxrule=0.3pt,
    arc=1mm,
    left=0.5mm,right=0.5mm,
    top=0.5mm,bottom=0.5mm
]
\codefile{structure/examples_new/sample_0066_gpt.tex}
\end{tcolorbox}
\end{minipage}
\hfill
\begin{minipage}[t]{0.485\linewidth}
\centering
\smalltitle{EdiTikZ-9B-RL}

\vspace{1mm}

\begin{tikzpicture}
\node[inner sep=0] (img) {%
    \includegraphics[
        width=\linewidth,
        height=2.9cm,
        keepaspectratio
    ]{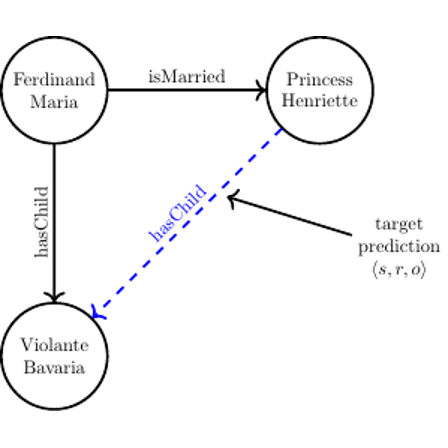}%
};

\node[
    anchor=west,
    right=1.5mm of img.east,
    align=left,
    fill=white,
    draw=gray!45,
    line width=0.3pt,
    rounded corners=1mm,
    inner xsep=1mm,
    inner ysep=0.8mm
] {%
    \fontsize{5.0}{5.8}\selectfont
    \begin{tabular}{@{}lr@{}}
    \textbf{TED}  & 0.690 \\
    \textbf{DSim} & 0.918 \\
    \textbf{EA}   & 1.000 \\
    \textbf{SP}   & 1.000 \\
    \textbf{VQ}   & 1.000 \\
    \textbf{AT}   & 335
    \end{tabular}
};
\end{tikzpicture}

\vspace{1mm}

\begin{tcolorbox}[
    colback=white,
    colframe=gray!45,
    boxrule=0.3pt,
    arc=1mm,
    left=0.5mm,right=0.5mm,
    top=0.5mm,bottom=0.5mm
]
\codefile{structure/examples_new/sample_0066_9b_rl.tex}
\end{tcolorbox}
\end{minipage}

\end{tcolorbox}

\caption{TikZ programs and rendered figures for a representative scientific-figure edit. Models receive the source figure and edit instruction and generate the edited TikZ program. Per-example TED, DSim, EA, SP, VQ, and AT are shown beside each prediction.}
\label{fig:tikz_code_example}

\end{figure*}

\begin{figure*}[p]
\centering

\newcommand{\codefile}[1]{%
    \lstinputlisting[
        style=mystyle,
        basicstyle=\fontsize{4.1}{4.5}\selectfont\ttfamily,
        frame=none,
        backgroundcolor=\color{white},
        aboveskip=1pt,
        belowskip=0pt,
        xleftmargin=1pt,
        xrightmargin=1pt
    ]{#1}%
}

\newcommand{\smalltitle}[1]{%
    {\fontsize{7}{7.5}\selectfont\bfseries #1}
}

\begin{tcolorbox}[
    width=\textwidth,
    colback=magenta!3,
    colframe=purple!40,
    title=Example with TikZ Code,
    enhanced,
    sharp corners=south,
    boxrule=0.5pt,
    left=2mm,
    right=2mm,
    top=1mm,
    bottom=1mm
]

\begin{minipage}[t]{0.29\linewidth}
\centering
\smalltitle{Source}

\vspace{1mm}

\includegraphics[
    width=\linewidth,
    height=2.9cm,
    keepaspectratio
]{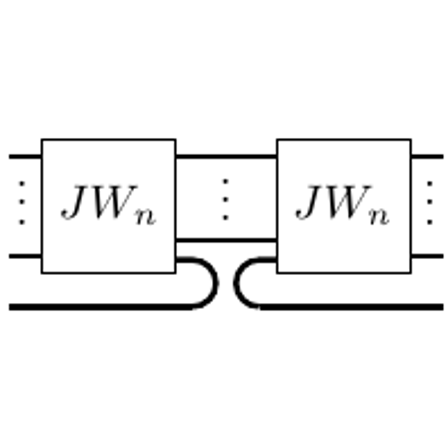}
\end{minipage}
\hfill
\begin{minipage}[t]{0.36\linewidth}
\centering
\smalltitle{Edit Instruction}

\vspace{-2mm}

\begin{tcolorbox}[
    colback=white,
    colframe=gray!45,
    boxrule=0.3pt,
    arc=1mm,
    left=1mm,right=1mm,
    top=1mm,bottom=1mm
]
\raggedright
\fontsize{5.4}{6.1}\selectfont
    Remove the right-hand $JW_n$ box, its external input/output lines, and the far-right vertical ellipsis. Remove the two horizontal lines. Move the loop from the gap between the boxes to a position beneath the remaining box, where it connects around the lower-left and lower-right sides. Increase the height of the remaining box upward from 2.0 to 2.3 units; the loop remains below the box rather than being contained within its vertical span. Reposition the two left input lines and the left and right vertical ellipses to suit the taller single-box layout.
\end{tcolorbox}
\end{minipage}
\hfill
\begin{minipage}[t]{0.29\linewidth}
\centering
\smalltitle{Ground Truth}

\vspace{1mm}

\includegraphics[
    width=\linewidth,
    height=2.9cm,
    keepaspectratio
]{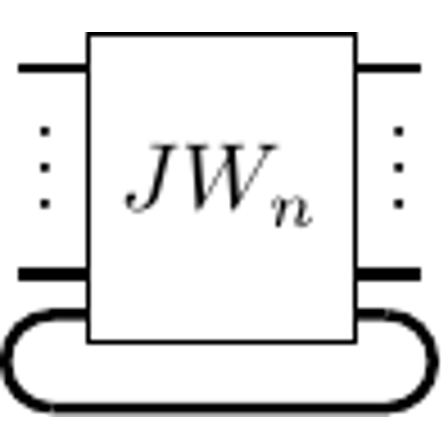}
\end{minipage}

\vspace{2mm}

\begin{minipage}[t]{0.44\linewidth}
\vspace{0pt}
\begin{tcolorbox}[
    width=\linewidth,
    colback=white,
    colframe=gray!45,
    boxrule=0.3pt,
    arc=1mm,
    left=0.5mm,right=0.5mm,
    top=0.5mm,bottom=0.5mm
]
\codefile{structure/examples_new/sample_0549_source.tex}
\end{tcolorbox}
\end{minipage}
\hfill
\begin{minipage}[t]{0.44\linewidth}
\vspace{0pt}
\begin{tcolorbox}[
    width=\linewidth,
    colback=white,
    colframe=gray!45,
    boxrule=0.3pt,
    arc=1mm,
    left=0.5mm,right=0.5mm,
    top=0.5mm,bottom=0.5mm
]
\codefile{structure/examples_new/sample_0549_gt.tex}
\end{tcolorbox}
\end{minipage}

\vspace{1mm}

\tcblower

\begin{minipage}[t]{0.485\linewidth}
\centering
\smalltitle{Gemini-3.1-Pro}

\vspace{1mm}

\begin{tikzpicture}
\node[inner sep=0] (img) {%
    \includegraphics[
        width=0.72\linewidth,
        height=2.9cm,
        keepaspectratio
    ]{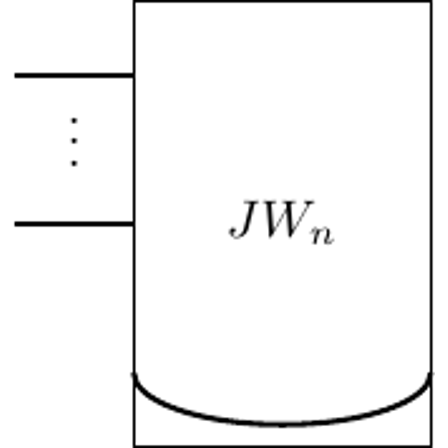}%
};

\node[
    anchor=west,
    right=1.5mm of img.east,
    align=left,
    fill=white,
    draw=gray!45,
    line width=0.3pt,
    rounded corners=1mm,
    inner xsep=1mm,
    inner ysep=0.8mm
] {%
    \fontsize{5.0}{5.8}\selectfont
    \begin{tabular}{@{}lr@{}}
    \textbf{TED}  & 0.680 \\
    \textbf{DSim} & 0.758 \\
    \textbf{EA}   & 0.500 \\
    \textbf{SP}   & 0.600 \\
    \textbf{VQ}   & 0.700 \\
    \textbf{AT}   & 165
    \end{tabular}
};
\end{tikzpicture}

\vspace{1mm}

\begin{tcolorbox}[
    colback=white,
    colframe=gray!45,
    boxrule=0.3pt,
    arc=1mm,
    left=0.5mm,right=0.5mm,
    top=0.5mm,bottom=0.5mm
]
\codefile{structure/examples_new/sample_0549_gemini.tex}
\end{tcolorbox}
\end{minipage}
\hfill
\begin{minipage}[t]{0.485\linewidth}
\centering
\smalltitle{EdiTikZ-9B-RL}

\vspace{1mm}

\begin{tikzpicture}
\node[inner sep=0] (img) {%
    \includegraphics[
        width=\linewidth,
        height=2.9cm,
        keepaspectratio
    ]{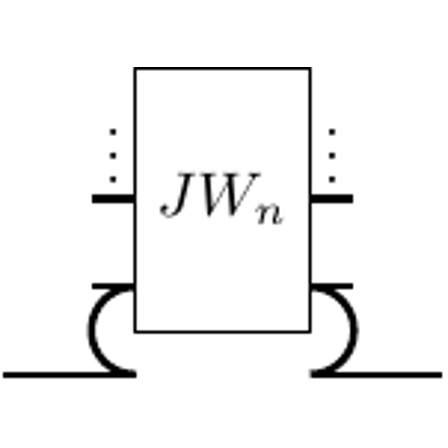}%
};

\node[
    anchor=west,
    right=1.5mm of img.east,
    align=left,
    fill=white,
    draw=gray!45,
    line width=0.3pt,
    rounded corners=1mm,
    inner xsep=1mm,
    inner ysep=0.8mm
] {%
    \fontsize{5.0}{5.8}\selectfont
    \begin{tabular}{@{}lr@{}}
    \textbf{TED}  & 0.659 \\
    \textbf{DSim} & 0.898 \\
    \textbf{EA}   & 0.500 \\
    \textbf{SP}   & 0.900 \\
    \textbf{VQ}   & 0.800 \\
    \textbf{AT}   & 323
    \end{tabular}
};
\end{tikzpicture}

\vspace{1mm}

\begin{tcolorbox}[
    colback=white,
    colframe=gray!45,
    boxrule=0.3pt,
    arc=1mm,
    left=0.5mm,right=0.5mm,
    top=0.5mm,bottom=0.5mm
]
\codefile{structure/examples_new/sample_0549_9b_rl.tex}
\end{tcolorbox}
\end{minipage}

\end{tcolorbox}

\caption{TikZ programs and rendered figures for a representative scientific-figure edit. Models receive the source figure and edit instruction and generate the edited TikZ program. Per-example TED, DSim, EA, SP, VQ, and AT are shown beside each prediction.}
\label{fig:tikz_code_example_2}

\end{figure*}

\begin{figure*}[p]
\centering

\newcommand{\codefile}[1]{%
    \lstinputlisting[
        style=mystyle,
        basicstyle=\fontsize{4.1}{4.5}\selectfont\ttfamily,
        frame=none,
        backgroundcolor=\color{white},
        aboveskip=1pt,
        belowskip=0pt,
        xleftmargin=1pt,
        xrightmargin=1pt
    ]{#1}%
}

\newcommand{\smalltitle}[1]{%
    {\fontsize{7}{7.5}\selectfont\bfseries #1}
}

\begin{tcolorbox}[
    width=\textwidth,
    colback=magenta!3,
    colframe=purple!40,
    title=Example with TikZ Code,
    enhanced,
    sharp corners=south,
    boxrule=0.5pt,
    left=2mm,
    right=2mm,
    top=1mm,
    bottom=1mm
]

\begin{minipage}[t]{0.29\linewidth}
\centering
\smalltitle{Source}

\vspace{1mm}

\includegraphics[
    width=\linewidth,
    height=2.9cm,
    keepaspectratio
]{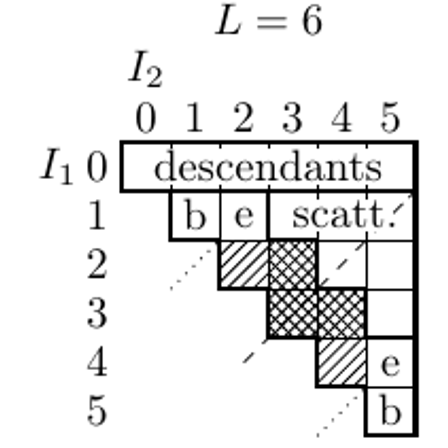}
\end{minipage}
\hfill
\begin{minipage}[t]{0.36\linewidth}
\centering
\smalltitle{Edit Instruction}

\vspace{-2mm}

\begin{tcolorbox}[
    colback=white,
    colframe=gray!45,
    boxrule=0.3pt,
    arc=1mm,
    left=1mm,right=1mm,
    top=1mm,bottom=1mm
]
\raggedright
\fontsize{4.4}{5.1}\selectfont
      The grid structure is expanded from a 6x6 matrix to a 7x7 matrix, adding a new column (index 6) and a new row (index 6) to the axes and the grid lines. The text label 'e' located at grid position (2, 1) is changed to 'b'." The fill pattern of the cell at grid position (3, 2) is changed from crosshatch to diagonal lines (north east lines). The fill pattern of the cell at grid position (4, 4) is changed from diagonal lines (north east lines) to crosshatch. A new filled cell with diagonal lines (north east lines) pattern is added at grid position (5, 4). A new filled cell with diagonal lines (north east lines) pattern is added at grid position (5, 5). The text label 'e' located at grid position (5, 4) is changed to 'b'. A new text label 'b' is added at grid position (6, 5). The dashed line indicating the boundary of the 'scatt.' region is extended to accommodate the larger grid size, now ending at the bottom right corner of the 7x7 grid.
\end{tcolorbox}
\end{minipage}
\hfill
\begin{minipage}[t]{0.29\linewidth}
\centering
\smalltitle{Ground Truth}

\vspace{1mm}

\includegraphics[
    width=\linewidth,
    height=2.9cm,
    keepaspectratio
]{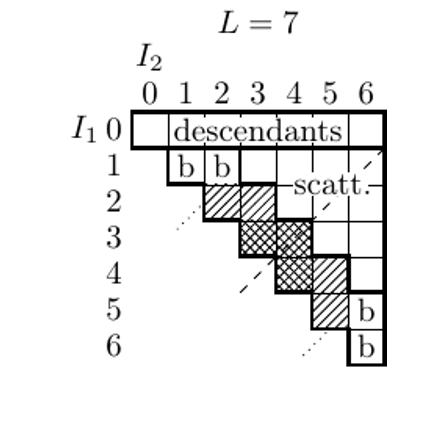}
\end{minipage}

\vspace{2mm}

\begin{minipage}[t]{0.44\linewidth}
\vspace{0pt}
\begin{tcolorbox}[
    width=\linewidth,
    colback=white,
    colframe=gray!45,
    boxrule=0.3pt,
    arc=1mm,
    left=0.5mm,right=0.5mm,
    top=0.5mm,bottom=0.5mm
]
\codefile{structure/examples_new/sample_0604_source.tex}
\end{tcolorbox}
\end{minipage}
\hfill
\begin{minipage}[t]{0.44\linewidth}
\vspace{0pt}
\begin{tcolorbox}[
    width=\linewidth,
    colback=white,
    colframe=gray!45,
    boxrule=0.3pt,
    arc=1mm,
    left=0.5mm,right=0.5mm,
    top=0.5mm,bottom=0.5mm
]
\codefile{structure/examples_new/sample_0604_gt.tex}
\end{tcolorbox}
\end{minipage}

\vspace{1mm}

\tcblower

\begin{minipage}[t]{0.485\linewidth}
\centering
\smalltitle{GPT-5.6-Sol}

\vspace{1mm}

\begin{tikzpicture}
\node[inner sep=0] (img) {%
    \includegraphics[
        width=0.72\linewidth,
        height=2.9cm,
        keepaspectratio
    ]{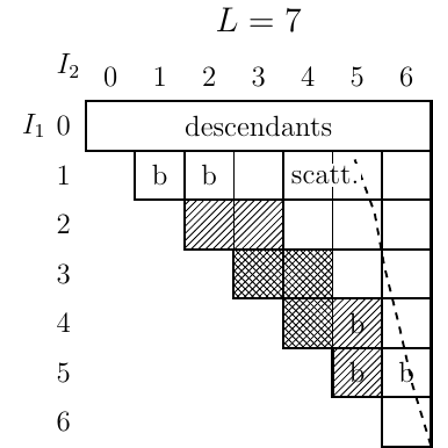}%
};

\node[
    anchor=west,
    right=1.5mm of img.east,
    align=left,
    fill=white,
    draw=gray!45,
    line width=0.3pt,
    rounded corners=1mm,
    inner xsep=1mm,
    inner ysep=0.8mm
] {%
    \fontsize{5.0}{5.8}\selectfont
    \begin{tabular}{@{}lr@{}}
    \textbf{TED}  & 0.647 \\
    \textbf{DSim} & 0.867 \\
    \textbf{EA}   & 0.500 \\
    \textbf{SP}   & 0.600 \\
    \textbf{VQ}   & 0.900 \\
    \textbf{AT}   & 671
    \end{tabular}
};
\end{tikzpicture}

\vspace{1mm}

\begin{tcolorbox}[
    colback=white,
    colframe=gray!45,
    boxrule=0.3pt,
    arc=1mm,
    left=0.5mm,right=0.5mm,
    top=0.5mm,bottom=0.5mm
]
\codefile{structure/examples_new/sample_0604_gpt.tex}
\end{tcolorbox}
\end{minipage}
\hfill
\begin{minipage}[t]{0.485\linewidth}
\centering
\smalltitle{EdiTikZ-4B-RL}

\vspace{1mm}

\begin{tikzpicture}
\node[inner sep=0] (img) {%
    \includegraphics[
        width=\linewidth,
        height=2.9cm,
        keepaspectratio
    ]{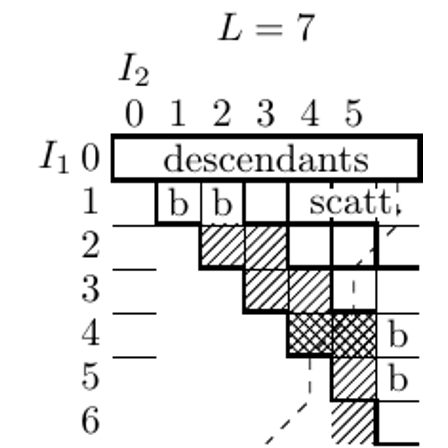}%
};

\node[
    anchor=west,
    right=1.5mm of img.east,
    align=left,
    fill=white,
    draw=gray!45,
    line width=0.3pt,
    rounded corners=1mm,
    inner xsep=1mm,
    inner ysep=0.8mm
] {%
    \fontsize{5.0}{5.8}\selectfont
    \begin{tabular}{@{}lr@{}}
    \textbf{TED}  & 0.804 \\
    \textbf{DSim} & 0.938 \\
    \textbf{EA}   & 0.200 \\
    \textbf{SP}   & 0.200 \\
    \textbf{VQ}   & 0.400 \\
    \textbf{AT}   & 1209
    \end{tabular}
};
\end{tikzpicture}

\vspace{1mm}

\begin{tcolorbox}[
    colback=white,
    colframe=gray!45,
    boxrule=0.3pt,
    arc=1mm,
    left=0.5mm,right=0.5mm,
    top=0.5mm,bottom=0.5mm
]
\codefile{structure/examples_new/sample_0604_4b_rl.tex}
\end{tcolorbox}
\end{minipage}

\end{tcolorbox}

\caption{TikZ programs and rendered figures for a representative scientific-figure edit. Models receive the source figure and edit instruction and generate the edited TikZ program. Per-example TED, DSim, EA, SP, VQ, and AT are shown beside each prediction.}
\label{fig:tikz_code_example_2}

\end{figure*}

\setlength{\aboverulesep}{0pt}
\setlength{\belowrulesep}{0pt}
\setlength{\tabcolsep}{2pt}
\renewcommand{\arraystretch}{1.03}
\setlength{\abovetopsep}{0pt}

\setlength{\descw}{0.25\textwidth}

\setlength{\imgw}{0.136\textwidth}

\setlength{\imgh}{2.6cm}

\newcolumntype{Y}{>{\raggedright\arraybackslash}p{\descw}}
\newcolumntype{C}{>{\centering\arraybackslash}p{\imgw}}

\newcommand{\exrowoneone}[6]{%
  \imgcell{#1} &
  \desc{#2} &
  \imgcell{#3} &
  \imgcellverygood{#4}
  {\fontsize{8}{8}\selectfont
  E:\textbf{7}\quad
  P:\textbf{7}\quad
  Q:\textbf{7}} 
  &
  \imgcell{#5}
  &
  \imgcellgood{#6}
  {\fontsize{8}{8}\selectfont
  E:\textbf{7}\quad
  P:\textbf{5}\quad
  Q:\textbf{5}} 
  \\
}

\newcommand{\exrowonetwo}[6]{%
  \imgcell{#1} &
  \desc{#2} &
  \imgcell{#3} &
  \imgcell{#4}
  &
  \imgcell{#5}
  &
  \imgcellverygood{#6}
  {\fontsize{8}{8}\selectfont
  E:\textbf{7}\quad
  P:\textbf{7}\quad
  Q:\textbf{7}} 
  \\
}

\newcommand{\exrowonethree}[6]{%
  \imgcell{#1} &
  \desc{#2} &
  \imgcell{#3} &
  \imgcellverygood{#4}
  {\fontsize{8}{8}\selectfont
  E:\textbf{7}\quad
  P:\textbf{7}\quad
  Q:\textbf{6}} 
  &
  \imgcellokay{#5}
  {\fontsize{8}{8}\selectfont
  E:\textbf{1}\quad
  P:\textbf{4}\quad
  Q:\textbf{5}} 
  &
  \imgcellverygood{#6}
  {\fontsize{8}{8}\selectfont
  E:\textbf{7}\quad
  P:\textbf{7}\quad
  Q:\textbf{7}} 
  \\
}

\newcommand{\exrowonefour}[6]{%
  \imgcell{#1} &
  \desc{#2} &
  \imgcell{#3} &
  \imgcellokay{#4}
  {\fontsize{8}{8}\selectfont
  E:\textbf{6}\quad
  P:\textbf{4}\quad
  Q:\textbf{4}}
  &
  \imgcell{#5}
  &
  \imgcell{#6}
  \\
}

\newcommand{\exrowonefive}[6]{%
  \imgcell{#1} &
  \desc{#2} &
  \imgcell{#3} &
  \imgcellokay{#4}
  {\fontsize{8}{8}\selectfont
  E:\textbf{4}\quad
  P:\textbf{4}\quad
  Q:\textbf{6}} 
  &
  \imgcellbad{#5}
  {\fontsize{8}{8}\selectfont
  E:\textbf{2}\quad
  P:\textbf{2}\quad
  Q:\textbf{2}} 
  &
  \imgcellgood{#6}
  {\fontsize{8}{8}\selectfont
  E:\textbf{5}\quad
  P:\textbf{5}\quad
  Q:\textbf{6}} 
  \\
}

\newcommand{\exrowonesix}[6]{%
  \imgcell{#1} &
  \desc{#2} &
  \imgcell{#3} &
  \imgcellverygood{#4}
  {\fontsize{8}{8}\selectfont
  E:\textbf{6}\quad
  P:\textbf{7}\quad
  Q:\textbf{7}} 
  &
  \imgcell{#5} 
  &
  \imgcellgood{#6}
  {\fontsize{8}{8}\selectfont
  E:\textbf{5}\quad
  P:\textbf{6}\quad
  Q:\textbf{7}} 
  \\
}

\newcommand{\exrowoneseven}[6]{%
  \imgcell{#1} &
  \desc{#2} &
  \imgcell{#3} &
  \imgcell{#4}
  &
  \imgcellokay{#5} 
  {\fontsize{8}{8}\selectfont
  E:\textbf{5}\quad
  P:\textbf{5}\quad
  Q:\textbf{4}} 
  &
  \imgcellverygood{#6}
  {\fontsize{8}{8}\selectfont
  E:\textbf{7}\quad
  P:\textbf{6}\quad
  Q:\textbf{7}} 
  \\
}

\begin{table}[t]
\centering
\caption{Scientific figure edits by GPT-5.6-Sol, Qwen3.5-9B, and EdiTikZ-9B-RL. Models receive the source image and VLM-generated edit instruction. Human annotations score edit application (E), source preservation (P), and visual quality (Q). Overall quality: \legendbox{green} very good, \legendbox{yellow} good, \legendbox{orange} bad, \legendbox{red} very bad.}
\label{tab:editikz_examples_1}
\begin{tabularx}{\textwidth}{@{}CYCCCC@{}}
    \toprule
    \headercell{Source} &
    \headercell{Edit Instruction} &
    \headercell{Ground Truth} &
    \headercell{GPT-5.6-Sol} &
    \headercell{Qwen3.5-9B} &
    \headercell{EdiTikZ-9B-RL} \\
    \midrule
    \exrowoneone
      {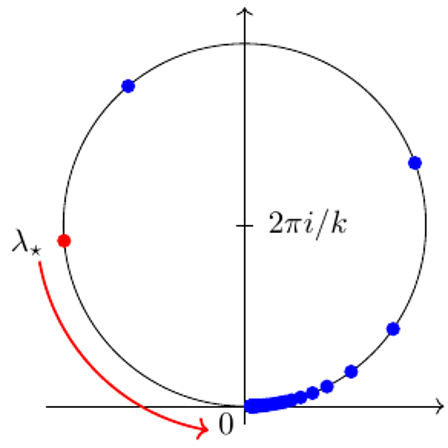}
      {The red curved arrow originating from the left side of the circle and pointing towards the origin is removed. The single red data point located on the left side of the circle (near the label \smath{\lambda_\star}) is changed to blue. A label \smath{\lambda_2} is added in the upper-left quadrant, positioned near the corresponding blue data point. A label \smath{\lambda_3} is added in the upper-right quadrant, positioned near the corresponding blue data point.}
      {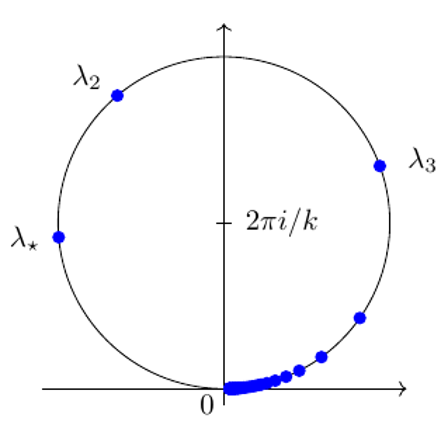}
      {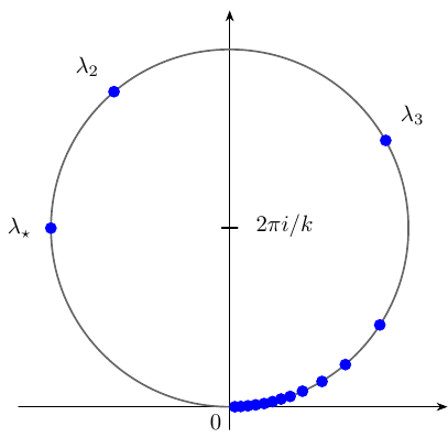}
      {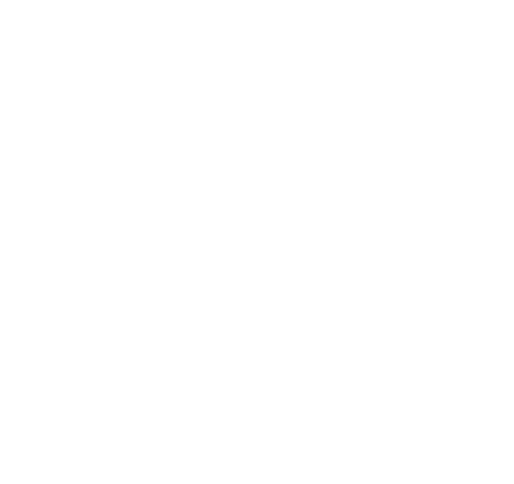}
      {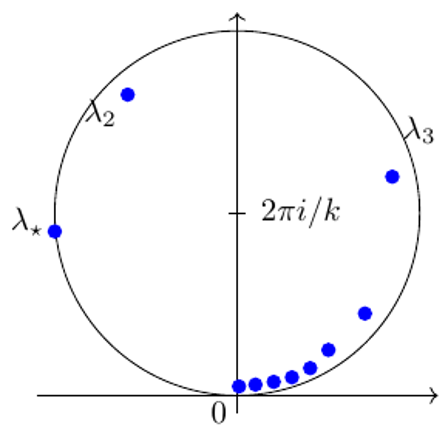}
    \midrule
    \exrowonetwo
      {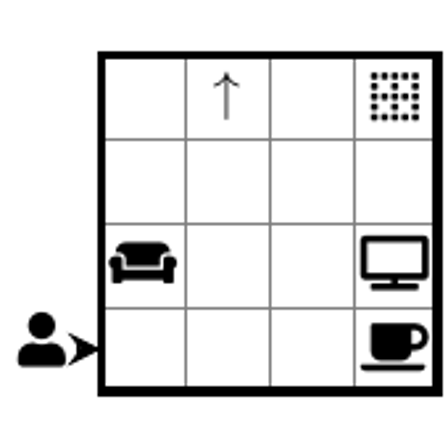}
      {The grid structure is reduced from a 4x4 layout to a 3x3 layout. This involves removing the second row and the third column of cells, effectively shrinking the main container area.}
      {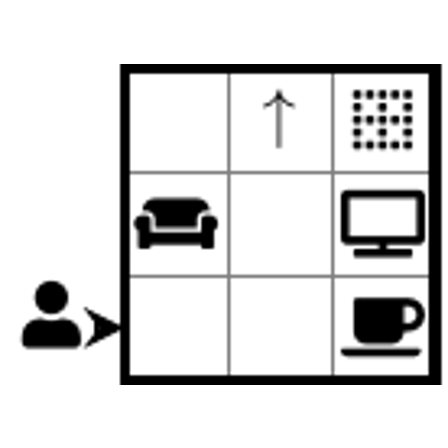}
      {structure/examples_1/not_compiled.png}
      {structure/examples_1/not_compiled.png}
      {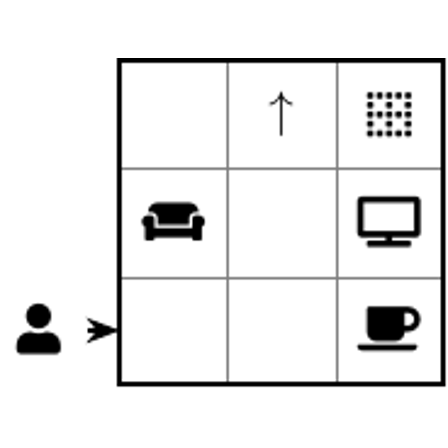}
    \midrule
    \exrowonethree
      {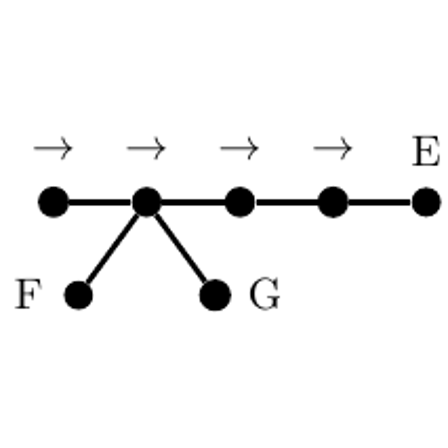}
      {The four right-pointing arrow symbols (\smath{\rightarrow}) located above the first four nodes on the top row are replaced with the Greek letters \smath{\alpha}, \smath{\beta}, \smath{\gamma}, and \smath{\delta} respectively, from left to right.}
      {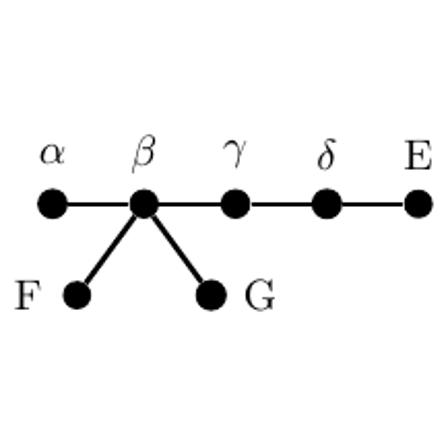}
      {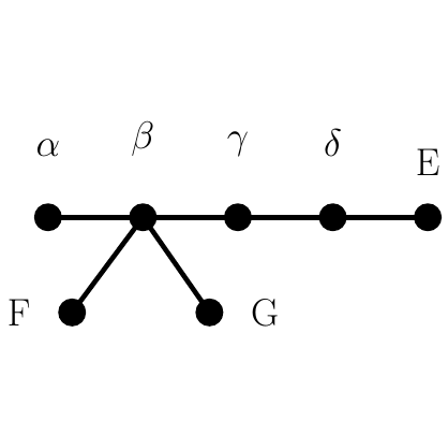}
      {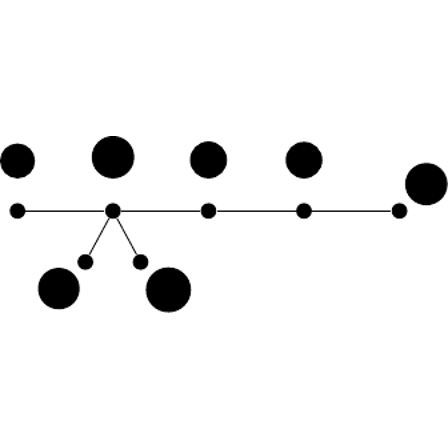}
      {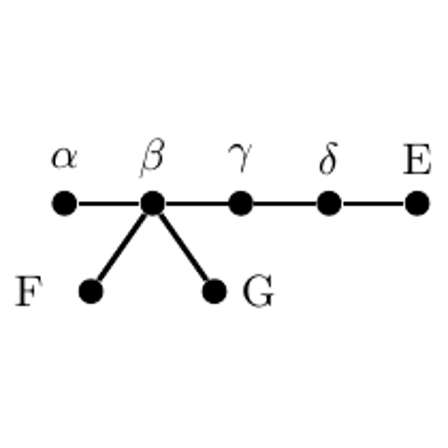}
    \midrule
    \exrowonefour
      {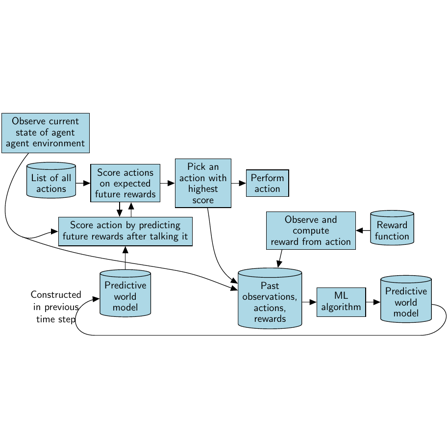}
      {The text in the bottom-left cylinder node (originally 'Predictive world model') is changed to 'Specifically incorrect predictive world model'. The color of the bottom-left cylinder node is changed from light blue to green. The text in the bottom-right cylinder node (originally 'Predictive world model') is changed to 'Correct predictive world model'. A new rectangular node labeled 'Model Editing' is added to the right of the 'Correct predictive world model' node. A new cylindrical node labeled 'Desired incorrect model elements' is added above the 'Model Editing' node. The color of the new 'Model Editing' node and the 'Desired incorrect model elements' node is set to green. A new arrow is added connecting the 'Correct predictive world model' node to the 'Model Editing' node. A new arrow is added connecting the 'Desired incorrect model elements' node to the 'Model Editing' node. The feedback loop arrow at the bottom right is modified to originate from the 'Model Editing' node and point back to the 'Specifically incorrect predictive world model' node.}
      {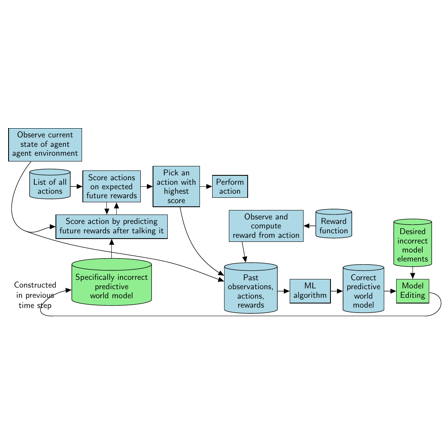}
      {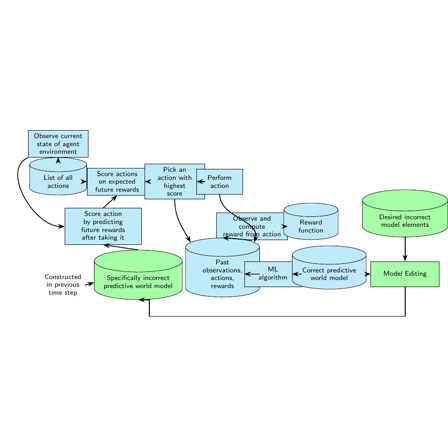}
      {structure/examples_1/not_compiled.png}
      {structure/examples_1/not_compiled.png}
    \midrule
    \exrowonefive
      {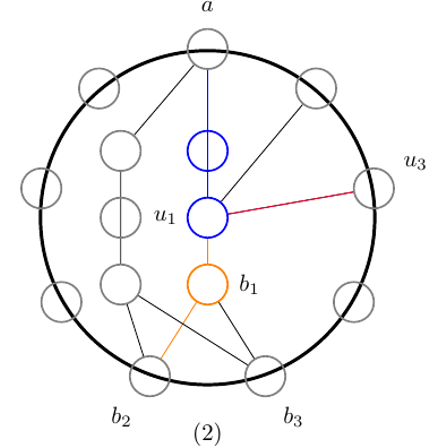}
      {The large outer circular boundary becomes invisible, while all nodes change from hollow outlines to lightly filled circles in their respective colors to match their outline color. The lower orange node is relabeled from \smath{b_1} to \smath{u_1} and move the label from the right side to the left side of the orange node. The central blue node is relabeled from \smath{u_1} to \smath{y} and move the label from the left side to the right side of the blue center node. The three inner gray nodes on the left shift downward. On the outer ring, label \smath{u_3} moves from nearby the node positioned at \smath{10^\circ} to  nearby the node positioned \smath{-30^\circ}. All inner pathes change to a slightly thicker style. A new green connection is added between the orange node and the node \smath{u_3}. The former red connection from the central blue node is removed. The black connection between center blue node \smath{y} and the node on the \smath{60^\circ} moves to between the first blue node on the top and the node on the \smath{60^\circ}. The black connection from the first blue node on the top to the node at approximately \smath{60^\circ} is removed. The large outer circular boundary from the node on left of the top node to node \smath{u_3} changes from black to green. The path between \smath{y} and \smath{u_1} changes from orange to blue.}
      {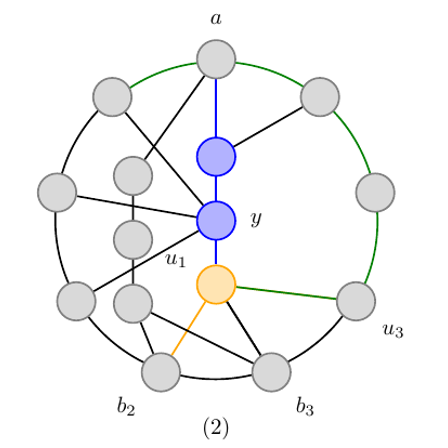}
      {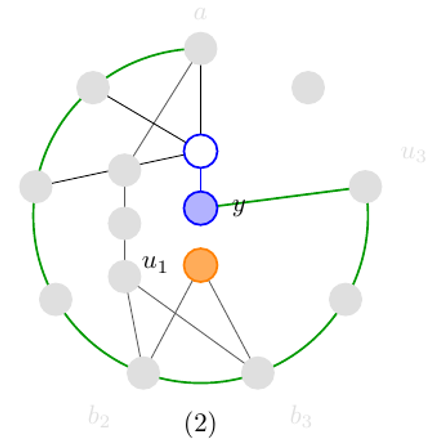}
      {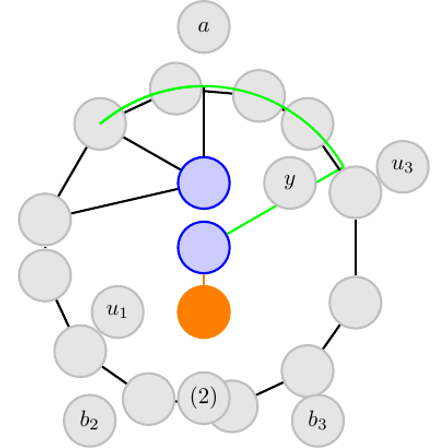}
      {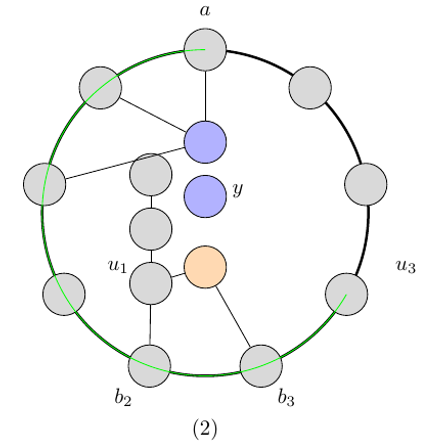}
    \midrule
    \exrowonesix
      {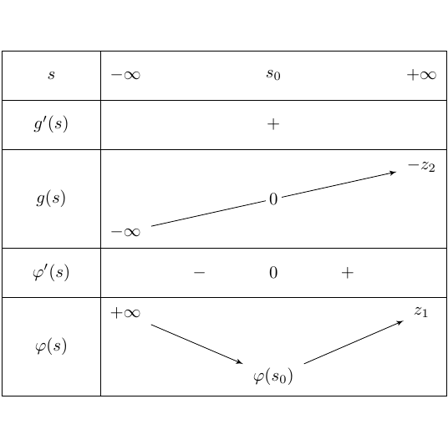}
      {In the row for \smath{g'(s)}, change the sign from '+' to '-'. In the row for \smath{g(s)}, change the starting value at \smath{-\infty} from \smath{-\infty} to \smath{+\infty} and reverse the direction of the arrow so it points downwards towards \smath{-z_2}. In the row for \smath{\varphi'(s)}, swap the signs in the intervals: change '-' to '+' in the interval between \smath{-\infty} and \smath{s_0}, and change '+' to '-' in the interval between \smath{s_0} and \smath{+\infty}. In the row for \smath{\varphi(s)}, change the starting value at \smath{-\infty} from \smath{+\infty} to \smath{-\infty} and reverse the direction of the first arrow so it points upwards towards \smath{\varphi(s_0)}. In the row for \smath{\varphi(s)}, reverse the direction of the second arrow so it points downwards from \smath{s_0} towards \smath{z_1}. In the row for \smath{\varphi(s)},\smath{\varphi(s_0)} is removed.}
      {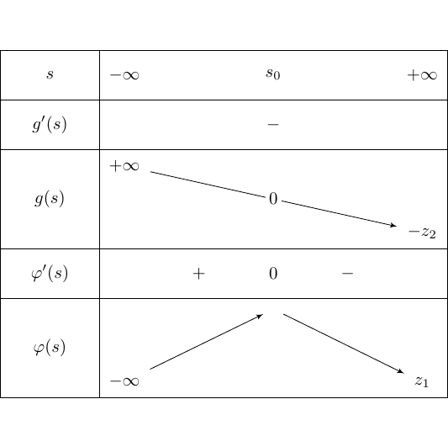}
      {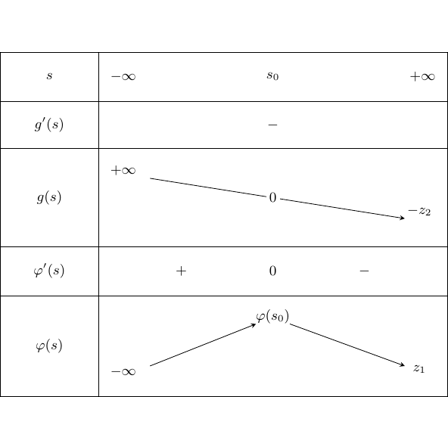}
      {structure/examples_1/not_compiled.png}
      {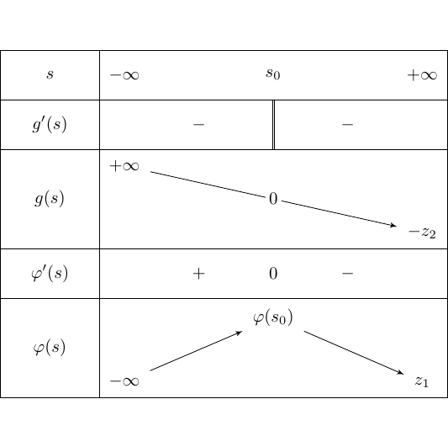}
    \midrule
    \exrowoneseven
      {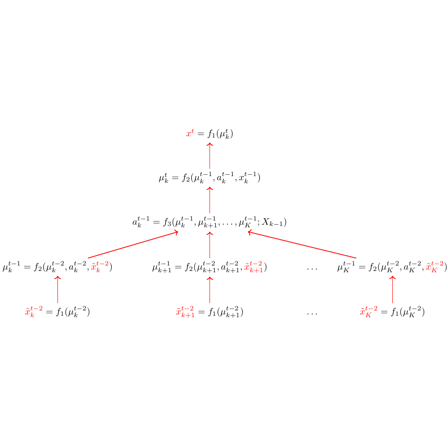}
      {The overall topology of the diagram is changed from a single central tree (one top node branching down to a row of three) to a split structure (two top nodes on the left and right with ellipsis dots between them, branching down to a single central node at the bottom). The top-level node text is changed from a single \smathsmall{x^t = f_1(\mu_k^t)} to two separate nodes: \smathsmall{\tilde{x}_{k+1}^t = f_1(\mu_{k+1}^t)} on the left and \smathsmall{\tilde{x}_K^t = f_1(\mu_K^t)} on the right. The second row nodes are changed from a single \smathsmall{\mu_{k}^t=f_2(\mu_k^{t-1},a_k^{t-1},x_k^{t-1})} equation to two separate equations: \smathsmall{\mu_{k+1}^t=f_2(\mu_{k+1}^{t-1},a_{k+1}^{t-1},x_{k+1}^{t-1})} on the left and \smathsmall{\mu_{K}^t=f_2(\mu_K^{t-1},a_K^{t-1},\tilde{x}_K^{t-1})} on the right. The third row nodes are changed from a single \smathsmall{a_{k}^{t-1}=f_3(\mu_k^{t-1},\mu_{k+1}^{t-1},...,\mu_K^{t-1};X_{k-1})} equation to two separate equations: \smathsmall{a_{k+1}^{t-1}=f_{3,k+1}(\mu_k^{t-1},...,\mu_K^{t-1};X_{k-1})} on the left and \smathsmall{a_{K}^{t-1}=f_{3K}(\mu_k^{t-1},...,\mu_K^{t-1};X_{k-1})} on the right. The fourth row is changed from a row of three \smathsmall{\mu} equations to a single central equation \smathsmall{\mu_k^{t-1} = f_2(\mu_k^{t-2}, a_k^{t-2}, x^{t-2})}. The bottom row is changed from a row of three \smathsmall{\tilde{x}} equations to a single central equation \smathsmall{x^{t-2} = f_1(\mu_k^{t-2})}. The connectivity arrows are updated to match the new structure: arrows now flow from the two top nodes down to the two middle nodes, which then converge onto the single bottom node.}
      {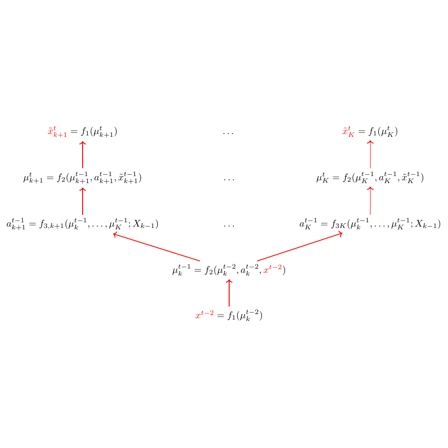}
      {structure/examples_1/not_compiled.png}
      {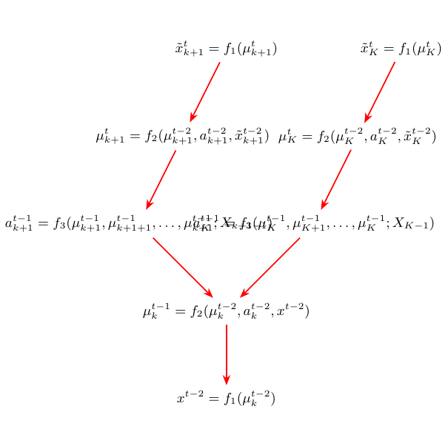}
      {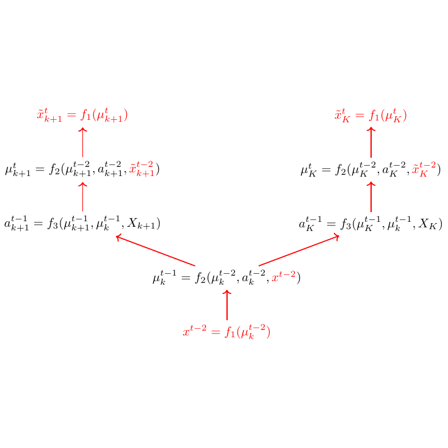}
    \bottomrule
\end{tabularx}
\end{table}

\setlength{\aboverulesep}{0pt}
\setlength{\belowrulesep}{0pt}
\setlength{\tabcolsep}{2pt}
\renewcommand{\arraystretch}{1.03}
\setlength{\abovetopsep}{0pt}

\setlength{\descw}{0.25\textwidth}

\setlength{\imgw}{0.136\textwidth}

\setlength{\imgh}{2.6cm}

\newcolumntype{Y}{>{\raggedright\arraybackslash}p{\descw}}
\newcolumntype{C}{>{\centering\arraybackslash}p{\imgw}}

\newcommand{\exrowtwoone}[6]{%
  \imgcell{#1} &
  \desc{#2} &
  \imgcell{#3} &
  \imgcellgood{#4}
  {\fontsize{8}{8}\selectfont
  E:\textbf{5}\quad
  P:\textbf{4}\quad
  Q:\textbf{7}} 
  &
  \imgcellokay{#5}
  {\fontsize{8}{8}\selectfont
  E:\textbf{3}\quad
  P:\textbf{4}\quad
  Q:\textbf{7}} 
  &
  \imgcellbad{#6}
  {\fontsize{8}{8}\selectfont
  E:\textbf{2}\quad
  P:\textbf{3}\quad
  Q:\textbf{4}} 
  \\
}

\newcommand{\exrowtwotwo}[6]{%
  \imgcell{#1} &
  \desc{#2} &
  \imgcell{#3} &
  \imgcellbad{#4}
  {\fontsize{8}{8}\selectfont
  E:\textbf{2}\quad
  P:\textbf{3}\quad
  Q:\textbf{3}} 
  &
  \imgcell{#5}
  &
  \imgcell{#6}
  \\
}

\newcommand{\exrowtwothree}[6]{%
  \imgcell{#1} &
  \desc{#2} &
  \imgcell{#3} &
  \imgcellokay{#4}
  {\fontsize{8}{8}\selectfont
  E:\textbf{3}\quad
  P:\textbf{4}\quad
  Q:\textbf{7}} 
  &
  \imgcellokay{#5}
  {\fontsize{8}{8}\selectfont
  E:\textbf{3}\quad
  P:\textbf{4}\quad
  Q:\textbf{6}} 
  &
  \imgcellokay{#6}
  {\fontsize{8}{8}\selectfont
  E:\textbf{3}\quad
  P:\textbf{3}\quad
  Q:\textbf{4}} 
  \\
}

\newcommand{\exrowtwofour}[6]{%
  \imgcell{#1} &
  \desc{#2} &
  \imgcell{#3} &
  \imgcell{#4}
  &
  \imgcell{#5}
  &
  \imgcellverygood{#6}
  {\fontsize{8}{8}\selectfont
  E:\textbf{7}\quad
  P:\textbf{6}\quad
  Q:\textbf{6}} 
  \\
}

\newcommand{\exrowtwofive}[6]{%
  \imgcell{#1} &
  \desc{#2} &
  \imgcell{#3} &
  \imgcell{#4}
  &
  \imgcellgood{#5}
  {\fontsize{8}{8}\selectfont
  E:\textbf{6}\quad
  P:\textbf{6}\quad
  Q:\textbf{6}} 
  &
  \imgcellgood{#6}
  {\fontsize{8}{8}\selectfont
  E:\textbf{5}\quad
  P:\textbf{5}\quad
  Q:\textbf{6}} 
  \\
}

\newcommand{\exrowtwosix}[6]{%
  \imgcell{#1} &
  \desc{#2} &
  \imgcell{#3} &
  \imgcellverygood{#4}
  {\fontsize{8}{8}\selectfont
  E:\textbf{6}\quad
  P:\textbf{6}\quad
  Q:\textbf{7}} 
  &
  \imgcellgood{#5}
  {\fontsize{8}{8}\selectfont
  E:\textbf{6}\quad
  P:\textbf{4}\quad
  Q:\textbf{6}} 
  &
  \imgcellverygood{#6}
  {\fontsize{8}{8}\selectfont
  E:\textbf{7}\quad
  P:\textbf{6}\quad
  Q:\textbf{6}} 
  \\
}

\newcommand{\exrowtwoseven}[6]{%
  \imgcell{#1} &
  \desc{#2} &
  \imgcell{#3} &
  \imgcellgood{#4}
  {\fontsize{8}{8}\selectfont
  E:\textbf{5}\quad
  P:\textbf{6}\quad
  Q:\textbf{5}} 
  &
  \imgcellgood{#5}
  {\fontsize{8}{8}\selectfont
  E:\textbf{6}\quad
  P:\textbf{5}\quad
  Q:\textbf{6}} 
  &
  \imgcellgood{#6}
  {\fontsize{8}{8}\selectfont
  E:\textbf{5}\quad
  P:\textbf{4}\quad
  Q:\textbf{7}} 
  \\
}

\begin{table}[t]
\centering
\caption{Scientific figure edits by Gemini-3.1-Pro, GPT-5.6-Sol, and EdiTikZ-9B-RL. Models receive the source image and VLM-generated edit instruction. Human annotations score edit application (E), source preservation (P), and visual quality (Q). Overall quality: \legendbox{green} very good, \legendbox{yellow} good, \legendbox{orange} bad, \legendbox{red} very bad.}
\label{tab:editikz_examples_1}
\begin{tabularx}{\textwidth}{@{}CYCCCC@{}}
    \toprule
    \headercell{Source} &
    \headercell{Edit Instruction} &
    \headercell{Ground Truth} &
    \headercell{Gemini-3.1-Pro} &
    \headercell{GPT-5.6-Sol} &
    \headercell{EdiTikZ-9B-RL} \\
    \midrule
    \exrowtwoone
    {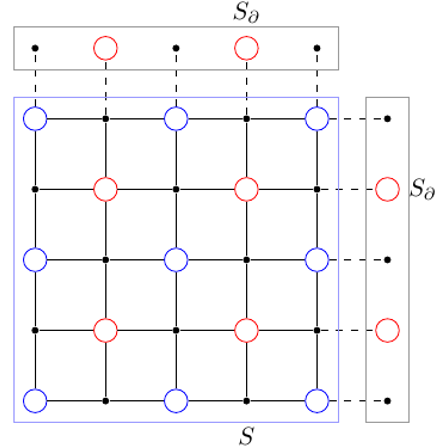}
    {The grid connectivity is changed. Specifically, the dashed lines connecting the outer boundary nodes to the main grid are replaced by solid lines. Solid lines are added between neighboring nodes nodes of the outer boundary. The blue rectangular region labeled 'S' is resized and repositioned. It now encloses the top-right 3x3 block of the grid, whereas it previously enclosed the entire 4x4 blocks. The label '\smath{S_{\partial}}' at the top is changed to '\smath{T}'. The label '\smath{S_{\partial}}' on the right is changed to '\smath{T}'. A new red rectangular box with a red node is added in the top-right corner. A new label '\smath{U}' is added above this red box.}
    {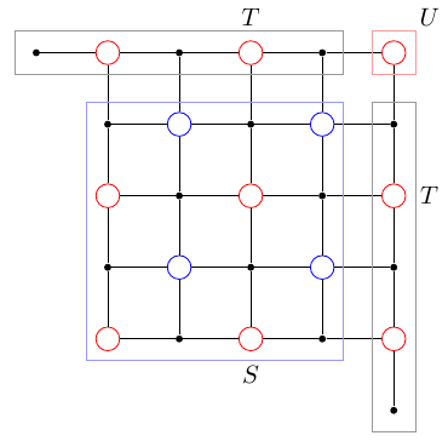}
    {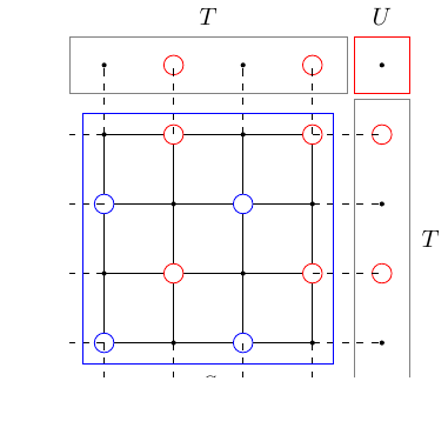}
    {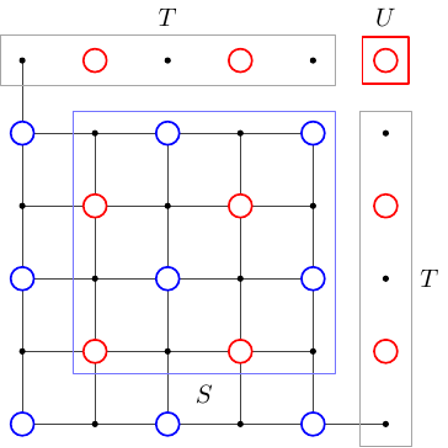}
    {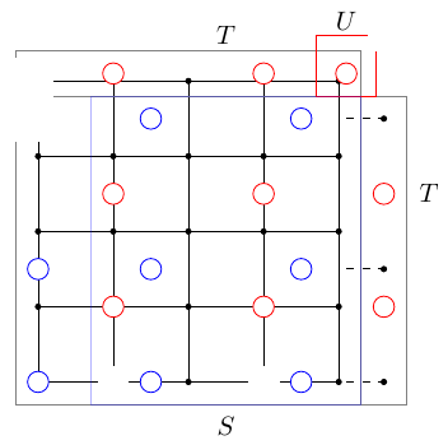}
    \midrule
    \exrowtwotwo
    {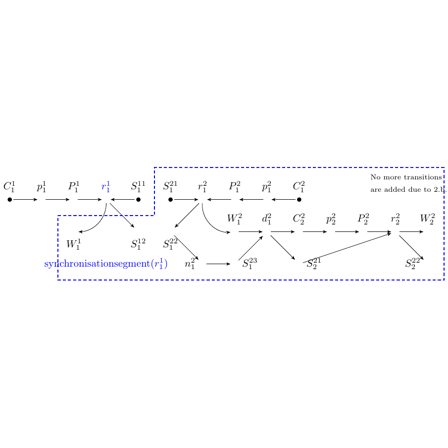}
    {Remove the blue transition node labeled \smath{r^1_1} and the nodes labeled \smath{p^1_1}, \smath{P^1_1}, \smath{W^1_1}, and \smath{S^12_1} and all connecting arrows at the left side of the diagram. Move the node labeled \smath{C^1_1} to the position previously occupied by \smath{r^1_1}, directly to the left of \smath{S^{11}_1}. Remove the blue dashed rectangular box and its label 'synchronisationsegment(\smath{r^1_1})'. Remove the text annotation 'No more transitions are added due to 2.b.' located in the top right corner. Add a red dashed path connecting \smath{C^1_1} to \smath{S^{11}_1}, then to \smath{S^{22}_1}, and finally to \smath{W_1^2}. Add the red text label '\smath{\text{lkc}(r^2_1)}' positioned below the red path to \smath{S^{22}_1}. Add an short vertical olive dashed path connecting \smath{W_2^2} to \smath{S^{22}_2} labeled '\smath{\text{lkc}(r^2_2)}' in olive to the right of the path. Add another short horizontal path connecting \smath{C_1^1} to \smath{S^{11}_1} just below the red path. The two middle points of these short paths are connected with two olive dashed line segments extending from the top left downwards and then to the right.}
    {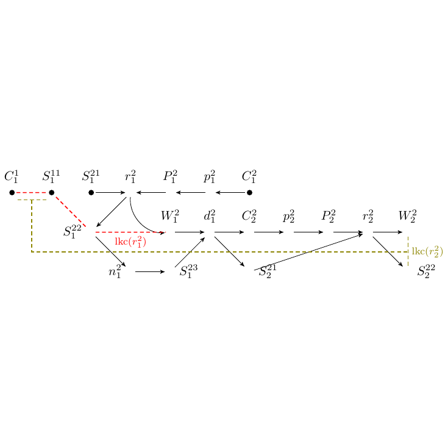}
    {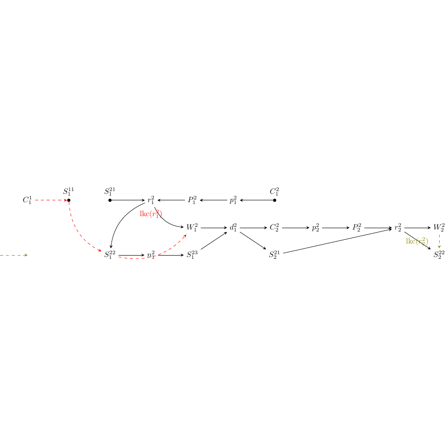}
    {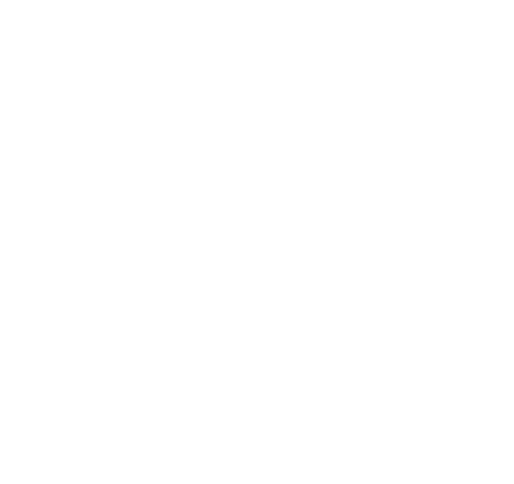}
    {structure/examples_2/not_compiled.png}
    \midrule
    \exrowtwothree
    {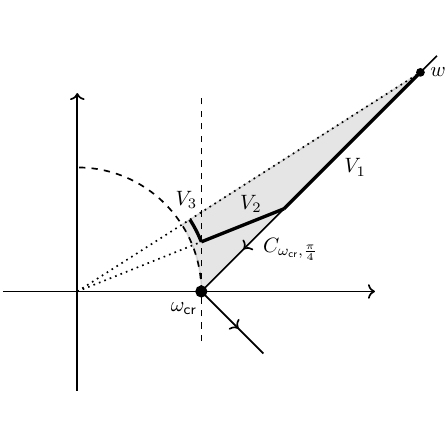}
    {The main diagonal ray and its vertex move leftward from approximately x=0.5 to x=0.3. The ray label changes from \smath{C_{\omega_{\mathsf{cr}},\frac{\pi}{4}}} to \smath{C_{a,\frac{\pi}{4}}} and move towards top right. The shaded region, dashed vertical line, dashed quarter-circle, solid arc, and segmented ultra-thick path to \smath{w} are removed. The labels \smath{V_1}, \smath{V_2}, \smath{V_3}, and \smath{\omega_{\mathsf{cr}}} are removed. A circular contour labeled \smath{\Gamma_w} is added near the origin, with arrowed arcs and two short arrowed internal segments around the dotted line to \smath{w}. Two small red points are added along the dotted line and one blue point is added on the lower-left part of \smath{\Gamma_w}. A horizontal double-headed arrow labeled \smath{q^R} is added below the x-axis. The main rays and the newly restored contour segments are drawn very thick.}
    {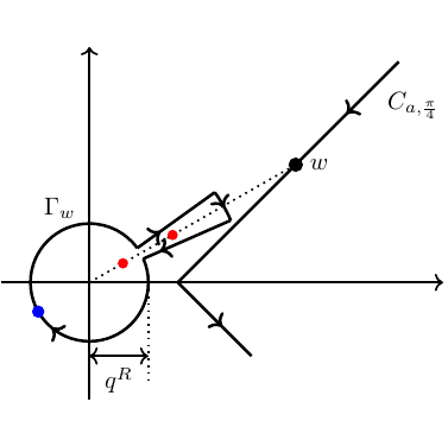}
    {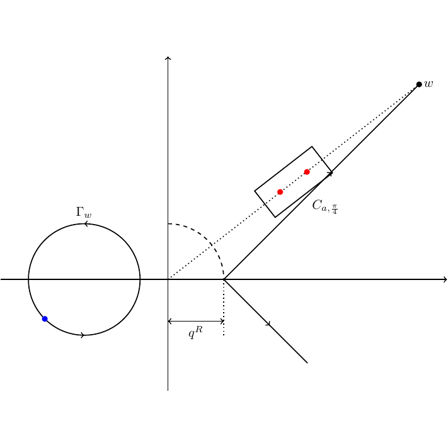}
    {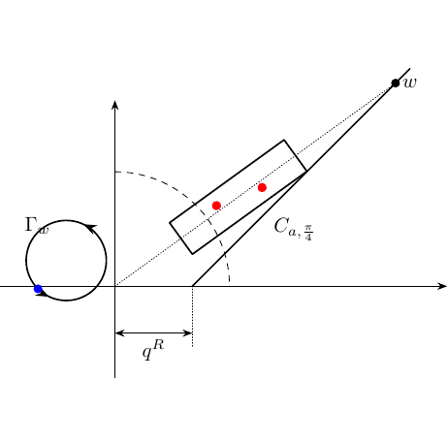}
    {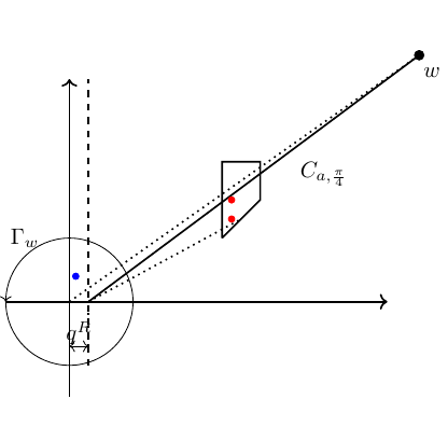}
    \midrule
    \exrowtwofour
    {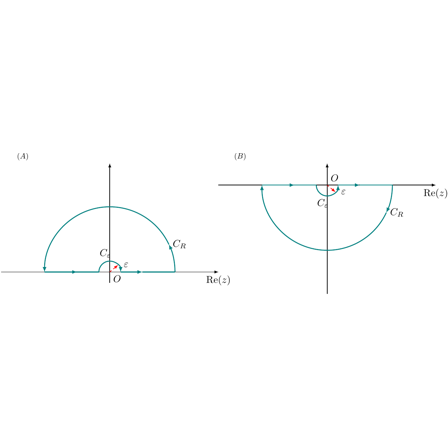}
    {Remove the second panel (labeled (B)) located on the right side of the figure, which depicts a contour in the lower half-plane. Remove the panel label '\smath{(A)}' from the top-left corner of the remaining diagram. Change the label '\smath{C_R}' on the large outer semicircle to '\smath{C_2}'. Change the label '\smath{C_\varepsilon}' on the small inner semicircle to '\smath{C_1}'. Change the label '\smath{\varepsilon}' indicating the radius of the small semicircle to '\smath{r}'. Change the label '\smath{O}' at the origin to '\smath{x_0}'. Add the label '\smath{R}' at the rightmost point of the large semicircle on the real axis. Remove the arrowhead from the red dashed line representing the radius \smath{r}.}
    {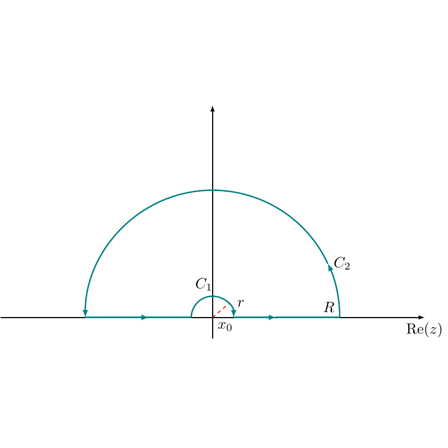}
    {structure/examples_2/not_compiled.png}
    {structure/examples_2/not_compiled.png}
    {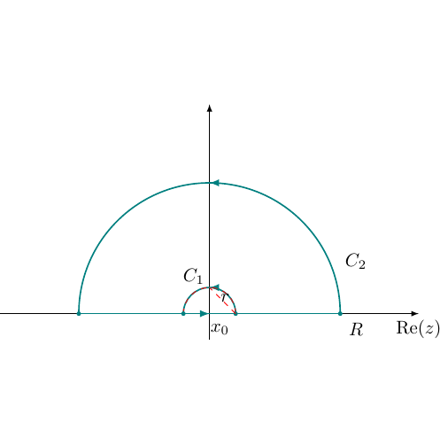}
    \midrule
    \exrowtwofive
    {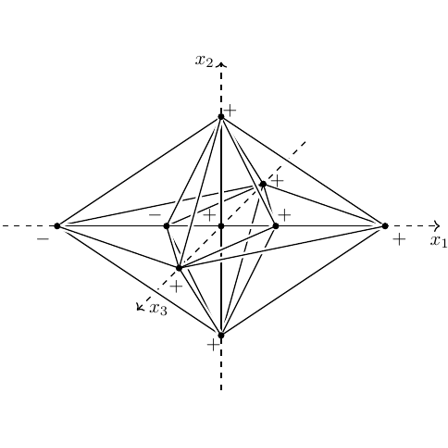}
    {The \smath{x_3} axis (dashed line pointing towards the viewer) is extended to a longer length, and its label '\smath{x_3}' is moved further out along the axis. The sign label '\smath{+}' at the point on the positive \smath{x_1} axis (inner point) is changed to '\smath{-}'. The sign label '\smath{-}' at the point on the negative \smath{x_1} axis (inner point) is changed to '\smath{+}'. The sign label '\smath{+}' at the point on the positive \smath{x_3} axis (front point) is changed to '\smath{-}'. The sign label '\smath{+}' at the point on the negative \smath{x_3} axis (back point) is changed to '\smath{-}'. A gray shaded region is added covering a large irregular area inside the diagram, roughly spanning from the left vertex marked '\smath{-}' across the central area toward the right vertex marked '\smath{+}', overlaying the underlying triangulated structure. Two closed red outline shapes are added: a larger red polygon/curve (partly dashed, partly solid) tracing an irregular boundary around the left-to-center portion of the diagram near the '\smath{-}' vertex, and a smaller red triangular outline near the right-center portion of the diagram close to the '\smath{-}' label on the right side, both overlapping the gray shaded region.}
    {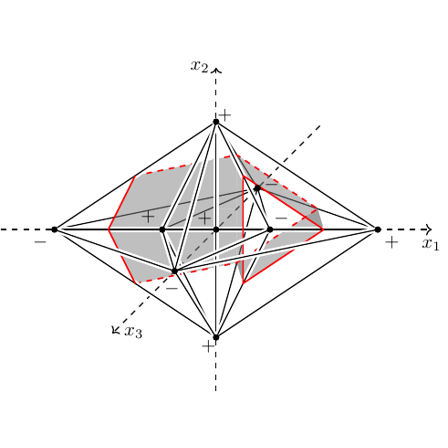}
    {structure/examples_2/not_compiled.png}
    {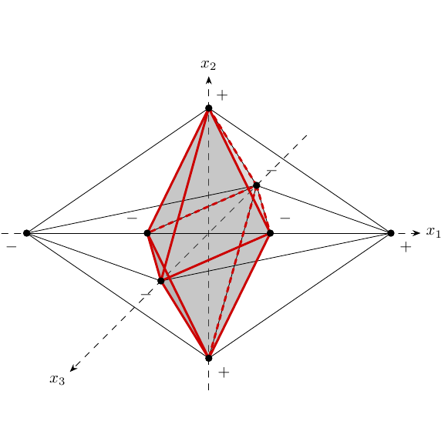}
    {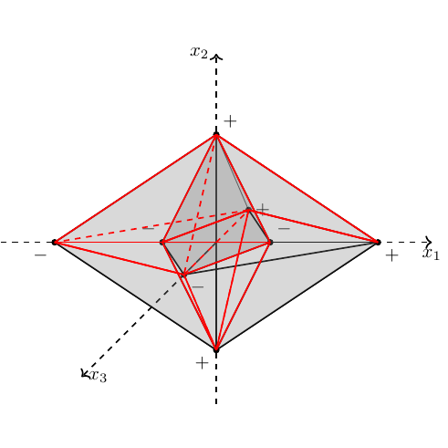}
    \midrule
    \exrowtwosix
    {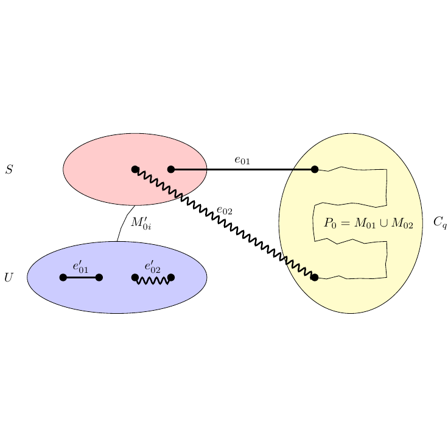}
    {The wavy line connecting the red ellipse (S) to the yellow ellipse (Cq) is removed together with its label ‘e02’. The wavy line connecting the two points inside the blue ellipse (U) is removed. A wavy line connecting the blue ellipse (U) to the yellow ellipse (Cq) is added to connect the rightmost point in U to the bottom point in Cq at it is labeled ‘e11’. The label 'e01' on the top horizontal line is changed to 'e12'. The label 'e01'' on the left horizontal line in U is changed to 'e13'. The label 'e02'' on the right wavy line in U is removed. The label 'M0i'' is changed to 'M1i''. The label 'P0 = M01 U M02' inside the yellow ellipse is changed to 'P1 = M11 U M12'.}
    {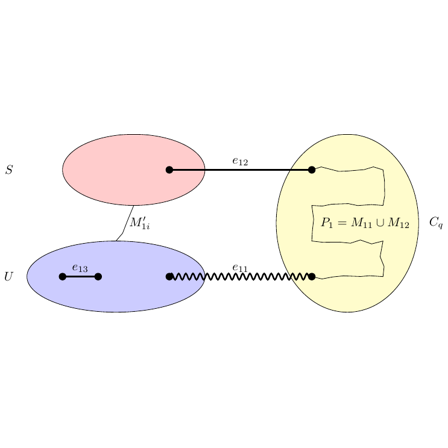}
    {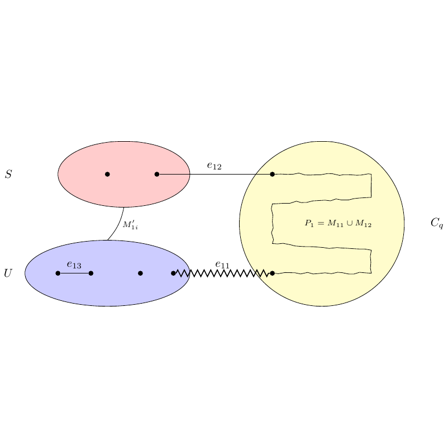}
    {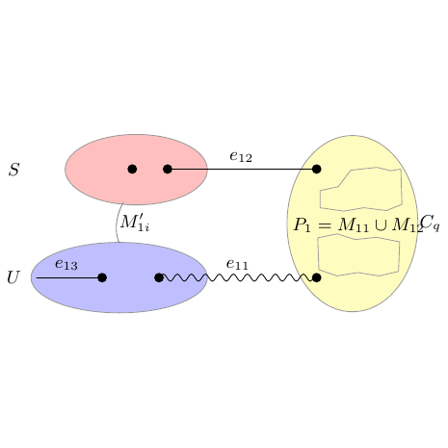}
    {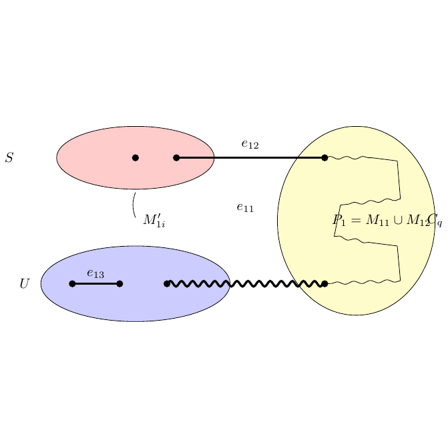}
    \midrule
    \exrowtwoseven
    {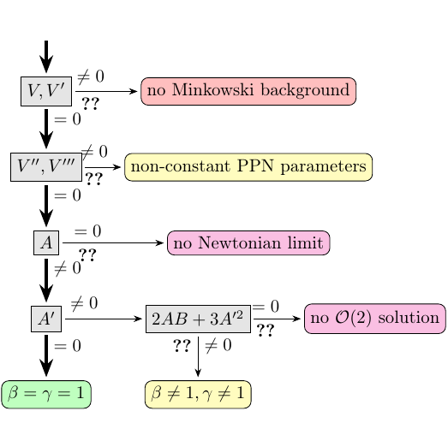}
    {The yellow result node 'beta != 1, gamma != 1' (positioned beside the bottom-row 'beta = gamma = 1' node) is repositioned to be reached from the 'C' branch's '\smath{2AB+3C^2}' box via the '!=0' path. The green result node labeled 'beta = gamma = 1' is moved from the bottom of the main vertical flow, positioned directly below 'A'' and reached by the '=0' branch arrow, to its original position on the right side, reached via a bold arrow from 'A''. The entire lower section of the flowchart is added. This includes the nodes 'B', 'B'', 'A'B'', and '\smath{2A'B' - C'^2}', the result node 'no O(4) solution', and the arrows connecting these added nodes. The mathematical expression in the decision box to the right of the fourth level is changed from '\smath{2AB + 3A'^2}' to '\smath{2AB + 3C^2}', with its '=0' branch continuing to point to 'no O(2) solution' and its '??' branch left unresolved as before. The decision node labeled 'A'' at the fourth level of the flowchart is changed to a node labeled 'C'.}
    {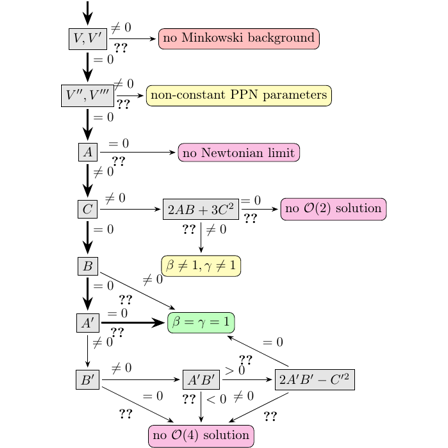}
    {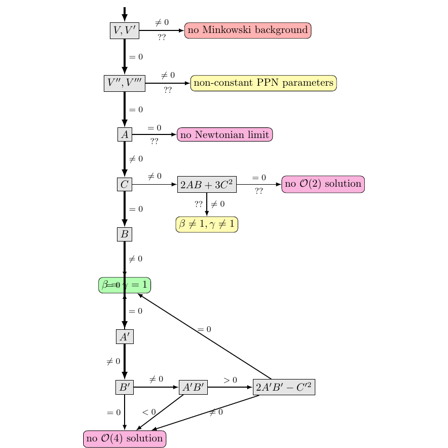}
    {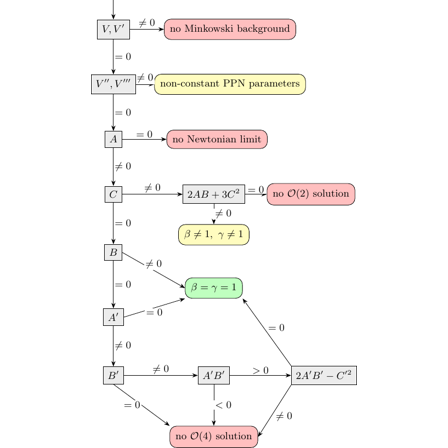}
    {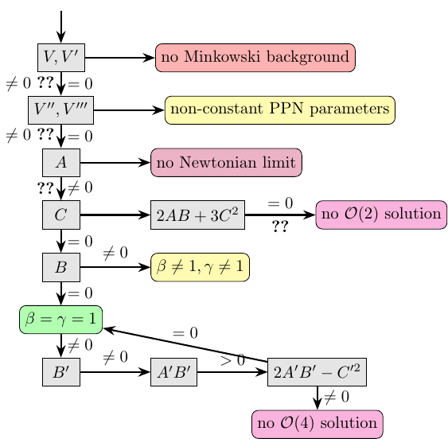}
    
    \bottomrule
\end{tabularx}
\end{table}

\setlength{\aboverulesep}{0pt}
\setlength{\belowrulesep}{0pt}
\setlength{\tabcolsep}{2pt}
\renewcommand{\arraystretch}{1.03}
\setlength{\abovetopsep}{0pt}

\setlength{\descw}{0.25\textwidth}

\setlength{\imgw}{0.136\textwidth}

\setlength{\imgh}{2.6cm}

\newcolumntype{Y}{>{\raggedright\arraybackslash}p{\descw}}
\newcolumntype{C}{>{\centering\arraybackslash}p{\imgw}}

\newcommand{\descsmall}[1]{%
  \begin{minipage}[t]{\linewidth}
    \vspace{-6pt}%
    \raggedright
    \fontsize{3.5}{4.0}\selectfont
    \setlength{\parskip}{0pt}%
    \setlength{\parindent}{0pt}%
    #1%
  \end{minipage}%
}

\newcommand{\exrowthreeone}[6]{%
  \imgcell{#1} &
  \desc{#2} &
  \imgcell{#3} &
  \imgcellverygood{#4}
  {\fontsize{8}{8}\selectfont
  E:\textbf{7}\quad
  P:\textbf{7}\quad
  Q:\textbf{5}} 
  &
  \imgcellokay{#5}
  {\fontsize{8}{8}\selectfont
  E:\textbf{7}\quad
  P:\textbf{3}\quad
  Q:\textbf{1}} 
  &
  \imgcellverygood{#6}
  {\fontsize{8}{8}\selectfont
  E:\textbf{7}\quad
  P:\textbf{7}\quad
  Q:\textbf{7}} 
  \\
}

\newcommand{\exrowthreetwo}[6]{%
  \imgcell{#1} &
  \desc{#2} &
  \imgcell{#3} &
  \imgcell{#4}
  &
  \imgcell{#5}
  &
  \imgcellverygood{#6}
  {\fontsize{8}{8}\selectfont
  E:\textbf{5}\quad
  P:\textbf{7}\quad
  Q:\textbf{7}} 
  \\
}

\newcommand{\exrowthreethree}[6]{%
  \imgcell{#1} &
  \desc{#2} &
  \imgcell{#3} &
  \imgcellverygood{#4}
  {\fontsize{8}{8}\selectfont
  E:\textbf{6}\quad
  P:\textbf{6}\quad
  Q:\textbf{7}} 
  &
  \imgcellbad{#5}
  {\fontsize{8}{8}\selectfont
  E:\textbf{4}\quad
  P:\textbf{2}\quad
  Q:\textbf{2}} 
  &
  \imgcellgood{#6}
  {\fontsize{8}{8}\selectfont
  E:\textbf{6}\quad
  P:\textbf{5}\quad
  Q:\textbf{7}} 
  \\
}

\newcommand{\exrowthreefour}[6]{%
  \imgcell{#1} &
  \desc{#2} &
  \imgcell{#3} &
  \imgcellokay{#4}
  {\fontsize{8}{8}\selectfont
  E:\textbf{3}\quad
  P:\textbf{4}\quad
  Q:\textbf{6}} 
  &
  \imgcell{#5}
  &
  \imgcellverygood{#6}
  {\fontsize{8}{8}\selectfont
  E:\textbf{7}\quad
  P:\textbf{7}\quad
  Q:\textbf{7}} 
  \\
}

\newcommand{\exrowthreefive}[6]{%
  \imgcell{#1} &
  \desc{#2} &
  \imgcell{#3} &
  \imgcellgood{#4}
  {\fontsize{8}{8}\selectfont
  E:\textbf{4}\quad
  P:\textbf{6}\quad
  Q:\textbf{6}} 
  &
  \imgcellokay{#5}
  {\fontsize{8}{8}\selectfont
  E:\textbf{3}\quad
  P:\textbf{5}\quad
  Q:\textbf{2}} 
  &
  \imgcellverygood{#6}
  {\fontsize{8}{8}\selectfont
  E:\textbf{7}\quad
  P:\textbf{6}\quad
  Q:\textbf{6}} 
  \\
}

\newcommand{\exrowthreesix}[6]{%
  \imgcell{#1} &
  \desc{#2} &
  \imgcell{#3} &
  \imgcell{#4}
  &
  \imgcell{#5}
  &
  \imgcellbad{#6}
  {\fontsize{8}{8}\selectfont
  E:\textbf{3}\quad
  P:\textbf{2}\quad
  Q:\textbf{1}} 
  \\
}

\newcommand{\exrowthreeseven}[6]{%
  \imgcell{#1} &
  \descsmall{#2} &
  \imgcell{#3} &
  \imgcellokay{#4}
  {\fontsize{8}{8}\selectfont
  E:\textbf{4}\quad
  P:\textbf{4}\quad
  Q:\textbf{6}} 
  &
  \imgcellbad{#5}
  {\fontsize{8}{8}\selectfont
  E:\textbf{2}\quad
  P:\textbf{2}\quad
  Q:\textbf{5}} 
  &
  \imgcellokay{#6}
  {\fontsize{8}{8}\selectfont
  E:\textbf{3}\quad
  P:\textbf{3}\quad
  Q:\textbf{5}} 
  \\
}

\begin{table}[t]
\centering
\caption{Scientific figure edits by GPT-5.6-Sol, Qwen3.5-4B, and EdiTikZ-4B-RL. Models receive the source image and VLM-generated edit instruction. Human annotations score edit application (E), source preservation (P), and visual quality (Q). Overall quality: \legendbox{green} very good, \legendbox{yellow} good, \legendbox{orange} bad, \legendbox{red} very bad.}
\label{tab:editikz_examples_1}
\begin{tabularx}{\textwidth}{@{}CYCCCC@{}}
    \toprule
    \headercell{Source} &
    \headercell{Edit Instruction} &
    \headercell{Ground Truth} &
    \headercell{GPT-5.6-Sol} &
    \headercell{Qwen3.5-4B} &
    \headercell{EdiTikZ-4B-RL} \\
    \midrule
    \exrowthreeone
    {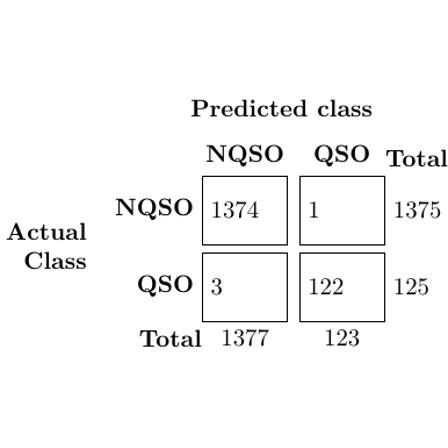}
    {The value in the top-left cell (Actual NQSO, Predicted NQSO) changed from 1374 to 1213. The value in the top-right cell (Actual NQSO, Predicted QSO) changed from 1 to 1073. The value in the bottom-left cell (Actual QSO, Predicted NQSO) changed from 3 to 1. The value in the bottom-right cell (Actual QSO, Predicted QSO) changed from 122 to 213.}
    {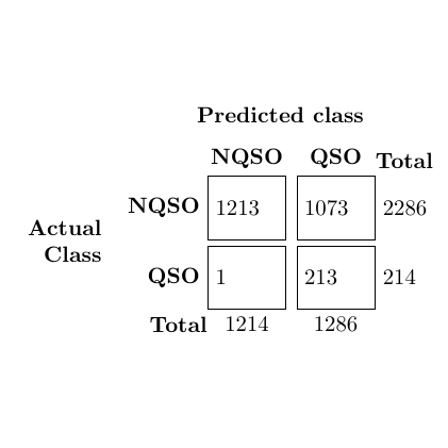}
    {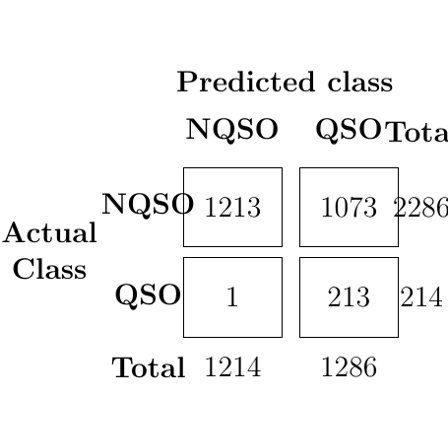}
    {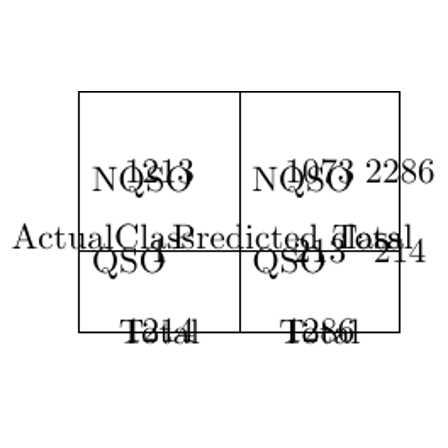}
    {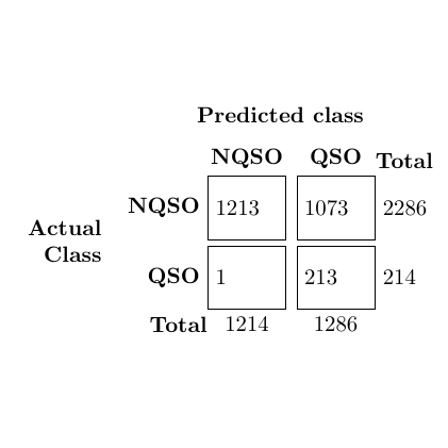}
    \midrule
    \exrowthreetwo
    {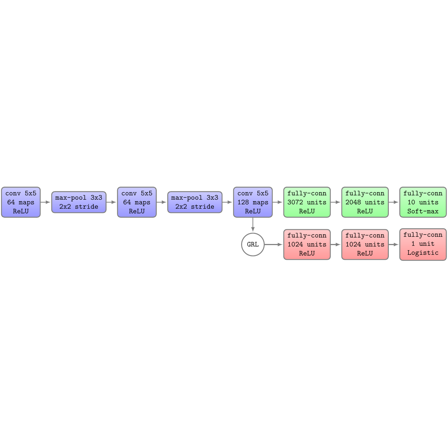}
    {In the first blue node, change the text '64 maps' to '32 maps'. In the second blue node (max-pool), change the text '3x3' to '2x2'. In the third blue node, change the text '64 maps' to '48 maps'. In the fourth blue node (max-pool), change the text '3x3' to '2x2'. Remove the fifth blue node labeled 'conv 5x5 128 maps ReLU' from the main sequence. In the first green node, change the text '3072 units' to '100 units'. In the second green node, change the text '2048 units' to '100 units'. Remove the first red node labeled 'fully-conn 1024 units ReLU' from the bottom branch. In the second red node, change the text '1024 units' to '100 units'. Move the 'GRL' circle node one position to the left so it is located below the second max-pool node instead of the third convolution node. Change the connection arrow pointing to the 'GRL' node so it originates from the second max-pool node instead of the third convolution node. Change the connection arrow pointing to the first green node so it originates from the second max-pool node instead of the third convolution node. Change the node fill colors from clue to grey, from bright green to a darker green, and from a bright red to a darker red/brown.}
    {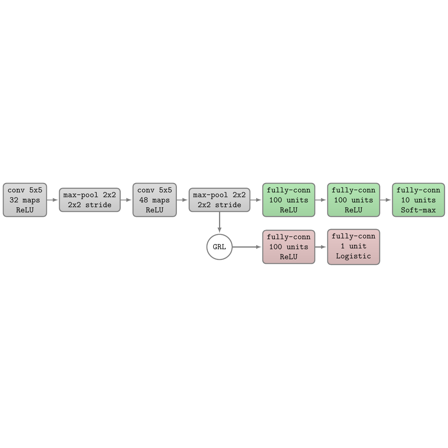}
    {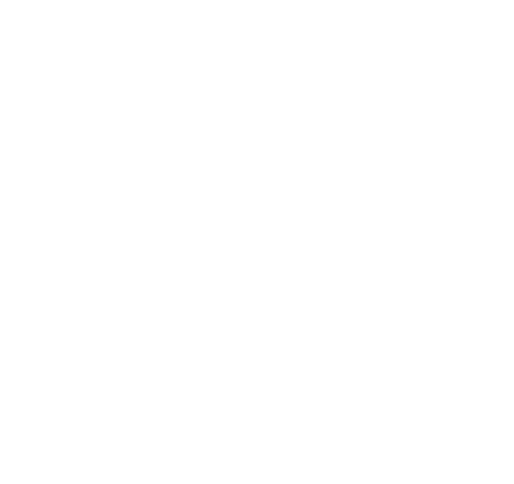}
    {structure/examples_3/not_compiled.png}
    {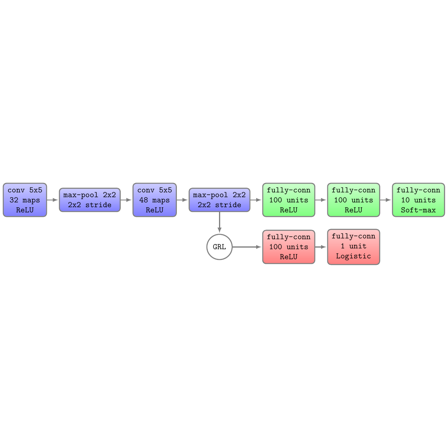}
    \midrule
    \exrowthreethree
    {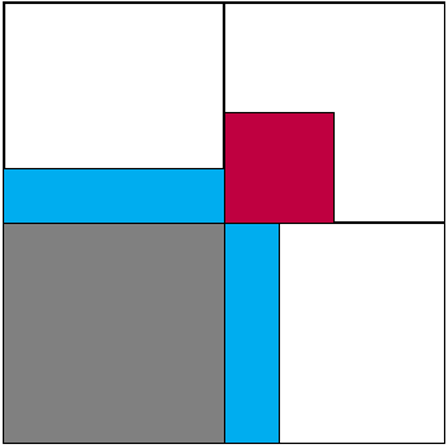}
    {The purple square in the top-right quadrant is replaced by a purple right-angled triangle. The triangle's vertices are at the center of the figure, the point to the right of the center on the horizontal midline, and the point above the center on the vertical midline. Two small red right-angled triangles are added. One is located in the top-left quadrant, adjacent to the top-left corner of the new purple triangle. The other is in the bottom-right quadrant, adjacent to the bottom-right corner of the new purple triangle. Four mathematical labels are added to the four main quadrants of the figure. The label '\smath{d^{1/3}}' is added to the gray square in the bottom-left. The label '\smath{d^{1/2}}' is added to the white square in the top-left. The label '\smath{d^{1/2}}' is added to the white square in the bottom-right. The label '\smath{d}' is added to the white square in the top-right.}
    {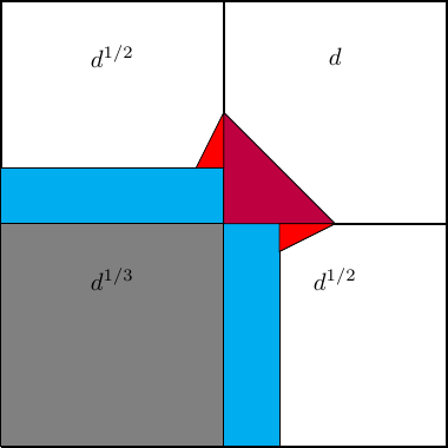}
    {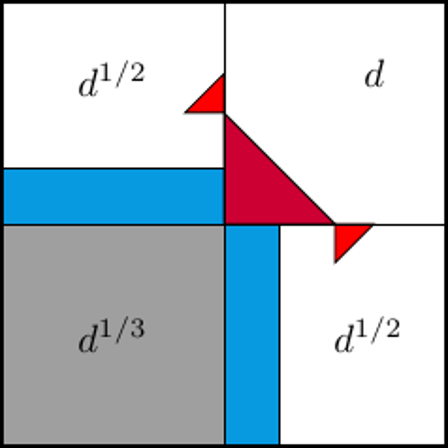}
    {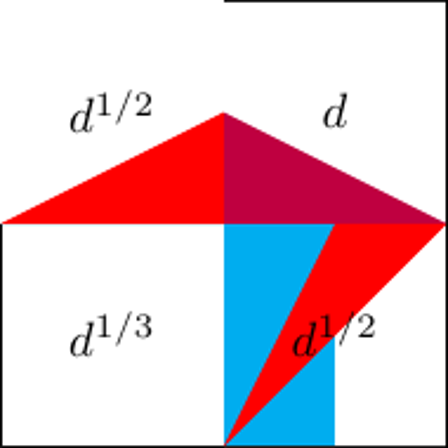}
    {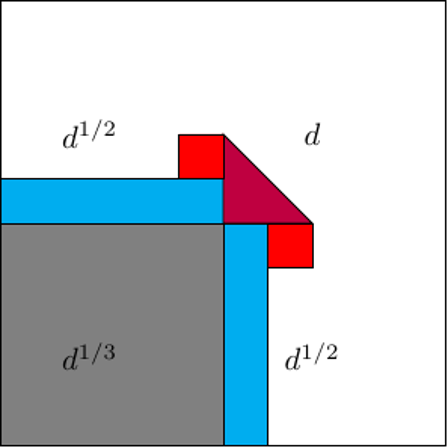}
    \midrule
    \exrowthreefour
    {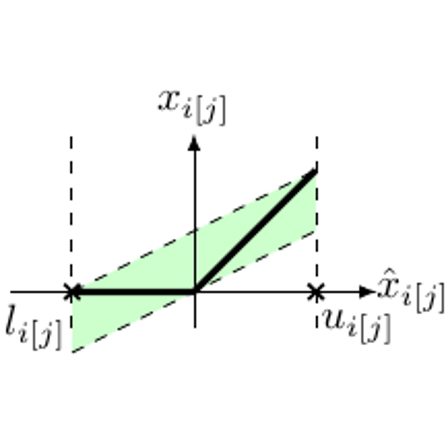}
    {The light green shaded region is changed from a parallelogram (bounded by y=0.5x and y=0.5x-0.5) to a triangle (bounded by the x-axis, the line y=0.5x, and the vertical line x=1). The dashed line segment forming the lower boundary of the original parallelogram (from (0, -0.5) to (2, 0.5)) is removed. The dashed line segment forming the upper boundary of the region (from (0,0) to (2,1)) is made thicker (ultra thick). A thick, red vertical line segment is added, extending from the origin (0,0) upwards to (0, 0.5). The vertical dashed lines at x=0 and x=2 are extended downwards to y=-0.5. The horizontal axis arrow is extended further to the right, and its label '\smath{\hat{x}_{i[j]}}' is moved further to the right. The vertical axis arrow is extended further downwards and upwards, and its label '\smath{x_{i[j]}}' is moved further upwards.}
    {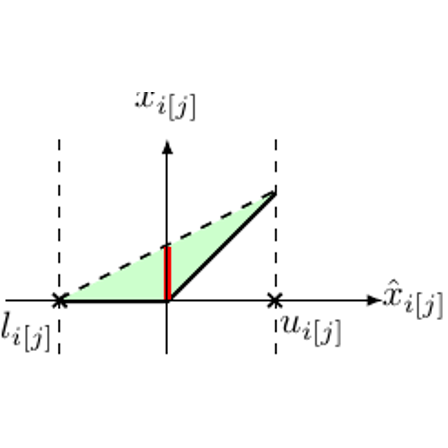}
    {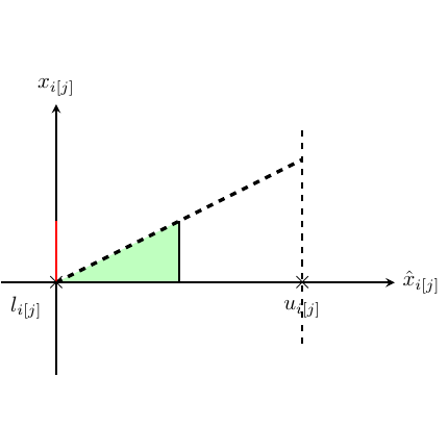}
    {structure/examples_3/not_compiled.png}
    {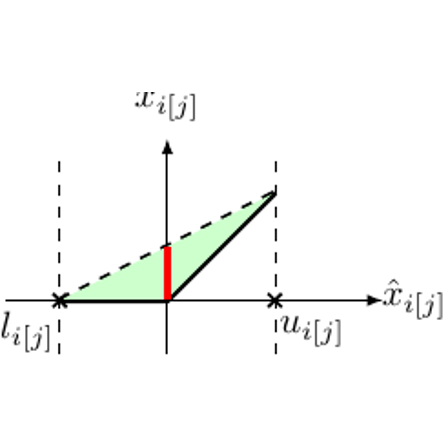}
    \midrule
    \exrowthreefive
    {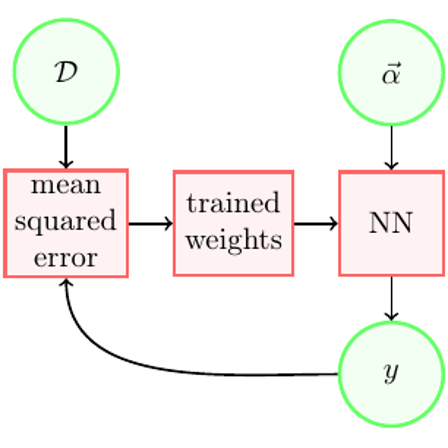}
    {Change the text in the top-right square node from 'NN' to 'kernel \ matrices'. Change the text in the middle square node from 'trained \ weights' to '\smath{\Theta}, \smath{\sigma_n}' and change its shape from a square to a circle. Change the text in the left square node from 'mean \ squared \ error' to 'Negative \ Log \ Likelihood'. Change the text in the bottom-right circle node from '$y$' to '\smath{\mathcal{N}(\mu, \sigma)}' and change its shape from a circle to a square. Move the top-left circle node labeled '\smath{\mathcal{D}}' from above the left square node to above the middle circle node. Change the straight arrow from the top-left circle '\smath{\mathcal{D}}' to point to the left square node 'Negative Log Likelihood' with a curved arrow. Add a new curved arrow from the top-left circle '\smath{\mathcal{D}}' pointing to the top-right square node 'kernel matrices'.}
    {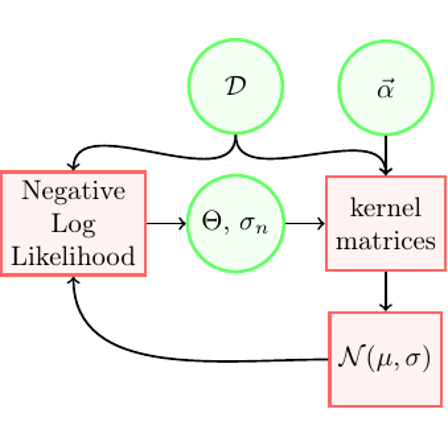}
    {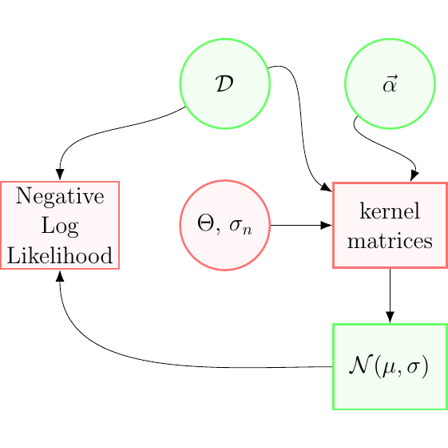}
    {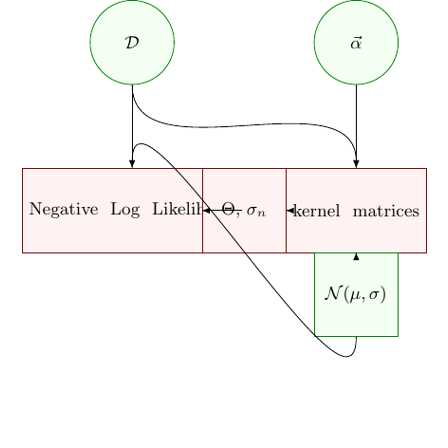}
    {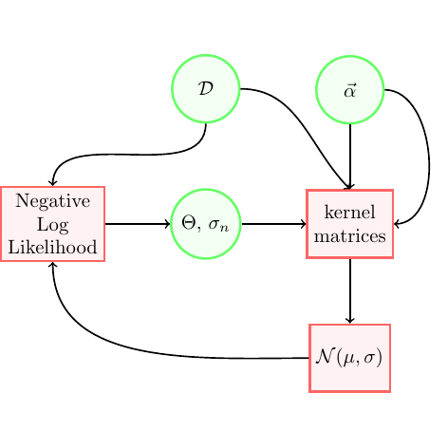}
    \midrule
    \exrowthreesix
    {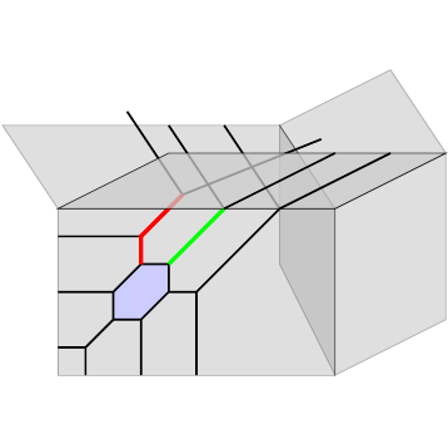}
    {The light blue filled hexagonal region is removed from the front face of the structure, located between the red and green highlighted paths. The red line is shortened to touch only the edge connecting the front and upper faces. The short black grid line on the upper-left top surface extending farther backwards is shortened. The black diagonal grid-line connected to the red line extending farther backwards on the upper-right top face is now shortened.}
    {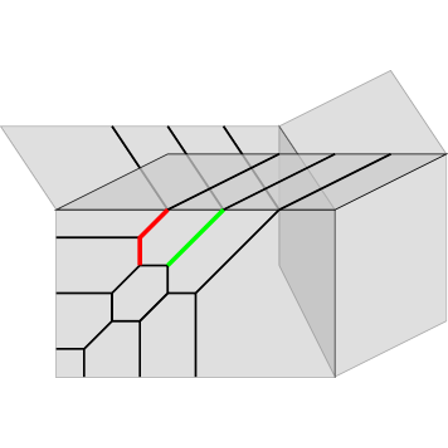}
    {structure/examples_3/not_compiled.png}
    {structure/examples_3/not_compiled.png}
    {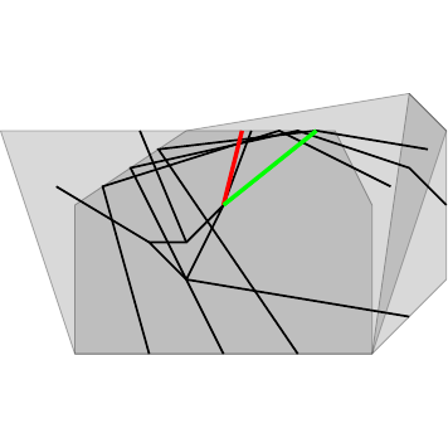}
    \midrule
    \exrowthreeseven
    {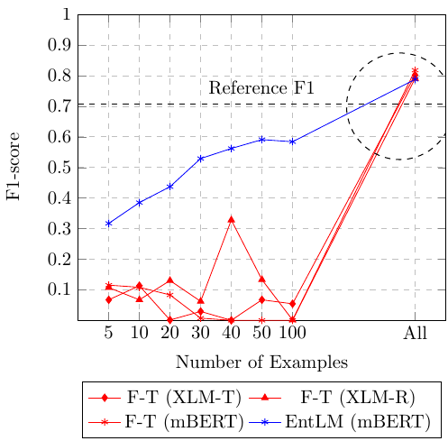}
    {The horizontal dashed line at y=0.707 spanning the width of the plot, labeled 'Reference F1', is removed. The legend entries are updated to: 'F-T (XLM-T)', 'F-T (XLM-T sentiment)', 'PET (XLM-T)', and 'PET (XLM-T sentiment)'. The single dashed circle annotation (without a label) enclosing the convergence point of all curves near x='All' is removed. Two dashed oval annotations are added: one labeled '(a)' positioned around the point at x=100, and one labeled '(b)' positioned around the cluster of points near x='All'. The data series using red asterisk markers (F-T mBERT), staying low and volatile around 0.05-0.14 across most x-values with minor fluctuations, then rising sharply to approximately 0.83 at 'All', is removed. A new data series using blue triangle markers (PET XLM-T sentiment) is added as a distinct series. The plotted data series for the blue diamond-marker curve (previously EntLM mBERT) is replaced with new coordinate values representing PET (XLM-T), starting at approximately 0.53 (x=5), rising through 0.61 (x=10), 0.67 (x=20), 0.82 (x=30), 0.73 (x=40), 0.76 (x=50), 0.79 (x=100), and reaching approximately 0.87 at 'All'. The plotted data series for the red triangle-marker curve (previously F-T XLM-R) is replaced with new coordinate values representing F-T (XLM-T sentiment), starting high at approximately 0.92 (x=5), rising to 0.94 (x=10), staying around 0.92 (x=20), dipping to 0.87 (x=30-40), rising to 0.92 (x=50), dropping to 0.8 (x=100), and reaching approximately 0.91 at 'All'. The plotted data series for the red diamond-marker curve (previously F-T XLM-T on F1-score scale) is replaced with new coordinate values representing F-T (XLM-T) on the accuracy scale, starting at approximately 0.43 (x=5), dropping to 0.38 (x=10), then rising steadily through 0.5 (x=20), 0.55 (x=30), 0.62 (x=40), 0.64 (x=50), 0.8 (x=100), and reaching approximately 0.87 at 'All'. The Y-axis label is changed from 'F1-score' to 'Accuracy'.}
    {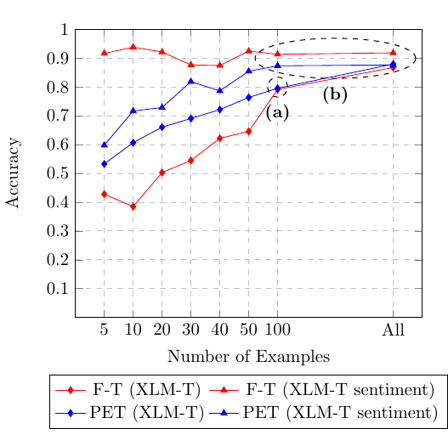}
    {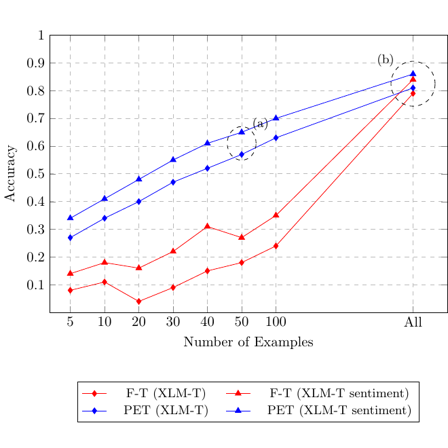}
    {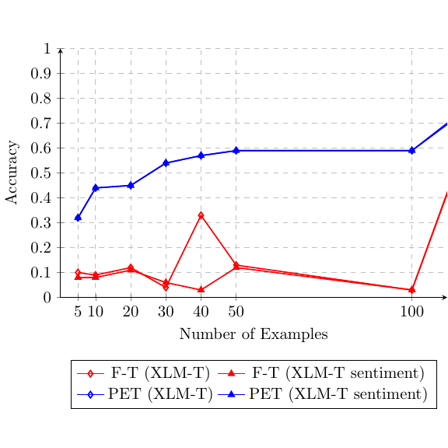}
    {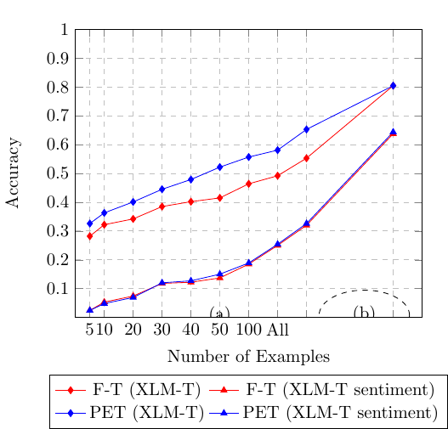}
    \bottomrule
\end{tabularx}
\end{table}

\end{document}